\documentclass{article}
\usepackage{PRIMEarxiv}
\usepackage{float}
\usepackage{xargs}
\usepackage{multirow}
\usepackage{tikz}       % Required by forest
\usetikzlibrary{positioning}  % for "below=..., 
\usetikzlibrary{backgrounds,calc}
\usepackage{forest}     % Tree drawing
\usepackage{xcolor}     % Colors
\usepackage{subcaption}

\usepackage{cite}       % For handling citations properly
\usepackage{adjustbox}
\usepackage[utf8]{inputenc} % allow utf-8 input
\usepackage[T1]{fontenc}    % use 8-bit T1 fonts
\usepackage{hyperref}       % hyperlinks
\usepackage{url}            % simple URL typesetting
\usepackage{booktabs}       % professional-quality tables
\usepackage{amsfonts}       % blackboard math symbols
\usepackage{nicefrac}       % compact symbols for 1/2, etc.
\usepackage{microtype}      % microtypography
\usepackage{lipsum}
\usepackage{fancyhdr}       % header
\usepackage{graphicx}       % graphics
\graphicspath{{media/}}     % organize your images and other figures under 
\usepackage{todonotes}
\usepackage{amsmath, amssymb}
\usepackage{tikz}
\usepackage{pdflscape}
\usepackage{mathtools}
\newcommand\mat[1]{\mathbf{#1}}

\renewcommand\vec[1]{\mathbf{#1}}

\renewcommand\vec[1]{\mathbf{#1}}

\renewcommand{\l}{\left}
\renewcommand{\r}{\right}

\usetikzlibrary{arrows.meta,positioning,shadows,shapes.misc}
\usepackage[dvipsnames]{xcolor}

\newcommand{\myparagraph}[1]{%
  \vspace{0.6em}\noindent\textbf{#1}\par\vspace{0.em}%
}

\title{Mixed-Precision Quantization for Language Models: Techniques and Prospects}

\author{%
\begin{tabular}{@{}c@{}}
Mariam Rakka$^1$, Marios Fournarakis$^2$, Olga Krestinskaya$^3$, Jinane Bazzi$^3$, \\ 
Khaled N. Salama$^3$, Fadi Kurdahi$^1$, Ahmed M. Eltawil$^3$, Mohammed E. Fouda$^{4,*}$
\end{tabular}\\[0.6em]
$^1$University of California Irvine, Irvine, CA, USA,\\
$^2$Wayve AI, London, UK,\\
$^3$King Abdullah University of Science and Technology, Thuwal, Saudi Arabia,\\
$^4$RAIN AI, San Francisco, USA. \\
$^*$ Email: \texttt{foudam@uci.edu}%
}

\begin{document}
\maketitle

\begin{abstract}
{
The rapid scaling of language models (LMs) has resulted in unprecedented computational, memory, and energy requirements, making their training and deployment increasingly unsustainable. Quantization has emerged as an essential compression technique to reduce model size, alleviate memory bottlenecks, and accelerate inference. However, while uniform low-bit quantization (e.g., INT8, INT4) provides significant efficiency gains, it can degrade accuracy in sensitive components of transformer-based LMs. Mixed-precision quantization offers a promising alternative by selectively allocating precision across layers or within tensors to balance efficiency and accuracy. This survey provides a comprehensive overview of Mixed-Precision quantization frameworks for LMs (MXPLMs). We first review quantization fundamentals, including uniform and non-uniform quantizers, quantization granularity, and methods widely used in post-training quantization. We then categorize and compare recent MXPLM frameworks according to their bit allocation strategies and precision configurations across weights, activations, and key-value caches. A comparative analysis highlights differences in perplexity, zero-shot task performance, and deployment trade-offs. Furthermore, we contrast MXPLMs with earlier mixed-precision quantization methods for deep neural networks, identifying strategies that transfer and those that face challenges in the LM setting. Finally, we summarize open issues and future directions, including hardware-aware design, activation quantization, and scalable optimization methods for billion-parameter models. By consolidating recent advances, this work serves as a reference for understanding the current landscape and research prospects of mixed-precision quantization for large-scale language models.
}
\end{abstract}

% keywords can be removed

\section{Introduction}

\begin{figure}[t!]
    \centering
    %\vspace{-0.5cm}
    \includegraphics[width=\textwidth]{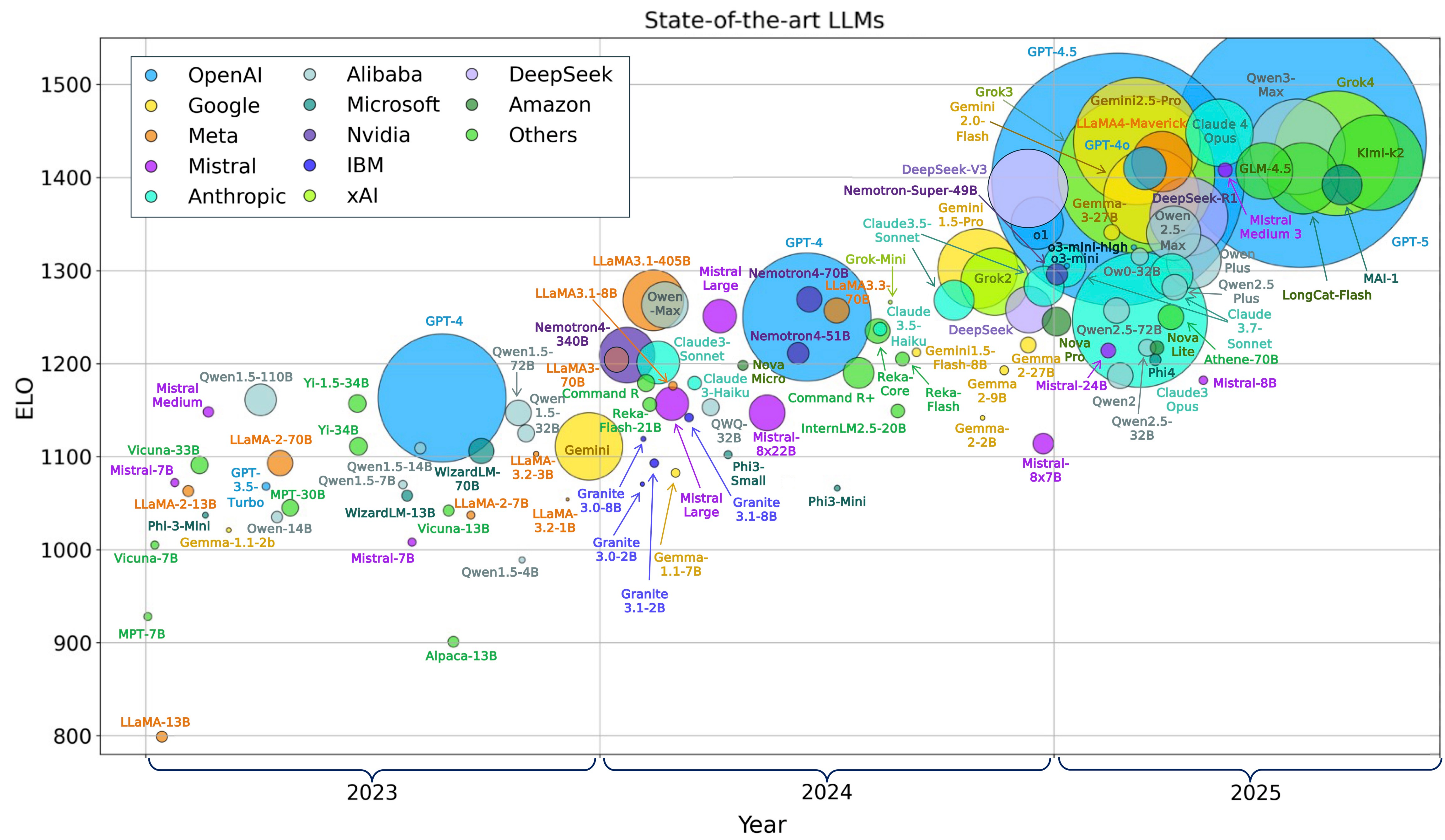}
    %\vspace{-0.6cm}
    \caption{ELO scores of state-of-the-art LMs released over the last three years. Scores are obtained from \cite{chiang2024chatbot}, where the circle size represents the total number of parameters in a model. For models whose sizes were not officially disclosed, the values shown reflect the average of multiple open-source estimates.}
    \label{fig:LLMs}
    %\vspace{-0.5cm}
\end{figure}

Over the past three years, the size of language models (LMs) has grown exponentially, placing increasing demands on modern computing infrastructure and driving the need for more advanced hardware (Fig. \ref{fig:LLMs}).
Model sizes have increased, outpacing the growth of hardware and memory capabilities.
For example, OpenAI’s GPT-3, released in 2020 with 175 billion (175B) parameters, requires an estimated 3.14×10$^{23}$ floating-point operations for training, equivalent to 355 GPU-years on an NVIDIA V100 cluster and roughly \$4.6 million in compute cost for a single run \cite{li2020gpt3overview}.
Deploying such models requires extensive cloud infrastructure and GPU resources due to their high compute and memory requirements.
In 16-bit precision, the 175B-parameter GPT-3 model requires roughly 350 GB just to store its weights, necessitating model-parallel inference across multiple GPUs \cite{li2020gpt3overview}.
Further scaling of these models will amplify both compute and memory demands.
Therefore, model compression techniques and efficient methods for quantizing and processing such models with lower bit precision are essential to sustainably support the growth of LMs.

Since the Transformer architecture and multi-head self-attention were introduced in 2017 \cite{vaswani2017attention}, language models have rapidly scaled in size and capability. This trend started with early models such as BERT \cite{devlin2019bert}, GPT \cite{radford2018improving}, and T5 \cite{raffel2020exploring}, released between 2018 and 2019. By 2020, models such as GPT-2 (1.5B parameters) \cite{radford2019language} and GPT-3 (175B parameters) \cite{brown2020language} demonstrated few-shot prompting and strong transfer learning capabilities. 
GPT-3 marked the rise of foundation models, massive pretrained models that serve as a general platform for a wide range of downstream tasks. Early examples included Google’s PaLM (540B parameters) \cite{chowdhery2023palm}, DeepMind’s Chinchilla \cite{hoffmann2022training}, and Microsoft and NVIDIA’s 530B-parameter Megatron-Turing NLG \cite{smith2022using}, released in 2022.
By 2023, models like GPT-4 \cite{achiam2023gpt} introduced multimodality, enabling input across text and images within a unified framework.
In general, multi-modal foundation models are AI systems designed to process and reason over multiple types of data (modalities), including text, images, audio, video, tables, and graphs, within a single unified architecture \cite{liang2024survey}. These include OpenAI’s GPT-4 \cite{openai2023gpt4}, Anthropic’s Claude 3 \cite{anthropic2024claude3}, Google’s Gemini \cite{pichai2023gemini}, among others. Further, the most recent models, like GPT-5 \cite{openai2025gpt5}, are more complex and designed to be more agentic. For example, GPT-5 has been explicitly optimized for agentic tasks, autonomously chaining multiple tool calls to address complex multi-step problems \cite{openai2025gpt5}. Similarly, Google’s Gemini 2 emphasizes the development of AI assistants that can interpret their environment, reason several steps ahead, and autonomously take actions on behalf of the user under supervision \cite{google2024gemini2}. Building on this trajectory, Anthropic’s Claude 4 offers advanced tool-use and extended reasoning capacities, supporting multi-step workflows across thousands of sequential steps and substantially expanding the capabilities of AI-driven agents
\cite{anthropic2025claude4}. 
The advanced reasoning abilities and problem-solving skills of such models represent an early step toward Artificial General Intelligence (AGI) \cite{yang2024harnessing, liang2024survey}.

As illustrated in Fig. \ref{fig:LLMs}, the size of Language Models (LMs) and the number of parameters have been increasing each year to achieve improvements in model capability. Fig. \ref{fig:LLMs} shows the state-of-the-art LMs that have appeared in the last few years and their ELO scores, a ranking metric that quantifies how likely one model is to outperform another in a head-to-head evaluation \cite{chiang2024chatbot}. While scaling up has brought significant gains in reasoning, language understanding, and generation, this growth trajectory raises concerns about sustainability due to increasing computational, memory, and energy demands. To address these challenges, model compression techniques, such as quantization, are essential for reducing model size, decreasing memory requirements, and improving inference efficiency, all while maintaining competitive performance.

Quantization is a key model compression technique that benefits both training and inference of LMs.
Many state-of-the-art LMs are memory-bound during inference \cite{qualcomm2024npu}, making model size reduction crucial for alleviating memory bottlenecks and enabling deployment on smaller or more cost-effective hardware.
For instance, deploying the 27B-parameter Gemma 3 model in BF16 (16-bit brain floating-point) format requires 54 GB of GPU memory, whereas INT4 (4-bit integer) quantization reduces this requirement to only 14.1 GB, which is nearly a 4x reduction \cite{team2025gemma, yvinec2025gemma3qat}. This reduction significantly decreases the number of GPUs needed for distributed inference.
Beyond memory savings, quantization also improves system throughput and latency. For example, the FP16 version of the LLaMA 3–8B model achieves 135.79 tokens/sec on the NVIDIA H100, while the INT8 and INT4 versions achieve 158.90 and 211.50 tokens/sec, respectively \cite{nvidia2025ptq}.

While low-bit quantization formats, such as INT8 and INT4, offer substantial memory and speed benefits, they can compromise accuracy, particularly when quantizing sensitive layers, such as attention projections or embedding matrices, in transformer architectures. To balance efficiency and performance, mixed-precision quantization is a promising approach for model compression, enabling the selective allocation of precision to preserve accuracy while reducing resource usage.

{
A growing body of related works addresses efficiency for LMs, with different emphases on quantization breadth versus mixed-precision depth. \cite{gong2024survey} presents a dedicated survey of \emph{low-bit} Large LMs (LLMs) spanning fundamentals, numeric formats, system/toolchain aspects, and algorithmic strategies for training and inference; mixed precision is covered as one instrument within the broader low-bit toolbox rather than the organizing principle. Complementary surveys organize \emph{efficient LLM inference} end-to-end, covering data-, model-, and system-level optimizations (e.g., KV cache, attention scaling, batching, scheduling), where quantization appears as one category among many \cite{zhou2024survey,wan2023efficient}. Broader compression overviews position quantization alongside pruning and distillation to map the overall landscape rather than concentrating on mixed-precision design \cite{zhu2024survey}. 

Several widely cited works are evaluation-centric rather than surveys; for instance, \cite{li2024evaluating} benchmark post-training quantization (PTQ) across weights, activations, and KV cache on multiple model families and task types, providing practical guidance but not a taxonomy of mixed-precision frameworks. There are also survey-adjacent syntheses focused specifically on quantization for LLMs, offering broad methodological coverage without centering mixed-precision allocation \cite{lang2024comprehensive}. Beyond LLM-specific surveys, \cite{rakka2024review} provides a cross-domain survey of \emph{mixed-precision neural networks} that categorizes frameworks by optimization strategies (e.g., reinforcement learning-based search) and quantization/rounding choices; we reference this work for general mixed-precision taxonomies and contrasts with earlier DNN literature, while our focus remains on LMs. Related community overviews for Transformer compression \cite{tang2024survey} and quantization in adjacent domains (e.g., ViTs and general DNNs) offer transferable taxonomies and hardware perspectives \cite{du2024model,yang2025survey}.

Our article is intentionally \emph{narrower and deeper} on \textbf{mixed precision} for LMs. First, it \emph{formalizes} mixed precision at both inter- and intra-layer levels and clarifies scope by excluding the common misnomer in which a-) only weights are quantized while activations remain floating point (e.g., W-int4/A-FP16) and b-) weights being uniformly quantized to one precision and activations being uniformly quantized to a different precision (e.g., W-int4/A-int8) from being labeled as "mixed precision". Second, it surveys \emph{MXPLM} frameworks with an explicit emphasis on \textbf{bit-allocation strategies} across weights, activations, and KV cache, and it provides a structured taxonomy (MPW; MPW{+}UPA; MPW{+}MPA) that distinguishes inter- and intra-layer allocation. Third, it offers a comparative discussion across frameworks (perplexity and zero-shot performance, where available) together with deployment considerations and hardware-awareness notes germane to billion-parameter models. These contributions complement prior low-bit and efficiency surveys by \emph{centering} mixed precision as the organizing axis and by analyzing how allocation policies (by layer, tensor, group) interact with LM-specific sensitivities and real-world deployment constraints.

To aid the reader, we first consolidate \emph{quantization fundamentals}: uniform vs. nonuniform, symmetric vs. asymmetric, and common granularities (per tensor, per channel, per group, per token) because these choices directly shape feasible mixed precision allocations and their accuracy/efficiency trade-offs in LMs. Within this background, we review families of methods that inform or enable mixed precision in practice, including second-order (OBS/GPTQ-style) approaches, equivalent transformations (e.g., AWQ/SmoothQuant), and rotation-based techniques (e.g., QuaRot), focusing on how these choices affect allocation granularity, sensitive layers, and activation/KV handling.

Compared to low-bit overviews \cite{gong2024survey} and general efficiency surveys \cite{zhou2024survey,wan2023efficient}, our contribution is to \textbf{center} mixed precision and to articulate the interactions between allocation policies and LM-specific sensitivities. Where broader compression surveys necessarily distribute attention across pruning and distillation \cite{zhu2024survey}, we provide a \emph{focused synthesis} of MXPLM frameworks and their evaluation patterns. And whereas evaluation papers emphasize empirical outcomes under uniform low-bit settings \cite{li2024evaluating}, our discussion integrates those insights into a \emph{framework-level} perspective on when and why mixed precision preserves accuracy while reducing memory and latency.

}

% {In this paper, we provide a comprehensive overview of recent mixed-precision quantization frameworks for language models. Our main contributions are summarized as follows:}
% \begin{itemize}
%     \item {We survey recent Mixed-Precision Language Model (MXPLM) frameworks, with a focus on their bit allocation optimization techniques and the strategies they employ to preserve accuracy under low-bit quantization.}
%     \item {We provide a comparative analysis across these frameworks in terms of perplexity, zero-shot task performance, and deployment.}
%     \item {We draw a comparison between MXPLM frameworks and earlier mixed-precision quantization frameworks for DNNs, highlighting which optimization strategies have carried over and why certain approaches have not yet been widely adopted for LMs.}
%     \item {We identify key challenges and outline future research directions, including hardware-aware MXPLMs, improved activation quantization, and scalable optimization methods tailored to billion-parameter models.}
    
% \end{itemize}

{The rest of the paper is organized as follows: Section~\ref{section:quantization_methods} introduces quantization methods and background. Section~\ref{sec:mixed_precisoion_lm_frameworks} presents recent MXPLM frameworks. Section~\ref{sec:insights&discussions} provides insights and discussions, including comparative evaluation across frameworks. Section~\ref{sec:quantization_compatible_hw} discusses quantization-compatible hardware, and Section~\ref{sec:future_directions} discusses prospects and future directions. Finally, Section~\ref{sec:conclusion} concludes the work.}

\section{Quantization methods}
\label{section:quantization_methods}
This section provides an overview of quantization fundamentals and discusses various types of quantizers. We categorize quantizers based on their properties: \textit{uniform} vs. \textit{non-uniform} and \textit{symmetric} vs. \textit{asymmetric}. We also explore quantization granularity (e.g., per-tensor, per-channel, per-group). Finally, we summarize and compare popular quantization techniques designed to improve the efficiency of LM deployment.

In practice, these quantization schemes are most commonly applied using the post-training quantization (PTQ) strategy \cite{frantar2022gptq, lin2024awq, park2024any, zeng2024abq, xiao2023smoothquant,zhang2025quantized}, where quantization is applied to a pre-trained model without requiring retraining. Its key advantage is the ability to scale efficiently to models with billions of parameters, where retraining is often impractical. Although alternative approaches, such as quantization-aware training (QAT) \cite{liu2023llm, chen2024efficientqat, cui2024cherry} and fully quantized training \cite{yu2024collage}, incorporate quantization during training to improve robustness under low-precision constraints, their significant computational and memory overheads limit their use in large language models. As a result, PTQ remains the most widely used method for compressing LMs, enabling efficient deployment across various platforms.

\subsection{Quantizers}
\label{sec:quantizers}
Quantizers map high-precision numerical values to lower-precision formats. In the context of LMs, integer quantization is one of the most widely adopted techniques, where high-precision floating-point values (e.g., FP32 or FP16) are converted into lower-precision integer formats such as INT8 or INT4. This transformation significantly reduces memory footprint and computational demands while preserving model accuracy. Recent studies have shown that INT4 and INT8 quantization can preserve over 99\% of full-precision accuracy on benchmarks, such as the Massive Multitask Language Understanding (MMLU) and the AI2 Reasoning Challenge (ARC) benchmarks, when combined with proper calibration and quantization-aware techniques \cite{gong2024survey, lang2024comprehensive}. For instance, methods, such as GPTQ \cite{frantar2022gptq} and QLoRA \cite{dettmers2023qlora}, report less than 1\% degradation in average accuracy when compressing LLaMA and GPT-style models to INT4. 

Furthermore, the storage requirement for models, such as GPT-3, decreases from 350 GB in FP16 to approximately 90 GB in INT4, enabling deployment on resource-constrained systems, including single GPUs or edge devices \cite{wang2024comprehensive}. Besides memory savings, integer operations are simpler and more hardware-efficient, enabling faster inference and lower power consumption on accelerators that natively support integer arithmetic. As such, integer quantization is a key enabler for scalable, efficient, and cost-effective deployment of large-scale models without compromising performance.

In the following sections, we describe the most common quantization schemes used in LMs, starting with uniform quantization, which includes both symmetric and asymmetric variants, followed by non-uniform quantization. These schemes are illustrated in Fig.~\ref{fig:quantization}.

\subsection{Uniform quantization}
\label{section:uniform_quantization}

Uniform quantization is the simplest and most widely adopted form of quantization in LMs due to its straightforward implementation and compatibility with hardware accelerators. It divides the numerical range into equal-sized intervals, resulting in uniformly spaced quantization levels. A key parameter in uniform quantization is the scale factor $S$, which defines the fixed step size between consecutive quantization levels. This scale enables the mapping of real-valued inputs to discrete integer levels and vice versa.

\subsubsection{Symmetric uniform quantization}
\label{section:sym_uniform_quant}
The simplest and most commonly used uniform quantization is symmetric quantization, where the quantized range is (roughly) centered around zero\footnote{Integer formats have an odd number of bin; therefore, it is common to assign an additional bin to the positive or negative range or drop one bin to make the grid fully symmetric around zero.}. In this scheme, the original floating-point values are mapped symmetrically around zero in the integer domain. In this case, the integer value corresponding to the real-valued zero, referred to as the zero point, is fixed at $0$. This eliminates the need for an offset, simplifying both the quantization and dequantization computations. Symmetric uniform quantization is characterized by two parameters: (1) a real-valued scaling factor $S$ that defines the step size of each quantization interval and (2) a bitwidth $b$, which determines the number of representable quantization levels. Given a real-valued input $x\in[\beta, \alpha]$, the quantization to a signed integer with $b$-bit precision is defined as:
\begin{align}
x_q &= \mathrm{clamp}\left( \left\lfloor \frac{x}{S} \right\rceil,\ -(2^{b-1}-1),\ 2^{b-1}-1 \right), \\
S &= \frac{\max{(|\alpha|,|\beta|)}}{2^{b-1}-1},
\label{eq:sym}
\end{align}

where $\lfloor \cdot \rceil$ denotes the round-to-nearest operator. The above choice for the scaling factor is based on a \textit{min-max} range \cite{nagel2021quantization_whitepaper} and ensures that the maximum absolute value is represented in the new grid. The clamping function ensures that the result remains within the representable range. It is defined as:
\begin{align}
\mathrm{clamp}(x; q_{\min}, q_{\max}) = 
\begin{cases}
q_{\min}, & \text{if } x < q_{\min},\\
x, & \text{if } q_{\min} \leq x \leq q_{\max}, \\
q_{\max}, & \text{if } x > q_{\max},
\end{cases}
\end{align}
where $q_{\min}$ and $q_{\max}$ correspond to the minimum and maximum integers representable by the chosen bitwidth $b$. For instance, in 8-bit signed symmetric quantization, $q_{\min}$ = -127 and $q_{\max}$ = 127. Fig. \ref{fig:quantization}(a) illustrates an example of 8-bit (INT8) symmetric uniform quantization, where the quantization grid is centered around zero, and zero in the FP domain aligns exactly with zero in the quantized INT domain. To approximate the original real-valued input $x$ from its quantized form $x_q$, a dequantization step is applied:
\begin{equation}
x \approx S \cdot x_q
\end{equation}

\subsubsection{Asymmetric uniform quantization}
\label{section:asym_uniform_quant}
While symmetric quantization is simple and effective for zero-centered data distributions, it becomes suboptimal when the data range is significantly skewed or not centered around zero. In such cases, many quantization levels may be wasted on underutilized regions, leading to reduced precision for the actual data. To address this limitation, asymmetric uniform quantization introduces an integer zero-point offset $Z$ that shifts the quantization grid, allowing it to better align with the input distribution. The asymmetric uniform quantization of a real-valued input $x\in[\beta, \alpha]$ using a \textit{min-max} range is defined in the following equations and illustrated in Fig.\ref{fig:quantization} for an INT8 example:
\begin{align}
x_q &= \mathrm{clamp}\left( \left\lfloor \frac{x}{S} \right\rceil + Z,\ 0,\ 2^b-1 \right), \\
S &= \frac{\alpha - \beta}{2^b - 1}; \quad Z = \left \lfloor \frac{-\beta}{S}\right  \rfloor
\label{eq:asym}
\end{align}

\subsection{Non-uniform quantization}
\label{section:non_uniform_quant}
Unlike uniform quantization, non-uniform quantization uses variable step sizes, allocating different levels of precision across the numerical range. It assigns finer precision (smaller intervals) to regions where the data is denser and coarser precision (larger intervals) to sparser regions. This flexibility can improve representation efficiency and quantization accuracy, especially for skewed data distributions.

The non-uniform quantization operation is formally defined as:
\begin{equation}
x_q = D_i, \quad \text{if} \quad x \in [\Delta_i,\ \Delta_{i+1})
\label{eq:non-uniform}
\end{equation}
where ${D_i}$ denotes the discrete quantization levels and ${\Delta_i}$ represents the corresponding quantization intervals. When the input value $x$ falls within the interval $[\Delta_i,\ \Delta_{i+1})$, it is mapped to the quantization level $D_i$. Unlike uniform quantization, neither the levels ${D_i}$ nor the intervals ${\Delta_i}$ are evenly spaced.

Non-uniform quantization has seen practical success in recent LM compression efforts. For instance, QLoRA \cite{dettmers2023qlora} introduces the NF4 (NormalFloat-4) quantization format; a data-aware, non-uniform quantizer designed to better capture the distribution of pre-trained model weights by allocating higher resolution near zero. This approach has been demonstrated to maintain downstream task performance while significantly reducing memory and computational requirements during fine-tuning. 

\begin{figure}[h]
    \centering
    \begin{subfigure}[b]{0.38\linewidth}
    \centering
    \includegraphics[width=\linewidth]{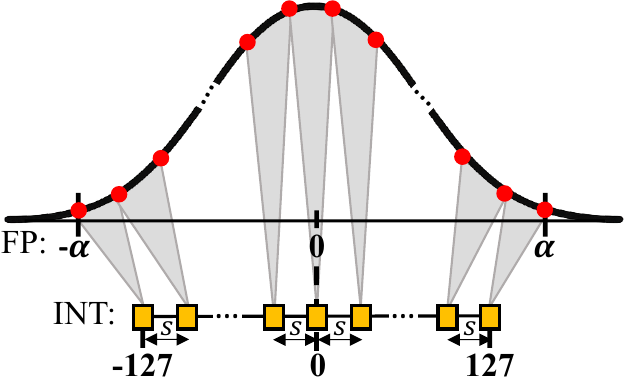}
    \caption{Signed symmetric uniform quantization}
    \end{subfigure}
    \hspace{0.05\linewidth}
    \begin{subfigure}[b]{0.38\linewidth}
    \centering
    \includegraphics[width=\linewidth]{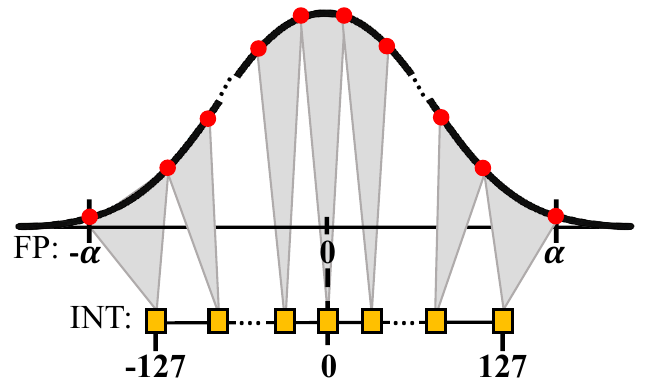}
    \caption{Signed symmetric non-uniform quantization}
\end{subfigure}
    
    \vspace{0.7mm}

\begin{subfigure}[b]{0.34\linewidth}
        \centering
        \includegraphics[width=\linewidth]{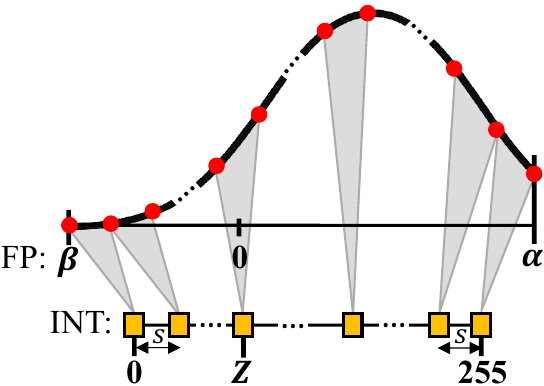}
        \caption{Asymmetric uniform quantization}
    \end{subfigure}
    \hspace{0.1\linewidth}
    \begin{subfigure}[b]{0.34\linewidth}
        \centering
        \includegraphics[width=\linewidth]{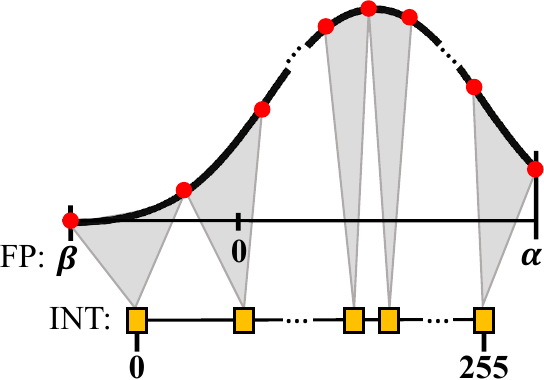}
        \caption{Asymmetric non-uniform quantization}
    \end{subfigure}

    \caption{Illustration of different quantization schemes using 8-bit representation.}
    \label{fig:quantization}
\end{figure}

As with uniform quantization, non-uniform schemes can be designed in symmetric or asymmetric forms, depending on whether the quantization grid is centered at zero or shifted to better match the data distribution, as illustrated in Fig. \ref{fig:quantization}.

\subsubsection{Quantization granularity}
\label{section:quantization_granularity}
Quantization in LMs can be applied at varying levels of granularity, each offering different trade-offs between computational efficiency and model accuracy. The granularity level determines how many elements share the same quantization parameters, which impacts both the precision of the quantized model and its hardware efficiency.

At the coarsest level, \textbf{per-tensor quantization} assigns a single scaling factor and zero-point to an entire tensor \cite{zhang2024flattenquant}. This method is computationally efficient and memory-friendly, making it promising for deployment on resource-constrained devices. Per-tensor quantization is more commonly used for weight tensors, which can be pre-processed offline to become quantizable. In contrast, it is very rarely used for activations, whose dynamic nature and higher variation across hidden dimensions and tokens can lead to significant accuracy degradation. An example of successful per-tensor \textit{static} quantization can be found in SmoothQuant \cite{xiao2023smoothquant}, where weights and activation of the OPT-175B and BLOOM-176B models are quantized to 8 bits with less than $1\%$ drop in average accuracy on zero-shot tasks.

A finer level of granularity is \textbf{per-channel quantization}, where each channel in a tensor, typically corresponding to rows or columns of a weight matrix, is assigned its own quantization parameters \cite{kim2023memory}. This approach strikes a balance between efficiency and accuracy by reducing quantization error compared to per-tensor quantization, while maintaining a moderate level of computational complexity. For these reasons, per-channel quantization is very commonly used for weight quantization with great success \cite{frantar2022gptq, lin2024awq, dettmers2023spqr, dettmers2022LLMint8}.

However, applying per-channel quantization to activations in LMs introduces practical challenges. Specifically, assigning different scaling factors to each channel complicates matrix multiplications with weight matrices, which typically assume uniform scaling across the input. To avoid this complexity and preserve the structure of matrix operations, a widely used alternative is \textbf{per-token quantization}, where each token in the input sequence is assigned a single scaling factor \cite{yao2022zeroquant}. While this approach is computationally convenient, it is suboptimal in terms of accuracy, as activation values in LMs vary more across channels than across tokens. To address this, recent works such as RPTQ \cite{yuan2023rptq} and MergeQuant \cite{wang2025mergequant} propose solutions like channel clustering and static per-channel calibration. Nonetheless, per-token quantization remains widely used due to its simplicity and compatibility with efficient hardware execution.

Further increasing granularity, \textbf{per-group quantization} divides each channel into smaller sub-groups, typically of fixed size (e.g., 128 elements), with each group assigned its own quantization parameters. Group quantization is particularly effective for low-bit quantization (e.g., 4-bit), offering improved accuracy by reducing intra-group variation without significantly increasing computational cost. Per-group quantization has become a standard technique in many recent LM quantization approaches \cite{heo2023rethinking, li2023llm, zheng2024mixllm, guan2024aptq}.

In practice, LMs often employ hybrid quantization strategies, where different model components use varying levels of granularity \cite{lee2023flexround, zhao2024atom, lee2025lrq}. For instance, weights are often quantized using per-channel quantization to maintain accuracy with minimal overhead, while activations may use per-token or per-group quantization to capture dynamic input variations better. This adaptive approach helps optimize both memory efficiency and model performance.

\subsection{Quantization methods for LMs}
\label{sec:quantization_methods}
\begin{figure}[t]
  \centering
\tikzset{
  root/.style={
    rounded corners, draw=black!60, fill=white, drop shadow,
    inner sep=5pt, minimum width=40mm, align=center
  },
  group/.style n args={1}{
    rounded corners=6pt, draw=#1!55, fill=#1!15, drop shadow,
    inner sep=6pt, minimum width=30mm, align=center
  },
  leaf/.style n args={1}{
    rounded corners=5pt, draw=#1!60, fill=#1!15, drop shadow,
    inner sep=4pt, minimum width=20mm, align=center
  },
  leaf/.default=black,
  edge/.style={line width=1pt, draw=black!65}
}

\begin{tikzpicture}[
  font=\sffamily,
  >=Stealth,
  node distance=9mm and 16mm
]

\colorlet{colPink}{red!60}
\definecolor{colGreen}{RGB}{183,159,220}   % pastel purple
\definecolor{colBlue}{RGB}{144,200,190}    % pastel teal
\definecolor{colOrange}{RGB}{205,170,125}   

\node[root] (root) {Quantization Methods};

\node[group=colPink,  below left=12mm and 25mm of root] (ET) {Equivalent Transformations};
\node[group=colGreen, below left=12mm and -15mm of root]               (OBS) {\vspace{1cm} OBS\vspace{1cm} };
\node[group=colBlue,  below right=12mm and -15mm of root](RL) {Rotated LMs};
\node[group=colOrange,below right=12mm and 20mm of root](AP) {AnyPrecision};

% ------------------- LEAVES -------------------
% ET (pink)
\node[leaf=colPink,  below=5mm of ET] (LMM) {\hyperref[LLM-MQ]{LLM-MQ}};
% \node[leaf=colPink, below=4mm of LMM] (ABQ) {\hyperref[methd:abq-llm]{ABQ-LLM}};

% OBS (green)
\node[leaf=colGreen, below left =5mm and -14mm of OBS] (ResQ1) {\hyperref[ResQ]{ResQ}};
\node[leaf=colGreen, below right =5mm and -14mm of OBS] (APTQ) {\hyperref[APTQ]{APTQ}};
\node[leaf=colGreen, below=4mm of APTQ] (Atom) {\hyperref[Atom]{Atom}};
\node[leaf=colGreen, below=4mm of ResQ1] (PB) {\hyperref[PB-LLM]{PB-LLM}};
\node[leaf=colGreen, below=4mm of PB] (OWQ) {\hyperref[OWQ]{OWQ}};
\node[leaf=colGreen, below=4mm of Atom] (SpQR) {\hyperref[SpQR]{SpQR}};
\node[leaf=colGreen, below=34mm of OBS] (MixLLM) {\hyperref[MixLLM]{MixLLM}};

% RL (blue)
\node[leaf=colBlue, below=5mm of RL] (ResQ2) {\hyperref[ResQ]{ResQ}};

% AP (orange)
\node[leaf=colOrange, below=5mm of AP] (PMDP) {\hyperref[PMDP]{PMDP}};

% ------------------- CONNECTIONS -------------------
\draw[edge,-{Stealth[length=3mm]}] (root) -- (ET);
\draw[edge,-{Stealth[length=3mm]}] (root) -- (OBS);
\draw[edge,-{Stealth[length=3mm]}] (root) -- (RL);
\draw[edge,-{Stealth[length=3mm]}] (root) -- (AP);

% ------------------- BACKGROUND -------------------
\begin{scope}[on background layer]
  \draw[rounded corners=8pt, fill=black!7, draw=black!20, line width=0.6pt]
    ($ (current bounding box.north west) + (-8pt, 8pt) $)
    rectangle
    ($ (current bounding box.south east) + ( 8pt,-8pt) $);
\end{scope}

\end{tikzpicture}
  \caption{Taxonomy of mixed-precision methods based on the quantization methods they employ.}
  \label{fig:quantization_methods_taxonomy}
\end{figure}
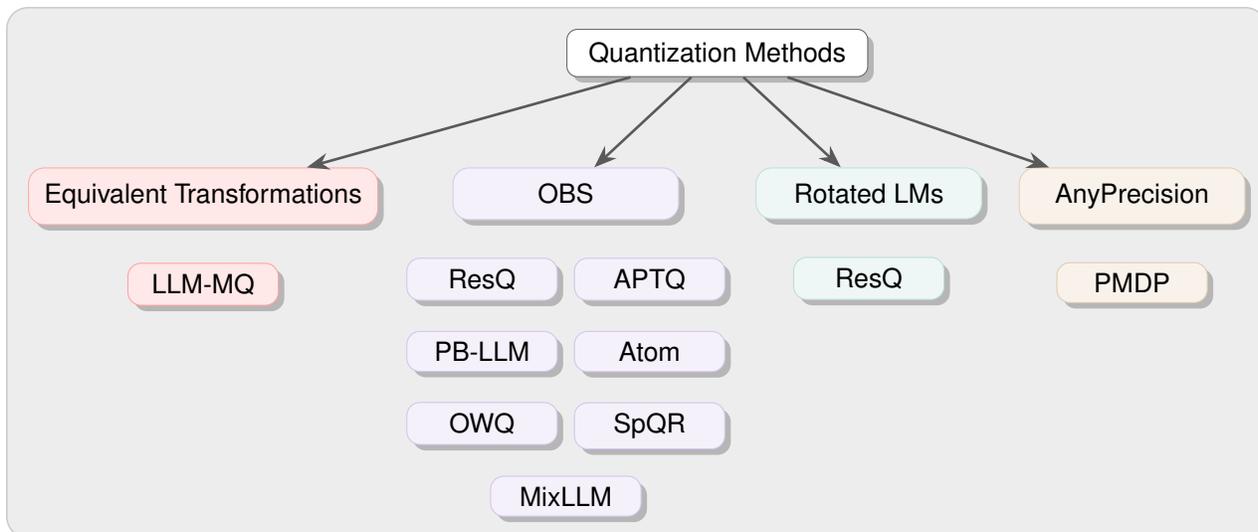

The substantial body of literature and methods that cover LM quantization confirms the effectiveness of quantization in accelerating LM inference. Despite the seemingly vast collection of methods and papers, most of them can be classified under a handful of families that share a common underlying principle. In this section, we highlight three primary quantization method families and discuss the most relevant and impactful papers related to each family.

\subsubsection{OBS-inspired methods}
\label{sec:OBS-inspired methods}
These methods trace their origin back to 1993 with the \textit{optimal brain surgeon} (OBS) \cite{hassibi1993obs}, which used second-order derivatives to prune a neural network. To determine which weights to prune, the authors calculate the resulting change in error induced by the removal of each weight (also known as \textit{saliency}). They then iteratively remove the weights with the smallest saliency and update the remaining weights to compensate for the increase in error that results from eliminating the weight.

Later in 2022, \textit{optimal brain compression} (OBC)\cite{fantar2022obq} extended the idea of OBS to quantization by jointly optimizing weight sparsity and low-bit precision using blockwise second-order approximations that scale to large models. Their formulation enables practical quantization and pruning of modern larger neural networks. However, it is the influential work of GPTQ\cite{frantar2022gptq} that solidified the OBS idea for LM quantization by tailoring the second-order approximation to the unique structure and scale of transformer models: (1) by recognizing that the exact order of quantization in large transformers did \textit{not} matter, they eliminate the search for least salient weights, and (2) the Cholesky decomposition of the Hessian allowed for a parallelized and iterative computation. We later provide a detailed description of the GPTQ algorithm.

Recent advancements have extended GPTQ's efficacy, particularly in achieving extremely low-bit quantization. Output-adaptive Calibration (AOC)\cite{edalati2025oac} directly addresses a limitation of GPTQ and similar methods by minimizing the distortion of the model's final output loss (e.g., cross-entropy) rather than just the layer-wise output reconstruction error. This "output-adaptive" calibration explicitly considers the global impact of quantization on the model's performance, leading to superior accuracy, especially at very low bit widths, such as 2-bit or binary, where traditional layer-wise methods often suffer significant degradation. 

\textbf{Generative pre-trained transformer quantization (GPTQ)}~\cite{frantar2022gptq} is a highly effective and widely adopted post-training quantization (PTQ) technique designed to compress large-scale transformer models to 3–4 bits per weight with minimal accuracy degradation. It builds upon the Optimal Brain Compression (OBC)\cite{fantar2022obq} framework by applying second-order approximations to minimize the output error induced by quantization, but introduces two key innovations that make it practical for LM-scale models: (1) a relaxation of the saliency-based quantization order, and (2) an efficient Cholesky-based approximation of the Hessian inverse.

GPTQ treats quantization as a reconstruction problem, where each weight matrix $\mat{W}$ is quantized to $\widehat{\mat{W}}$ so as to minimize the squared output error with respect to a calibration set:
\begin{equation}
     \arg\min_{\widehat{\mat{W}}} \| \mat{W}\mat{X} -  \widehat{\mat{W}}\mat{X} \|^2_2
\end{equation}
where $\mat{X}$ denotes the input activations. GPTQ processes each linear layer \textit{independently} and partitions the weight matrix into blocks of $B$ consecutive columns. Within each block, weights are quantized iteratively in a fixed left-to-right order, an important simplification over previous OBS-inspired methods, which required searching for the least salient weight at each step. GPTQ empirically demonstrates that this simplification has a negligible impact on accuracy while drastically reducing computational cost.

To correct for the error introduced by quantizing each column, GPTQ uses second-order information based on a Hessian matrix $\mat{H} = \mat{X}\mat{X}^\top$ that captures local curvature. Instead of explicitly computing the inverse Hessian, which is computationally expensive, GPTQ approximates $\mat{H}^{-1}$ using its Cholesky decomposition:
\begin{equation}
    \mat{H}^{-1} \approx (\mat{L}\mat{L}^\top)^{-1} = \mat{L}^{-\top} \mat{L}^{-1}
\end{equation}
where $\mat{L}$ is a lower-triangular matrix. This enables efficient updates and stable numerical inversion during calibration.

For each column $j$ in a block, the quantized weights and quantization error are computed as:
\begin{subequations}
\begin{align}
    \mat{Q}_{:,j} &\leftarrow \text{quant}(\mat{W}_{:,j}) \\
    \mat{E}_{:,j-i} &\leftarrow \frac{\mat{W}_{:,j} - \mat{Q}_{:,j}}{[\mat{H}^{-1}]_{jj}}
\end{align}
\end{subequations}
The error is propagated to the remaining unquantized columns using second-order corrections:
\begin{equation}
    \mat{W}_{:,j:(i+B)} \leftarrow \mat{W}_{:,j:(i+B)} - \mat{E}_{:,j} \cdot \mat{H}^{-1}_{j,j:(i+B)}
\end{equation}
After all columns in a block are quantized, a final refinement is applied to subsequent blocks:
\begin{equation}
    \mat{W}_{:,(i+B):} \leftarrow \mat{W}_{:,(i+B):} - \mat{E}\cdot \mat{H}^{-1}_{i:(i+B),(i+B):}
\end{equation}
By combining these efficient updates with a fixed ordering and Cholesky-based Hessian inversion, GPTQ provides a practical and scalable solution for second-order PTQ of LMs.

\subsubsection{Equivalent transformations}
\label{sec:equivalent_transformations}
Another category of techniques improves quantization by applying mathematically \textit{equivalent transformations} (ET) to the model weights or activations before quantization, making their distributions more amenable to low-bit representation while preserving the original network outputs.
These methods employ simple reparameterizations, such as per-channel scaling or shifting, that do not change any layer’s output in higher precision but have the effect of spreading out or reducing troublesome outlier values. Activation-aware weight quantization (AWQ)~\cite{lin2024awq} is a popular method in this category that applies per-channel weight scaling guided by activation statistics. We describe the technique in more detail below.

Several methods have extended the equivalent transformation (ET) paradigm beyond AWQ by varying both the type of transformation and the way parameters are selected. SmoothQuant~\cite{xiao2023smoothquant} introduced the idea of balancing quantization difficulty between weights and activations using per-channel scaling factors that are predefined based on layer norms. Outlier Suppression+~\cite{wei2023outlier} builds on this by adding shifting operations alongside scaling, allowing it to align activation channels and suppress consistently large outliers. It applies grid search for scaling and uses predefined shift values. Finally, OmniQuant~\cite{shao2024omniquant} takes this further by treating ET parameters as learnable, optimizing them via gradient descent to directly minimize quantization loss. While more computationally expensive, OmniQuant enables fine-grained adaptation of both scaling and shifting operations, supporting both weight and activation quantization.

\textbf{Activation-aware weight quantization (AWQ)}~\cite{lin2024awq} is a weight-only post-training quantization method designed for LMs. The central observation is that a small number of weight channels (0.1\%–1\%) contribute disproportionately to large-magnitude activations. Quantizing these \textit{salient} weights naively leads to large output errors. AWQ addresses this by rescaling the weight and activation channels such that the overall product remains unchanged, while salient weights are protected via reduced dynamic range during quantization. Similar methods have been adopted in earlier literature to alleviate issues with per-tensor quantization of MobileNets~\cite{nagel2019datafreequantizationweightequalization}.

Concretely, for a given weight matrix $\mat{W}$, and input activation $\mat{X}$, AWQ introduces a set of per-channel scales $\vec{s} = (s_1,\dots,s_n)$ expressed a diagonal matrix $\mat{S} = \mathrm{diag}(\vec{s})$ to rewrite the layers output:
\begin{equation}
      \mat{W} \mat{X} = \l(\mat{W} \mat{S}\r) \l(\mat{S}^{-1} \mat{X}\r),
\end{equation}
ensuring the layer output remains unchanged under full-precision computation. AWQ then applies quantization to the rescaled weight matrix:
\begin{equation}
    \widehat{\mat{W}}= Q \l( \mat{W}  \mat{S}\r),
\end{equation}
The resulting quantized weights can be folded back into the model, and the inverse scaling into preceding normalization layers, allowing AWQ to introduce no additional runtime cost.

The optimal scaling factors $\vec{s}^*$ are determined via grid search on a small calibration dataset, solving the following optimization problem:
\begin{equation}
    \vec{s}^* = \arg\min_{\vec{s}} \| Q\l(\mat{W} \cdot \text{diag}\l(\vec{s}\r)\r) (\text{diag}\l(\vec{s}\r)^{-1} \cdot \mat{X}) - \mat{W}\mat{X} \|
\end{equation}
After optimal scaling is applied, the quantization error is significantly reduced for important weight channels, particularly in the presence of high-variance activations.  In practice, AWQ achieves competitive 4-bit weight-only quantization accuracy with full-precision activations and has been shown to outperform or match methods like GPTQ in some settings, particularly where speed and implementation simplicity are prioritized.

\subsubsection{Rotated LMs}
\label{sec:rotated_llms}
A third line of attack for quantizing LMs involves removing heavy-tailed outliers by making the weight and activation distributions more \textit{coherent}~\cite{chee2024quip2bitquantizationlarge} or uniform through orthogonal transformations. These methods apply random or learned orthonormal \textit{rotations} to either the weight matrices or the activations, so that troublesome outliers are dispersed across dimensions. Intuitively, a rotation changes the basis of the vectors. For example, rotating a 2D vector $(1, 10)$ by 45° yields approximately $(7.8, 6.4)$, thereby eliminating the extreme value.

The core principle enabling rotation-based quantization is computational invariance. For a linear operation  $\mat{Y}=\mat{X}\mat{W}$, where $\mat{X}$ is the input activation and $\mat{W}$ is the weight matrix, an orthogonal matrix $\mat{R}$ (satisfying $\mat{R}^{\top}\mat{R}=\mat{I}$) can be introduced without changing the output:
\begin{align}
\mat{Y}=\mat{X}\mat{W} = \mat{X}\l(\mat{R}^{\top}\mat{R}\r)\mat{W} = \l(\mat{X} \mat{R}^{\top}\r) \l(\mat{R}\mat{W}\r),
\end{align}
where $\mat{X}'=\mat{X} \mat{R}^{\top}$ and $\mat{W}'=\mat{R}\mat{W}$ represent the rotated input  weight matrix respectively. The weights can be readily rotated and then quantized \textit{offline}, but for invariance to hold, the inputs must also be transformed accordingly during inference in an \textit{online} manner. Randomized Hadamard transformations have been employed recently due to their efficient hardware implementation, leading to impressive low-bit quantized performance. QuIP\#~\cite{tseng2024quipbetterllmquantization} first applied randomized Hadamard transformation along with vector quantization to achieve sub-4-bit weight-only quantization of Llama2 models. QuaRot~\cite{Ashkboos2025QuaRot} extends by applying \textit{online} Hadamard transformations to the attention module to remove outlier features in keys and values, enabling low-bit quantization of activations \& KV cache (see section below for more details).

QuaRot and QuIP\# are only part of a growing family of rotation-based quantization methods that aim to remove statistical outliers and improve numerical conditioning before quantization. SpinQuant~\cite{liu2025SpinQuant} goes a step further using backpropagation to learn the optimal rotation matrices jointly with quantization parameters. SpinQuant can better tailor transformations to model-specific distributions, albeit at a higher calibration cost. DuQuant~\cite{Lin20204DuQuant} extends the rotation approach by using dual transformations: a combination of rotations and permutations. DuQuant first applies a “zigzag” permutation, reordering channels to even out the activation outliers across rotation blocks, followed by a minor rotation to \textit{smooth} the activation landscape. These methods highlight a rich design space where orthogonal transformations – whether fixed, random, or learned – serve as a preprocessing tool to make quantization both more robust and more aggressive, opening the door to fully 4-bit quantized models across weights, activations, and KV cache.

\textbf{Outlier-free 4-Bit inference in rotated LMs (QuaRot)}~\cite{Ashkboos2025QuaRot} builds on the principle of output-invariant orthogonal transformations by strategically inserting randomized Hadamard matrices at multiple locations within each transformer block. These transformations aim to uniformly redistribute activation energy and suppress large outliers that typically impede low-bit quantization. Unlike earlier methods that focused solely on weight transformations, QuaRot introduces a mix of \textit{offline} (fused) and \textit{online} (runtime) transformations across the attention and feed-forward pathways, each targeting a distinct quantization bottleneck.

Specifically, QuaRot introduces a global rotation matrix \( \mat{Q} \) to the residual stream between transformer blocks, which is folded into the pre- and post-projection weights of the FFN. Within the FFN itself, a randomized Hadamard transformation $\mat{H} $ is applied online before the down-projection and absorbed post-hoc into the weight matrix. In the attention path, two uses of $ \mat{H}_{dh} $ target value projections and KV cache activations: it is first fused into the value and output projection matrices using Kronecker structure, and then applied online to queries and keys \textit{after} RoPE, which prevents folding due to positional encoding interference. This online transformation is later inverted to preserve attention semantics. These combined mechanisms enable QuaRot to reduce quantization error across weights, activations, and KV buffers, allowing for end-to-end 4-bit quantization with minimal accuracy loss on LLaMA and OPT models.

% \Marios{fix the figure}
% \begin{figure}[t]
%     \centering
%     \includegraphics[width=0.75\linewidth]{Figs/image.png}
%     \caption{QuaRot orthogonal transformations applied to multi-head self-attention.}
%     \label{fig:quarot-attention}
% \end{figure}
% \begin{table}[h]
% \centering
% \caption{QuaRot transformations: notation, location, and implementation details.}
% \label{tab:quarot_rotations}
% \resizebox{\textwidth}{!}{%
% \begin{tabular}{|l|c|c|l|}
% \hline
% \textbf{Trasnformation} & \textbf{Pre/Post} & \textbf{Online?} & \textbf{Notes} \\
% \hline
% Global rotation \( \mat{Q} \) before FFN         & Pre  & No  & Folded into FFN input/output weights \\
% Hadamard \( \mat{H} \) in FFN block               & Post & Partially & Applied online then fused into down-projection \\
% Kronecker-fused \( \mat{H}_{dh} \) into \( \mat{W}_v, \mat{W}_o \) & Pre  & No  & Offline fusion into value/output projections \\
% Online \( \mat{H}_{dh} \) on Q/K after RoPE       & Pre  & Yes & Cannot be folded due to RoPE; inverted post-attention \\
% \hline
% \end{tabular}%
% }
% \end{table}

By combining rotation-aware preprocessing with efficient runtime rotations, QuaRot enables uniform 4-bit quantization across weights, activations, and KV cache buffers with virtually no accuracy drop in large-scale LMs, such as Llama-2 and OPT.  Its design showcases how orthogonal transformations, when inserted with careful awareness of transformer structure, can yield significant reductions in quantization error without compromising compatibility or inference speed.

\subsubsection{Other quantization methods} 
\label{sec:Other quantization methods}
Not all methods neatly fall within the three families of methods described above, and some recent work utilizes a combination of these principles to achieve the best quantized accuracy \cite{lin2024qserve}. Any-Precision LM \cite{park2024any} is another PTQ framework that addresses the need for different-sized LMs without incurring the high memory and computational costs associated with storing separate models. This is particularly useful when the same LM has to be served on devices with heterogeneous hardware or latency requirements.

\myparagraph{Any-precision LLM: low-cost deployment of multiple, different-sized LLMs}
\label{Any-Precision LLM}
% \nonuniformlayertag \quad \asymmetriclayertag \quad \perchannellayertag

Unlike conventional quantization techniques that produce a fixed-precision model, Any-Precision LM extends the concept of Any-Precision DNNs to LMs, enabling a single model to dynamically support multiple bitwidths. The core idea behind Any-Precision LM is to store an $n$-bit quantized model in a way that enables the derivation of lower-bit models ($(n-1)$-bit, $(n-2)$-bit, ...) by selecting only the most significant bits (MSBs). This approach allows for adaptive precision selection at runtime, optimizing the trade-off between model quality and inference latency. The framework employs incremental upscaling, starting from a low-bit seed model and progressively refining it to higher bitwidths, along with a bitplane-based memory representation. In this representation, each bit position of the quantized weights is stored separately, enabling efficient access to different precision levels. Mathematically, the Any-Precision Quantization process can be expressed as follows. Given a weight matrix $W$, the quantized model at bitwidth $b$ is obtained as:
\begin{equation}
    Q_b(W) = \text{Truncate}(Q_n(W), b)
\end{equation}
where $Q_n(W)$ represents the highest-precision quantized model (parent model), and $\text{Truncate}(\cdot, b)$ extracts the most significant $b$ bits from each quantized weight.
To minimize quantization error, Any-Precision LM employs a non-uniform quantization strategy based on clustering (e.g., SqLLM\cite{kim2023squeezellm}), which divides weight centroids into sub-clusters during upscaling, preserving weight distributions across bitwidths.

\subsection{Discussion}
\label{section:quant_methods_comparison}

To better understand the relative strengths and trade-offs between quantization methods, we compare the three main families, OBS-inspired methods, equivalent transformations (ET), and rotated LMs, along three critical dimensions: \textit{computational effort}, \textit{quantization scope}, and \textit{deployment complexity}.

\paragraph{Computational effort.} OBS-inspired methods are the most computationally demanding among the three families. Techniques like GPTQ and OAC require computing approximate second-order statistics, such as the Hessian of the layer outputs, to guide quantization. Despite the approximations and factorization introduced by the authors, e.g. Cholesky factorization or block-wise Hessian, these approaches still require significant calibration time and memory overhead. In contrast, equivalent transformation methods are much lighter: AWQ\cite{lin2024awq}, SmoothQuant\cite{xiao2023smoothquant}, and Outlier Suppression+\cite{wei2023outlier} rely only on statistics from a small calibration set (e.g., max activations or layer norms), and perform inexpensive operations such as per-channel scaling or shifting. Rotation-based methods fall somewhere in between. When using fixed, untrained orthogonal transforms, such as Hadamard matrices in QuaRot or QuIP, the preparation cost is negligible. However, methods such as SpinQuant, which learn rotations jointly with quantization parameters, require gradient-based optimization, resulting in higher offline cost akin to light fine-tuning.

\paragraph{Quantization scope.} OBS-based techniques, including GPTQ and OAC, are primarily focused on quantizing weights. They operate on individual linear layers and do not directly address the quantization of activations or the attention KV cache, which are typically kept in higher precision (e.g., FP16 or INT8). In contrast, ET methods offer greater flexibility. SmoothQuant\cite{xiao2023smoothquant} and OmniQuant\cite{shao2023omniquant} explicitly target both weights and activations by scaling one to make the other more quantizable, and can achieve full W8A8 or W4A8 quantization. However, these methods generally do not extend to KV cache quantization. Rotation-based methods offer the most comprehensive coverage. QuaRot\cite{Ashkboos2025QuaRot}, for instance, enables uniform 4-bit quantization of weights, activations, and KV cache by carefully applying orthogonal transformations to suppress outliers across all components. As such, they represent the most complete solution when extreme compression is needed across the entire model pipeline.

\paragraph{Deployment complexity.} From a deployment standpoint, ET and OBS-based methods are the most attractive: they modify weight magnitudes or normalization layers offline, and the quantized model requires no special operations during inference. In contrast, rotated LMs can introduce deployment challenges. While some rotations can be absorbed into weights offline, others, particularly those applied to activations or KV cache, must be executed online during inference. This adds runtime overhead and may require custom kernels to implement fast Hadamard transforms or channel permutations. Nevertheless, these costs are often modest and justifiable in settings where full low-bit quantization is essential. Some methods, such as DuQuant, attempt to strike a balance by limiting online operations to lightweight block-wise rotations or permutations that are hardware-friendly.

\section{Mixed-precision language model frameworks}
\label{sec:mixed_precisoion_lm_frameworks}

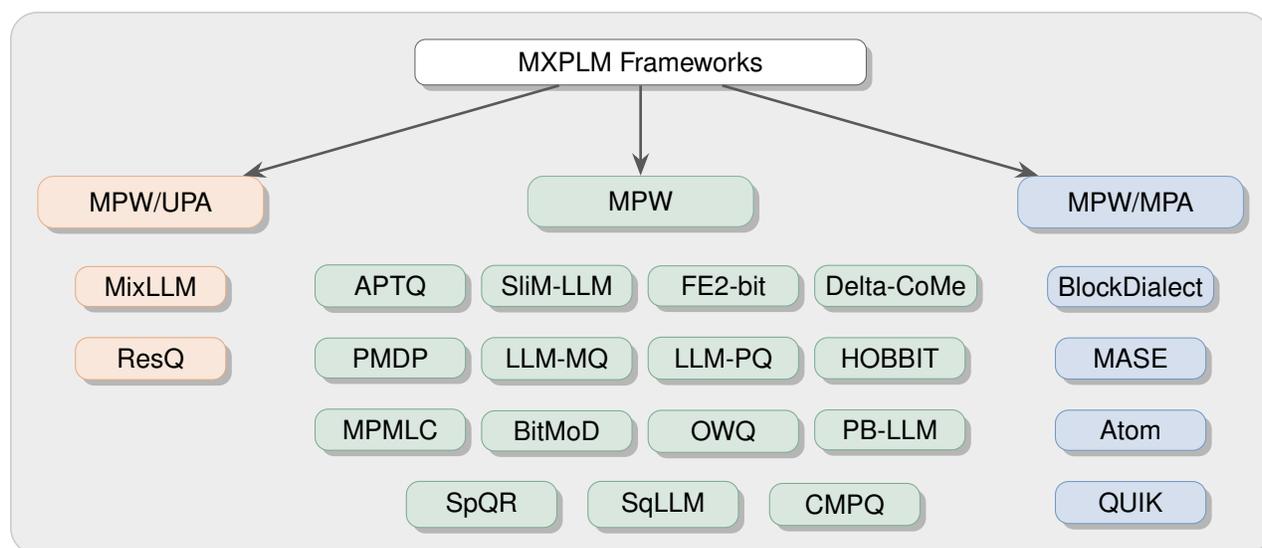
\begin{figure}[h]
  \centering
% \documentclass[tikz,border=6pt]{standalone}
% \usepackage[dvipsnames]{xcolor}
% \usetikzlibrary{arrows.meta,positioning,shadows,shapes.misc,fit,backgrounds,calc}

% \begin{document}
\begin{tikzpicture}[
  font=\sffamily,
  >=Stealth,
  node distance=9mm and 16mm,
  root/.style    ={rounded corners, draw=black!60, fill=white, drop shadow, inner sep=5pt, minimum width=60mm, align=center},
  group/.style   ={rounded corners=6pt, draw=#1!55, fill=#1!15, drop shadow, inner sep=6pt, minimum width=30mm, align=center},
  leaf/.style n args={1}{
  rounded corners=5pt,
  draw=#1!60,
  fill=#1!15,
  drop shadow,
  inner sep=4pt,
  minimum width=20mm,
  align=center
},
leaf/.default=black,
  edge/.style    ={line width=1pt, draw=black!65}
]
% --- Palette (tweak as you like)
\colorlet{colOrange}{Orange!85!black}
\colorlet{colGreen}{ForestGreen!70!black}
\colorlet{colBlue}{RoyalBlue!80!black}

% ---------- TOP ----------
\node[root] (root) {MXPLM Frameworks};

% Groups
\node[group=colOrange, below left=12mm and 20mm of root] (Gleft) {MPW/UPA};
\node[group=colGreen,  below=12mm of root]               (Gmid)  {MPW};
\node[group=colBlue,   below right=12mm and 20mm of root](Gright) {MPW/MPA};

% Left group leaves
\node[leaf=colOrange, below=5mm of Gleft] (MixLLM) {\hyperref[MixLLM]{MixLLM}};
\node[leaf=colOrange, below=4mm of MixLLM] (ResQ)  {\hyperref[ResQ]{ResQ}};

% Middle group leaves (2x2 grid)
\node[leaf=colGreen, below left =5mm and -14mm of Gmid] (Slim)   {\hyperref[SliM-LLM]{SliM-LLM}};
\node[leaf=colGreen, right=2mm of Slim]                (FE2)    {\hyperref[FE2-bit]{FE2-bit}};
\node[leaf=colGreen, below=4mm of Slim]                 (LLMMQ)  {\hyperref[LLM-MQ]{LLM-MQ}};
\node[leaf=colGreen, below=4mm of FE2]                  (LLMPQ)  {\hyperref[LLM-PQ]{LLM-PQ}};

\node[leaf=colGreen, left =2mm of Slim] (APTQ)   {\hyperref[APTQ]{APTQ}};
\node[leaf=colGreen, below=4mm of APTQ] (PMDP) {\hyperref[PMDP]{PMDP}};

\node[leaf=colGreen, right=2mm of FE2] (Delta) {\hyperref[Delta-CoMe]{Delta-CoMe}};
\node[leaf=colGreen, right=2mm of LLMPQ] (HOBBIT) {\hyperref[HOBBIT]{HOBBIT}};

\node[leaf=colGreen, below=4mm of PMDP] (MPMLC) {\hyperref[MPMLC]{MPMLC}};
\node[leaf=colGreen, below=4mm of LLMMQ] (BitMod) {\hyperref[BitMoD]{BitMoD}};
\node[leaf=colGreen, below=4mm of LLMPQ] (OWQ) {\hyperref[OWQ]{OWQ}};
\node[leaf=colGreen, below=4mm of HOBBIT] (PBLLM) {\hyperref[PB-LLM]{PB-LLM}};

\node[leaf=colGreen, below right =4mm and -8mm of MPMLC] (SpQR) {\hyperref[SpQR]{SpQR}};
\node[leaf=colGreen, right=4mm of SpQR] (SqLLM) {\hyperref[SqLLM]{SqLLM}};
\node[leaf=colGreen, right=4mm of SqLLM] (CMPQ) {\hyperref[CMPQ]{CMPQ}};

%right group
\node[leaf=colBlue, below =5mm of Gright] (Block) {\hyperref[BlockDialect]{BlockDialect}};
\node[leaf=colBlue, below= 4mm of Block]                  (MASE)  {\hyperref[MASE]{MASE}};
\node[leaf=colBlue, below= 4mm of MASE]                   (Atom)  {\hyperref[Atom]{Atom}};
\node[leaf=colBlue, below=4mm of Atom]                    (QUIK)  {\hyperref[QUIK]{QUIK}};

% Connect root to groups
\draw[edge,-{Stealth[length=3mm]}] (root) -- (Gleft);
\draw[edge,-{Stealth[length=3mm]}] (root) -- (Gmid);
\draw[edge,-{Stealth[length=3mm]}] (root) -- (Gright);

\path let \p1=(ResQ), \p2=(LLMPQ), \p3=(QUIK) in
      coordinate (baseline) at ($(0, {min(\y1,\y2,\y3)-1.2cm})$);

% ---------- GLOBAL BACKGROUND BOX ----------
\begin{scope}[on background layer]
  \draw[rounded corners=10pt, fill=black!7, draw=black!20, line width=0.6pt]
    ($ (current bounding box.north west) + (-10pt, 10pt) $)
    rectangle
    ($ (current bounding box.south east) + ( 10pt,-10pt) $);
\end{scope}

\end{tikzpicture}
% \end{document}
  \caption{Overview of mixed-precision LM (MXPLM) frameworks; UPW/UPA: uniform-precision weights/activations, MPW/MPA: mixed-precision weights/activations.}
  \label{fig:mxplm-frameworks}
\end{figure}

In this section, we summarize different recent mixed-precision language model (MXPLM) frameworks in the literature. We begin by clarifying the terminology of mixed-precision. In general, mixed-precision refers to the practice of allocating different bitwidths to various numerical elements, such as weights, activations, and key-value (KV) caches, either across layers or within a single tensor. In this work, we focus on mixed-precision across layers (\textit{inter-layer}) and across tensors of the same layer (\textit{intra-layer}).

To establish clarity, we define a "layer" in the context of LMs as a single transformer block, which typically includes a multi-head attention mechanism (projection, attention, linear) and a feed-forward network. Suppose a single weight tensor within such a transformer layer, for instance, the projection weights for keys ($Projection_K$), is partitioned into groups, and each group is quantized to a different precision. In that case, this constitutes \textit{intra-layer} mixed-precision. Since this approach inherently varies precision within a layer, it also naturally subsumes the \textit{inter-layer} case, where different transformer layers have different uniform precisions.

{Although it is common in the literature to refer to models with uniformly but differently quantized weights and activations (e.g., W-INT4/A-INT8) or uniformly quantized weights only (e.g., W-INT4/A-FP16) as “mixed-precision,” we do not adopt this convention.}. Our focus is on mixing precisions amongst weights within and across layers of the model. As such, we define three main categories of the MXPLM framework below:

\begin{enumerate}
    %\item \textit{Uniform-precision} weights (UPW) where all the weights in the model are quantized to the same precision, while activations are left in full-precision, e.g. Weights: int8, Activations: FP16.
    
    % \item \textit{Uniform-precision} weights \& \textit{uniform-precision} activations (UPW, UPA): where weights are quantized to a precision $p_1$ and activations are quantized to a precision $p_2$, but $p_1\neq p_2$, e.g., W-int4/A-int8.
    
    \item\textit{Mixed-precision} weights (MPW): where the weights in the model are quantized using mixed-precision while activations are left in full-precision, e.g., W-int\{4,8\}/A-FP16.
    
    \item\textit{Mixed-precision} weights \& \textit{uniform-precision} activations (MPW, UPA): where weights and activations undergo mixed and uniform-precision quantization, respectively. e.g,  W-int\{4,8\}/A-int8.
    
    \item\textit{Mixed-precision} weights \&  \textit{mixed-precision} activations (MPW, MPA): where weights and activations are quantized with mixed-precision, e.g. W-int\{4,8\}/A-\{int8, BF16\}.
\end{enumerate}

Fig. \ref{fig:mxplm-frameworks} shows the mixed-precision LM frameworks based on these categories. In the rest of this section, we elaborate on the formulation of the mixed-precision bit allocation in the different SOTA MXPLM frameworks shown in Fig. \ref{fig:mxplm-frameworks}. {We note that for our surveyed MXPLM works below, all frameworks are intra- and hence also inter-layer unless specified otherwise in their corresponding text.}

\subsection{Mixed-precision weights frameworks}
Here, we start with an overview of the frameworks with mixed precision quantization for weight parameters. 

\label{section:mpw_franeworks}
\subsubsection{SliM-LLM: salience-driven mixed-precision quantization for LLMs}
\label{SliM-LLM}

Huang et al.~\cite{huang2024slim} propose SliM-LLM, a mixed-precision framework that can be integrated into backbone PTQ methods such as GPTQ~\cite{frantar2022gptq} and OmniQuant~\cite{shao2023omniquant} (with the latter yielding $SliM$-$LLM^+$). The key idea is to quantize LMs using structured mixed-precision. SliM-LLM approximates the Hessian of each weight matrix from a small calibration set, and leverages this to compute weight saliency. Columns are grouped in blocks of 128, after which two steps are applied: (1) \textit{salience-determined bit allocation} (SBA) and (2) \textit{salience-weighted quantizer calibration} (SQC).

In SBA, each group is ranked by average salience and assigned a higher or lower bitwidth, while maintaining an overall average bitwidth of 2 or 3. Salient groups thus receive more bits, whereas less important groups are compressed more aggressively. The allocation is determined via a double-pointer search that solves:
\begin{equation}
  \arg\min_{g_1,\ldots,g_n}
D_{KL}\left(x w_f^\top \,\middle\|\, x \, Q\left(w_f \,\middle|\, [g_1,\ldots,g_n]\right)^\top\right)
\text{ subject to } \lvert G_{N-1} \rvert = \lvert G_{N+1} \rvert,
\end{equation}
where $g_i$ is the bitwidth of group~$i$, $Q(\cdot)$ denotes group-wise mixed-precision quantization, and $D_{KL}$ is the KL divergence between original and quantized outputs. The constraint $\lvert G_{N-1} \rvert = \lvert G_{N+1} \rvert$ ensures that the average bitwidth equals $N$ (typically 2 or 3). In $SliM$-$LLM^+$, SBA is integrated into OmniQuant with its learnable quantizer.

Even within a group, certain weights may remain disproportionately salient. To address this, SQC amplifies such “local outliers” by introducing a calibration parameter $\gamma$ into the quantization interval:
\begin{equation}  
\Delta = \gamma \cdot \frac{w_{\max} - w_{\min}}{2^N - 1}, 
\quad 
z = -\left\lfloor \frac{\gamma w_{\min}}{\Delta} \right\rceil.
\end{equation}
This expansion of the quantizer’s solution space is optimized by minimizing the saliency-weighted quantization loss:
\begin{equation}
  \arg\min_{\gamma} \; \mathcal{L}_s\big(w_s, \text{dequant}(\hat{w}_{sq}, \Delta, z)\big) 
+ \mathcal{L}_u\big(w_u, \text{dequant}(\hat{w}_{uq}, \Delta, z)\big),
\end{equation}
where $\mathcal{L}$ is the $\ell_2$ loss, $w_s$/$w_u$ are salient and less salient weights, and $\hat{w}_{sq}$/$\hat{w}_{uq}$ their quantized versions. Salient components are identified via a 3$\sigma$ rule on the saliency distribution. The search for $\gamma$ is performed over $[1-\lambda, 1+\lambda]$ with $2n$ candidates (empirically, $\lambda=0.1$, $n=50$). Importantly, SQC shares the same quantizer for $w_s$ and $w_u$, introducing no extra parameters or inference overhead.

By combining SBA’s global saliency allocation with SQC’s local refinement, SliM-LLM jointly enhances global and local quantization awareness of salient weights. Its group-wise (structured) design ensures hardware efficiency, avoiding scatter-gather codebooks or fine-grained bitmaps. When integrated into PTQ backbones like GPTQ~\cite{frantar2022gptq} or OmniQuant~\cite{shao2023omniquant}, it produces 2–3 bit LMs that maintain strong performance while running efficiently on GPUs.

\subsubsection{LLM-MQ: mixed-precision quantization for efficient LLM deployment}
%\interlayertag \quad \intralayertag
\label{LLM-MQ}

LLM-MQ~\cite{li2023llm} is a mixed-precision post-training quantization method for LMs, built on two main ideas: \textit{sparse outlier protection} and \textit{layer-wise precision allocation} using first-order (gradient-based) sensitivity. First, it detects outlier weights and keeps $0.5\%$ of them in FP16 (compressed sparse row, CSR, format) while quantizing the rest of the outliers into INT2. Second, LLM-MQ assigns the bitwidth per layer to either 2, 3, or 4 bits, given a certain target memory budget by leveraging a first-order Taylor approximation of the model’s output loss. Specifically, it measures how much the loss changes if the $i$th layer is quantized to $b$ bits, denoted as $s_{i,b}$. Then it solves an integer programming problem:
\begin{align}
& \arg\min_{\{c_{i,b}\}} \;\; \sum_{i=1}^{N} \sum_{b \in \{2,3,4\}} c_{i,b}\,s_{i,b}
\nonumber  \\
& \text{s.t.}
\quad
\sum_{b} c_{i,b} = 1,
\quad
\sum_{i=1}^{N} \sum_{b} c_{i,b}\,M\bigl(Q_b(W_i)\bigr)\;\le\;B,    
\end{align}
where $N$ is the LM's layer number, $c_{i,b}\in{0,1}$ is an indication of whether layer~$i$ uses bitwidth~$b$, $M(\cdot)$ is the memory usage of quantized weights, and $B$ is the target weight memory budget. Solving this yields per-layer bitwidths that minimize total quantization error while respecting the budget. Finally, LLM-MQ supplies GPU kernels that fuse dequantization with GEMV and organize 2/3/4-bit weights in efficient layouts, thereby achieving an end-to-end speedup over FP16 baselines.

\subsubsection{PMDP: progressive mixed-precision decoding for efficient LLM Inference}
\label{PMDP}

Chen et al.~\cite{chen2024progressive} introduce PMPD which combines two complementary strategies. The first is a \textit{phase-aware scheme}, where weight precision differs between the prefill phase and the decoding phase. The second is a progressive mixed-precision scheme, where the level of precision is adjusted across tokens in the sequence. In this setting, tokens that appear earlier are processed with higher precision, while tokens that appear later are more resilient to approximation and are, therefore, assigned lower precision. For example, the prefill phase may use 3-bit weights, and during decoding, the model gradually transitions from three 2-bit weights as the token position increases.  

These two strategies work together to maintain generation quality while improving efficiency. To support precision switching, PMPD introduces two kinds of runtime schedulers: one that is \textit{prompt-agnostic} and static, and another that is \textit{task-agnostic} and learned. Because precision changes both across phases and across tokens, the weights within a layer can be quantized to different precisions at different points in the inference process. 

The method operates in two stages. In the \textit{offline} stage, quantized variants of the model are generated, the phase-aware precision allocation is determined by selecting the lowest precisions that still meet the target quality, and the scheduler is configured. In the deployment stage, the scheduler assigns the chosen precisions to phases and triggers switching for each prompt. To avoid additional memory requirements when storing multiple weight precisions, PMPD relies on Any Precision LLM~\cite{park2024any}.  

The scheduler is designed to minimize the average bitwidth subject to a quality constraint. Considering decoding up to an output length $OL$, the search space includes switching times $\{st(p)\}$ for each precision $p$. The optimization problem is given as
\begin{align}
    \min{st(p)}, \quad &\forall p\in P\setminus\{p_{\min}\}
    \nonumber \\
    \text{s.t.}\quad &q_{\mathrm{ref}} - \epsilon \;\le\; q(S),  \nonumber \\
    &0 \;\le\; st(p) \;<\; OL,\quad p \in P,
    \nonumber \\
    &p \;>\; q \;\;\implies\;\; st(p) \;\le\; st(q) ,\quad p,q \in P
\end{align}
where $q(\cdot)$ measures the quality of the scheduler $S$, $P$ is the set of available precisions, $\epsilon$ is the tolerated quality drop, and $st(p)$ denotes the switching step for precision $p$.  

By progressively lowering precision from three to two bits across the decoding process, PMPD reduces memory bandwidth demands, which are a key bottleneck in autoregressive generation. This yields faster inference on resource-constrained devices while maintaining performance close to the original model.

\subsubsection{APTQ: attention-aware post-training mixed-precision quantization}
\label{APTQ}

APTQ~\cite{guan2024aptq} extends the second-order Hessian-based approach of GPTQ~\cite{frantar2022gptq} to the broader structure of transformer attention. Instead of treating each linear projection independently, APTQ formulates quantization at the level of the entire attention mechanism. This shift allows it to account for interactions between the $Q$, $K$, $V$, and $O$ matrices, capturing dependencies that are otherwise neglected.

Formally, APTQ defines an objective over the multi-head attention function $F$, seeking quantized weights $\widehat{W}$ that minimize the reconstruction error:
\begin{equation}
    \arg \min_{\widehat{W}} \| F(W,X) - F(\widehat{W},X)\|^2_2,
    \label{eq:aptq_objective}
\end{equation}
To solve this problem, APTQ employs a Levenberg–Marquardt approximation~\cite{lecun1989optimal} of the Hessian, keeping first-order derivative components while retaining sufficient curvature information for effective error correction. This design makes the optimization tractable while preserving the benefits of second-order updates.

For each projection in the attention block, APTQ derives both first- and second-order gradients with respect to quantized weights. Quantization error is then corrected by propagating updates to the remaining unquantized weights using the Hessian inverse:
\begin{equation} 
    E = -\frac{W_q - \text{quant}(W_q)}{\left[H^{-1}_{\widehat{W}}\right]_{qq}}, 
    \qquad 
    \delta_F = E \cdot \left(H^{-1}_{\widehat{W}}\right)_{:,q},
    \label{eq:aptq_update}
\end{equation}
where $E$ is the local quantization error for group $q$, $w_q$ are the original weights of group $q$, and $\delta_F$ denotes the propagated correction applied to the remaining unquantized weights of the layer . This formulation mirrors the reconstruction updates in GPTQ but adapts them to the multi-head attention setting.

Beyond improving attention quantization, APTQ also incorporates a Hessian-trace-based mixed-precision allocation. The trace of the Hessian provides a measure of sensitivity, and layers with larger values are assigned higher bitwidths. In practice, APTQ allocates 4 bits to the most sensitive projections, most notably the $K$ matrices, while compressing less sensitive components to 2 bits. Let $R$ denote the fraction of 4-bit weights across the model. The average precision is then given by:
\begin{equation}
    \text{Average bits} = 4R + 2(1-R),
    \label{eq:aptq_avg_bits}
\end{equation}
enabling a tunable trade-off between compression and accuracy.

This hybrid allocation scheme allows APTQ to push LMs into the 2–4 bit regime with limited accuracy loss. By integrating attention-aware objectives with Hessian-guided sensitivity analysis, APTQ demonstrates how structured mixed-precision quantization can reduce model size while maintaining performance, offering a practical path for deploying LMs under tight memory and latency constraints.

\subsubsection{MPMLC: harnessing DRAM and SSD for sustainable and accessible LLM inference with mixed-precision and multi-level caching}
\label{MPMLC}

MPMLC~\cite{peng2024harnessing} introduces \textit{M2Cache}, a co-design framework that combines mixed-precision quantization with multi-level caching to enable inference on commodity servers with limited GPU memory. The method utilizes a pre-trained predictor, Deja Vu~\cite{liu2023deja}, to dynamically estimate neuron activity. Important neurons are executed at higher precision (e.g., FP16), while less active ones are quantized more aggressively or offloaded to DRAM. This selective allocation conserves GPU HBM capacity without sacrificing accuracy on critical paths.

To support this dynamic scheme, MPMLC extends the memory hierarchy beyond HBM. A GPU-managed least recently used (LRU) cache reduces transfers between DRAM and GPU memory by keeping frequently accessed neurons in HBM. When both HBM and DRAM are insufficient, SSDs serve as a final cache tier, providing additional capacity at the cost of latency. By combining precision-aware execution with hierarchical caching, MPMLC enables the execution of large LMs on hardware that would otherwise be unable to host them.

Precision allocation is further refined through an uncertainty-guided search. The method evaluates how different ratios of low- and high-precision neurons affect decoding stability using the uncertainty score:
\begin{equation}
\mathrm{UQEst}(\mathrm{LM}, r_{\mathrm{low}}, r_{\mathrm{high}})
= -\sum_{i > j} \sum_{k} p^i_k \log(p^i_k),
\label{eq:mpmlc_uncertainty}
\end{equation}
where $j$ is the prompt length, $p^i_k$ is the probability assigned to token $k$ at generation step $i$, and $r_{\mathrm{low}}, r_{\mathrm{high}}$ are neuron ratios at different precisions. This iterative process identifies a balance between memory savings and output reliability. By retaining only the most essential neurons in higher precision and caching the rest, MPMLC provides a sustainable and accessible solution for LLM deployment on resource-limited GPUs.

\subsubsection{Delta-CoMe: training-free delta-compression with mixed precision}
\label{Delta-CoMe}

Delta-CoMe~\cite{ping2024delta} targets the efficient compression of \textit{delta weights}, i.e., the differences between a fine-tuned model and its base LLM. Instead of retraining, the method applies a quantization strategy guided by the statistical properties of these deltas. Singular Value Decomposition (SVD) reveals that delta weights follow a long-tailed spectrum: a few large singular values dominate, while the rest decay rapidly. Delta-CoMe exploits this structure to assign precision adaptively.

The framework allocates higher precision to singular vectors associated with large singular values, since these components carry most of the performance-critical information. Vectors linked to smaller singular values are quantized into lower-bit formats, and those corresponding to negligible singular values can be pruned entirely (0-bit). This creates a natural hierarchy of mixed-precision representations aligned with the importance of each component.

Formally, for a delta matrix, indices $r_{\text{begin}}$ and $r_{\text{end}}$ define the quantization range of singular vectors, and the precision assignment must satisfy:
\begin{equation}
k \times (r_{\text{end}} - r_{\text{begin}})(h_{\text{out}} + h_{\text{in}}) 
= 16 \times \alpha \times h_{\text{out}}h_{\text{in}},
\label{eq:delta_come_constraint}
\end{equation}
where $k$ is the bitwidth for a group of singular vectors, $h_{\text{out}}$ and $h_{\text{in}}$ are the output and input dimensions, and $\alpha$ is the target compression ratio. This constraint ensures that mixed-precision allocation meets the desired memory footprint.

Delta-CoMe evaluates three configurations: \textit{single precision} with all singular vectors at 3-bit, \textit{double precision}, a split between 3-bit and 8-bit representations, and \textit{triple precision}, a mix of {8,3,2}-bits, depending on singular value magnitude.  

Empirical results on mathematical reasoning tasks show that the triple-precision scheme achieves the best trade-off between accuracy and compression, outperforming uniform quantization baselines. By aligning quantization with the intrinsic structure of delta weights, Delta-CoMe provides a training-free yet effective strategy for compressing fine-tuned LLMs.

\subsubsection{FE2-bit: fast and efficient 2-bit LLM inference on GPU}
\label{FE2-bit}

Li et al.~\cite{li2023fast} propose FE2-bit, a framework that enables efficient 2-bit inference on GPUs. The method addresses three main challenges: (i) accuracy loss from uneven weight distributions when quantized uniformly, (ii) speed degradation caused by sparse outliers, and (iii) heavy computational overhead during GPU dequantization.

To mitigate these issues, FE2-bit adopts a group-wise mixed-precision scheme guided by sensitivity analysis. Each weight matrix is divided into groups of size $M$, and the sensitivity of group $i$ is computed as:
\begin{equation}
S_i = \sum_{m=1}^{M} \sum_{j=1}^{OC} \frac{w_{Mi+m,j}^2}{\left[H^{-1}\right]^2_{Mi+m}},
\label{eq:fe2bit_sensitivity}
\end{equation}
where $\mat{H}^{-1}$ is the offline-computed inverse Hessian, $w_{Mi+m,j}$ is the weight of element $m$ in group $i$ for output channel $j$, and $OC$ is the number of output channels. Groups with high sensitivity are quantized to 4 bits, while low-sensitivity groups are quantized to 2 bits. A small subset of very large weights is further preserved as 16-bit sparse outliers to reduce error without substantially increasing average bitwidth.

The framework refines this quantization through QAT, which reduces residual error. The same sensitivity analysis used for weights is then extended to input channels: channels with wider dynamic ranges are assigned 4-bit precision, while others remain at 2 bits. To support GPU deployment, FE2-bit incorporates memory alignment strategies to prevent accuracy degradation and minimize bitwidth overhead.

A key bottleneck in 2/4-bit inference is dequantization to 16-bit before matrix multiplication. FE2-bit introduces an asynchronous dequantization pipeline that utilizes shared memory to overlap the computation of group scales with the loading of quantized weights. This process is further accelerated by CUDA primitives that parallelize partial result reductions inside a warp. Together, these optimizations ensure that low-bit inference does not incur excessive runtime overhead.

Finally, FE2-bit carefully manages sparse outliers. Their ratio is capped at 0.5\%, and they are preferentially preserved in 2-bit groups, where their impact on accuracy is greatest. This balance of sensitivity-guided precision, GPU-aware optimizations, and controlled outlier retention allows FE2-bit to achieve accurate and efficient 2-bit inference for LLMs.

\subsubsection{LLM-PQ: serving LLMs on heterogeneous clusters with phase-aware partition and adaptive quantization}
\label{LLM-PQ}

LLM-PQ~\cite{zhao2024llm} enables inter-layer mixed-precision inference of LLMs on heterogeneous GPU clusters. Its goal is to reduce resource demands and computational costs by jointly optimizing precision assignment and model partitioning. The framework integrates two components: an \textit{offline assigner}, which determines layer partition, micro-batch sizing, and quantization bitwidths; and a \textit{distributed runtime}, which executes the plan to perform inference. The assigner relies on two complementary cost models. The first is an analytical memory model that predicts GPU memory occupation under different mixed-precision assignments. The second is a latency model that estimates the execution time of each partition during the two critical inference phases: \textit{prefill} and \textit{decode}. Latency prediction is based on a linear regression model calibrated for the target hardware. 

To measure the sensitivity of each layer to quantization, LLM-PQ introduces a \textit{variance indicator}, derived from an upper bound on output variance introduced by quantization. For layer $i$ and bitwidth $b$, the sensitivity is defined as:
\begin{equation}
w_{i,b}=\sum_{o \in O_i} D_{W_o}\left(S_{W_o}(b_i)\right)^2 G(X_o),
\label{eq:llmpq_variance}
\end{equation}
where $O_i$ is the set of linear operators in layer $i$, $W_o$ denotes the weights of operator $o$, $X_o$ is the input feature, and $G(\cdot)$ captures variance under deterministic or stochastic rounding. This metric ranks layers by their robustness to different quantization precisions and guides the allocation of bitwidths.

Using this indicator, LLM-PQ formulates the precision assignment and layer partition problem as an Integer Linear Programming (ILP) objective:
\begin{equation}
\min_{Z}\left(\left\lceil\frac{B}{\eta}-1\right\rceil T_{\text{max}}^{\text{pre}} + \left\lceil\frac{B}{\xi}-1\right\rceil (n-1)T_{\text{max}}^{\text{dec}} + T_{\text{pre}}+T_{\text{dec}}\right) + \theta \sum_{j=1}^{N}\sum_{i=1}^{L}\sum_{b \in \text{BITs}} z_{i,j,b} w_{i,b},
\label{eq:llmpq_ilp}
\end{equation}
where $z_{i,j,b}$ is a binary variable indicating if layer $i$ is placed on device $j$ at bitwidth $b$, $Z$ is the set of bitwidth assignments, $n$ is the number of generated tokens, and $B$, $\eta$, and $\xi$ denote the global batch size, prefill micro-batch size, and decode micro-batch size, respectively. The $T$ terms represent execution latencies in prefill and decode phases, while $\theta$ balances quality against acceleration. Constraints ensure that (i) each layer is assigned to exactly one device and precision, and (ii) per-device memory capacity is not exceeded.

To solve the ILP, LLM-PQ uses an off-the-shelf solver GUROBI~\cite{gurobi2023gurobi}. To address scalability challenges, LLM-PQ employs several optimizations: pruning by enumerating micro-batch sizes in the prefill phase, grouping layers to reduce solution space, and a \textit{bitwidth transfer heuristic} that exploits GPU performance asymmetries to swap precision assignments across devices while respecting memory constraints.

In addition, the authors propose an efficient plugin for on-the-fly quantized model loading, which reduces the required DRAM for model loading and improves recovery speed from potential failures. Particularly, they overlap the disk-to-CPU weight loading time with the on-GPU model quantization and the CPU-to-GPU memory copy, thereby determining the granularity of processed weights at runtime.

\subsubsection{HOBBIT: a mixed-precision expert offloading system for fast MoE inference}
\label{HOBBIT}

HOBBIT~\cite{tang2024hobbit} accelerates inference of MoE-based LLMs on memory-limited devices by combining mixed-precision expert loading, prefetching, and caching. The framework builds on sparse MoE layers~\cite{shazeer2017outrageously}, where feed-forward networks (FFNs) serve as experts, and introduces multiple precision versions: FP16 experts can be replaced by INT4, and INT8 experts by INT2. When a cache miss occurs, HOBBIT loads low-precision experts (INT4/INT2) in place of less important high-precision ones, thereby reducing loading latency without significant accuracy loss.

At the token level, HOBBIT employs a \textit{Dynamic Expert Loader} to decide which experts should be loaded in high or low precision. Expert importance is estimated using the magnitude of gating weights obtained during the Top-$k$ expert selection. For a token input $x$, the unimportance degree of expert $e_i$ is defined as:
\begin{equation}
s_{e_i}(x) =
\begin{cases}
0,& i=0, \\[6pt]
\displaystyle\sum_{j<i} \|G(x)_{e_j}\|, & i>0,
\end{cases}
\label{eq:hobbit_importance}
\end{equation}
where $\|G(x)_{e_j}\|$ is the gating weight magnitude of expert $e_j$. If $s_{e_i} > T_1$, expert $e_i$ is loaded in low precision, since its contribution is minimal. If $s_{e_i} \leq T_1$, it is loaded in high precision. A second threshold is also introduced to skip unimportant experts entirely. This mechanism allows HOBBIT to dynamically assign precision to experts based on runtime input during the prefill phase.

At the layer level, an \textit{Adaptive Expert Predictor} exploits the strong similarity of gating inputs across adjacent layers to prefetch future experts in mixed precision with minimal overhead. At the sequence level, a \textit{Multidimensional Cache Manager} combines four replacement policies: Least Recently Used (LRU), Least Frequently Used (LFU), Least High-Precision Frequently Used (LHU), and Farthest Layer Distance (FLD). Each expert $t$ is assigned a priority
\begin{equation}
p_t = w_{\text{lru}} p^{\text{lru}}_t + w_{\text{lfu}} p^{\text{lfu}}_t + w_{\text{lhu}} p^{\text{lhu}}_t + w_{\text{fld}} p^{\text{fld}}_t,
\label{eq:hobbit_priority}
\end{equation}
where the weights $w_{\cdot}$ are hyperparameters. The expert with the lowest priority is evicted on cache replacement.  

HOBBIT is implemented on Llama.cpp by redistributing model weights and computation patterns. All non-expert
weights and a fraction of multi-precision experts are stored in GPU memory, while full expert weights reside in CPU memory or SSD. This design, coupled with dynamic precision-aware loading, enables efficient MoE inference on
edge-class hardware. Evaluated on Mixtral and Phi‑MoE across Jetson Orin and RTX-4090 setups, HOBBIT’s $FP16+INT\{2,4,8\}$ mix cuts expert‑loading I/O by up to $4\times$ and yields $9.93\times$ decoding speed‑up ($3-4\times$ on desktop GPUs) and $60-80\%$ lower prefill latency, with a small accuracy loss.

\subsubsection{BitMoD: bit-serial Mixture-of-Datatype LLM Acceleration}
\label{BitMoD}
%\interlayertag \quad \intralayertag

BitMoD \cite{chen2024bitmod} introduces an innovative algorithm-hardware co-design method aimed at efficiently accelerating LMs via mixed-precision post-training per-group quantization of weights. The authors introduce a new fine-grained data type adaptation which utilizes a different numerical data type to
quantize a group of weights at 3-bit and 4-bit precision. Their new data type adaptation includes FP3 and FP4 extensions. For FP3, the extension allows the redundant zero, which is present due to the sign-magnitude representation that has both positive and negative zero, to be replaced by one of the four pre-defined special values. 

Four special values, divided into two sets, are chosen to balance encoding overhead and hardware complexity, and the special values must satisfy one of these two properties: 1) some values should fall inside the numerical range of FP3, ensuring they do not change its original absolute maximum, or 2) some values should fall outside the numerical range of FP3 to add asymmetry. The two values which are part of one set and which satisfy the first property (+3 and -3) are denoted as the new data type FP3-ER (ER for Extra Resolution). Since infinite values satisfy the second properties, the authors determine these two values which are part of one set in a way that minimizes the quantization error, and achieves a balanced asymmetry across all weight groups. By observing the quantization error of different values, BitMod adopts +6 and -6 as the special values that satisfy the second property, resulting in a new data type denoted as FP3-EA (EA for Extra Symmetry). Similarly +5 and -5 are added to FP4 to define FP4-ER, and +8 and -8 are added to define FP4-EA. As each weight group can be quantized with only one special value out of the possible four for FP3 or FP4 in addition to the basic values; i.e. those of FP3 and FP4, BitMoD relies on a fine-grained data type adaptation where each group is quantized using a different special value to minimize the Mean Squared Error (MSE): 
\begin{equation}
MSE=\frac{1}{N}\sum_{i=1}^N(W_{f,i}-W_{q,i})^2,
\end{equation}
where $W_{f,i}$ and $W_{q,i}$ are full-precision and quantized weight tensors, respectively. Specifically, the adaptation is based on a heuristic algorithm which eventually results in a mixed-precision quantization. 

After iterating through all special values and adding a special value to the set of basic values in every iteration, non-linear PTQ quantization is performed, and then the special value that results in the lowest MSE is assigned. It is worth noting that BitMod's algorithm can be vectorized on a GPU to simultaneously assign the best special value for all groups of a weight tensor, and that they build on top of a prior work (VS-Quant \cite{dai2021vs}) to quantize via second-level quantization the scaling factors to low-precision integers, hence reducing the dequantization cost. On the hardware side, BitMoD implements custom bit-serial processing elements capable of efficiently handling mixed-precision floating-point and INT weights, thus significantly enhancing computational efficiency and reducing hardware overhead. Please refer to the paper for more details on the microarchitecture of BitMoD's proposed custom hardware.

% BitMoD demonstrates significant performance improvements, achieving up to $1.69\times$ speedup compared to SOTA accelerators like ANT and $1.48\times$ over OliVe. Additionally, it provides better energy efficiency, demonstrating a $2.31\times$ improvement over the baseline FP16 accelerator, observed across representative large language models including OPT-1.3B, Phi-2B, Yi-6B, Llama-2 (7B, 13B), and Llama-3-8B.

\subsubsection{PB-LLM: partially Binarized Large Language Models}
%\interlayertag \quad \intralayertag
\label{PB-LLM}

PB-LLM \cite{shang2023pb} introduces a novel approach for quantizing LMs through partial binarization, enabling significant model compression while maintaining language reasoning capabilities. Under PTQ, PB-LLM, now denoted as PB-GPTQ, leverages a Hessian-guided reconstruction of the binarized weight matrix. Adapting concepts from GPTQ \cite{frantar2022gptq}, and for each column in the weight matrix, it iteratively binarizes the less significant un-salient weights, quantizes the salient weights at higher precision (e.g., INT4), and subsequently applies compensation to the remaining weights. Specifically, the optimization in PB-GPTQ is expressed through minimizing the layer-wise quantization error: 
\begin{equation}
\arg\min_{\widehat{W}} \|WX - \widehat{W}X\|_2^2, 
\end{equation}
where $\mat{W}$ is the original weight matrix that includes salient weights and un-salient (to-be-binarized) weights, $\widehat{W}$ is the quantized weight matrix, and $X$ represents inputs. To detect the salient weights using the Hessian criterion, PB-LLM calculates the saliency metric inspired by SparseGPT \cite{frantar2023sparsegpt} as follows:  $v_i=\frac{w_i^2}{[H^{-1}]^2_{ii}}$, where $H$ is the Hessian matrix of the layer-wise quantization error with respect to the weights. Accordingly, the un-salient weights are binarized while the salient weights are quantized to a higher precision, whereby the quantization for both cases is asymmetric per-channel. The iterative process repeats until all the weights are quantized. 

In the QAT framework that enhances the reasoning capacity of PB-LLM, PB-LLM freezes salient weights throughout training and optimizes scaling factors to mitigate the quantization error of quantized binary weights. For the first optimization, and at the beginning of QAT, PB-LLM filters out a number of weights from a pre-trained weight matrix and keeps them fixed throughout the training. For the second optimization, and inspired by AWQ \cite{lin2024awq}, PB-LLM analytically derives optimal scaling factors to minimize quantization error in residual binarized weights. The optimal values of column-wise scaling factor $\alpha$ are computed explicitly as: $\alpha^* = \frac{\|w_F\|_1}{n_{w_F}}$, after minimizing the L2 error. Here, $w_F$ is the vector of full-precision weights, and $n_{w_F}$ is its number of elements. This approach eliminates the need for empirical scaling factor searches, enhancing training efficiency.

% Experimental results on models such as OPT-1.3B and LLaMA-7B demonstrate that PB-LLM successfully retains model performance at low bitwidth quantization (as low as 1-bit with some higher-bit salient weights), achieving rapid convergence in QAT and significantly outperforming conventional binarization techniques.

\subsubsection{OWQ: outlier-aware weight quantization for ffficient fine-tuning and inference of LLMs}
\label{OWQ}
%\interlayertag \quad \intralayertag

The OWQ framework \cite{lee2024owq} introduces an outlier-aware (activation outliers) mixed-precision technique that combines INT3/4 and FP16 weight precisions to reduce memory and computational demands while minimizing quantization error effectively. The outlier-aware quantization has two steps: 1) OWQ first detects the weak columns (these are not quantized) by defining a sensitivity metric related to the layer-wise Hessian matrix, and 2) it quantizes the remaining weights to low-precision by relying on tuned quantization parameters. The $j-th$ weight column sensitivity metric is computed as follows: 
\begin{equation}
\text{sensitivity}_j = \lambda_j \|\Delta W_{:,j}\|_2^2,
\end{equation}
where $\lambda_j$ is the $j-th$ diagonal element of the Hessian matrix, and $\Delta W_{:,j}=W_{:,j}-\hat{W}_{:,j}$ represents the error for the $j-th$ output channel. OWQ can utilize this metric to select the top-k sensitive columns if the goal is to choose a specific number (k) of weak columns. OWQ then relies on OPTQ~\cite{frantaroptq} to quantize the rest of the weights in a sequential column-wise manner. OWQ modifies OPTQ by incorporating truncation with min-max quantization. After quantization, OWQ then stores the weak columns as FP16. To further enhance model robustness for certain tasks, OWQ utilizes Parameter-Efficient Fine-Tuning (PEFT) in conjunction with Weak Column Tuning (WCT). Specifically, WCT first quantizes the base model with OWQ and then fine-tunes only the weak columns that were kept in FP16 due to OWQ.

%Experimental evaluations across various LMs, including OPT and LLaMA, demonstrate that OWQ achieves significantly lower perplexity compared to existing methods at similar bit precisions, effectively matching or surpassing higher-bit methods such as GPTQ \cite{frantar2022gptq}. 

\subsubsection{SpQR: a sparse quantized representation for near-lossless LLM weight compression}
\label{SpQR}
%\interlayertag \quad \intralayertag

SpQR \cite{dettmers2023spqr} introduces a novel Sparse-Quantized Representation designed for near-lossless efficient LM compression. After identifying small groups of sensitive weights and individual outlier weights, SpQR keeps the outlier weights in high precision (16-bits), while compressing all other weights to 3/4 bits. To capture individual outliers, SpQR computes the sensitivity of each weight by measuring the layer-wise squared error between full-precision weights and quantized weights. The per-layer sensitivity metric guiding bit allocation is defined as: 
\begin{equation}
s_{ij} = \frac{(w_{ij} - Q(w_{ij}))^2}{2(XX^\top)^{-1}},
\end{equation}
where $s_{ij}$ is the sensitivity of weight $w_{ij}$, $X$ represents the calibration inputs, and ${(XX^\top)^{-1}}$ is the inverse Hessian matrix. These inputs are collected by running them through the model up to the particular layer. SpQR approximates this sensitivity metric by relying GPTQ \cite{frantar2022gptq}, which quantizes weight matrices column-by-column while adjusting the not-yet-quantized weights to compensate for the quantization error in each step. 

The framework isolates weights identified as outliers, which cause disproportionately large quantization errors, storing them in higher precision, while compressing the remaining majority of weights to 3/4 bits. Specifically, outliers detected via the sensitivity analysis are encoded using a Compressed-Sparse-Row (CSR) format \cite{hoefler2021sparsity} due to their scattered and unstructured nature. To capture small groups of sensitive weights, SpQR employs a bilevel quantization scheme with small groups (typically groups with 8-32 weights, where for each group of 8-32 consecutive weights, there is a separate quantization scale and zero-point), maintaining an efficient storage of quantization statistics by applying asymmetric min-max quantization to the quantization statistics themselves. Additionally, SpQR employs an optimized GPU-based decoding for SpQR format, leveraging the memory-bound nature of autoregressive inference on GPUs to hide decoding overheads with high compression
rates. Specifically, SpQR loads the quantized weights and group statistics into SRAM, dequantizes them to 16-bits, and performs matrix multiplication with 16-bit activations. SpQR also proposes a sparse matrix algorithm which handles outliers with load balancing, and runs faster than the cuSPARSE sparse algorithm in PyTorch \cite{paszke2019pytorch}.

% Extensive evaluations demonstrate that SpQR achieves near-lossless compression for models like LLaMA and Falcon, resulting in less than a $1\%$ relative accuracy drop in perplexity.

\subsubsection{SqLLM: dense-and-sparse quantization}
\label{SqLLM}
%\interlayertag \quad \intralayertag

What SqLLM \cite{kim2023squeezellm} is a PTQ framework that introduces two main innovations: 1) a sensitivity-based non-uniform quantization method using weighted K-means clustering guided by second-order information from the Hessian, and 2) a Dense-and-Sparse decomposition that extracts a small percentage of outlier and sensitive weights and stores them in FP16 using an efficient sparse format. SqLLM supports mixed-precision with weight representations composed of INT3/4 and FP16 for outliers, and activations retained in FP16. The optimization objective guiding mixed-precision bit allocation is based on sensitivity K-means clustering, and comprises minimizing the overall perturbation across all layers with respect to the final loss. SqLLM first approximates the objective function to get a form weighted by the scaling factor introduced by the second-order derivative of the Hessian. But since computing the Hessian is costly, SqLLM further approximates the objective based on the Fisher information matrix. Finally, the authors approximate the Fisher information matrix as a diagonal matrix as follows: 
\begin{equation}Q(w)^* = \arg\min_Q \sum_i F_{ii} (w_i - Q(w_i))^2,\end{equation}
where $Q(w)^*$ quantifies how much the model gets perturbed after quantization, $F_{ii}$ is the Fisher information approximation to the Hessian diagonal, and $Q(w_i)$ is the quantized value of weight $w_i$. This formulation biases the K-means centroids to be closer to more sensitive weights. In addition, the Dense-and-Sparse decomposition is used to filter out the outliers from the weight matrix and enhance the quantization resolution with minimal overhead. Particularly, SqLLM decomposes the weight matrix into a sparse matrix (S) with the outliers (stored in CSR format) and a dense matrix (D) without outliers. SqLLM also identifies sensitive outliers within the isolated sparse matrix by using the Fisher information and retains those weights in FP16. For efficient implementation, the authors develop 3/4-bit CUDA LUT-based kernels for matrix-
vector multiplication which load the non-uniformly quantized weights and dequantize them "piece-by-piece" to minimize the utilization of memory bandwidth. The quantized
matrices store 3/4-bit indices, corresponding to LUT
entries containing FP16 values. After dequantization,
all arithmetic is performed in FP16. They also develop kernels for sparse matrix-vector multiplication which load the sparse matrix in CSR format. SqLLM implements balanced hybrid kernels similar to \cite{flegar2017balanced}.

% SqLLM achieves high accuracy across various LLaMA, OPT, and Vicuna models by preserving only 0.45\% of outlier and sensitive weights in FP16, enabling 3/4 bit quantization with less than 0.1-0.5 perplexity drop. It outperforms AWQ \cite{lin2024awq} and SpQR \cite{dettmers2023spqr} in both compression and performance while offering up to $2.4\times$ latency speedup compared to FP16 baseline during inference on A6000 GPUs.

\subsubsection{CMPQ: channel-wise mixed-precision quantization for large language models}
\label{CMPQ}
%\interlayertag \quad \intralayertag

CMPQ \cite{chen2024channel} is a PTQ framework that supports mixed-precision quantization, quantizing non-sensitive weight channels to 2/3/4 bits, while preserving outlier channels in full precision (FP16). CMPQ develops an algorithm to adaptively quantize LMs under any given average bit constraint, applying channel-wise non-uniform quantization, where different weight channels within each layer are quantized to different bitwidths based on their sensitivity, identified using activation distributions. To perform non-uniform quantization, a K-means clustering algorithm is applied to each channel in the weight matrix, where each weight in the matrix is represented by its nearest centroid from the set of K centroids after clustering. CMPQ chooses the value of K depending on the assigned precision to the channel.

To determine the bit allocation for the channels, CMPQ computes the per-channel L2-norm of the activations and uses this metric to allocate precision according to an average bitwidth constraint $b\in[2,4]$: when the average bitwidth $b > 3$, the most salient channels; i.e. those with large activation norms, are assigned higher precision (4-bit) to boost model performance, when $b < 3$, the less important channels are quantized to lower precision (2-bit) to reduce quantization loss. For the case of $b=3$ (3-bit quantization), approximately 1\% of the most salient weight channels are protected by assigning them 4-bit precision to preserve accuracy. Additionally, CMPQ introduces an efficient outlier protection mechanism that detects both activation-based outliers and a small subset of quantization-sensitive outliers, storing them in a sparse FP16 format to minimize distortion in clustering and centroid alignment during quantization. Specifically, for activation-based outliers, channels corresponding to the top 0.45\% largest values in the L2-norm vector of the activation are kept in FP16. For detecting a small subset of quantization-sensitive outliers, CMPQ applies another K-means clustering step before quantization, and accordingly preserves 0.05\% of the magnitude-based outliers in FP16.

% CMPQ consistently outperforms integer-only and prior mixed-precision baselines across OPT and LLaMA2 models, especially under 2-bit and 3-bit constraints. It offers a strong trade-off between accuracy and memory savings, showing significant perplexity improvements on C4 and WikiText-2 benchmarks while incurring minimal latency and memory overhead during inference.

\subsection{Mixed weights, fixed activations quantization frameworks}

\subsubsection{MixLLM: LLM quantization with global mixed-precision between output-features and highly-efficient system design}
\label{MixLLM}
%\interlayertag \quad \intralayertag

MixLLM \cite{zheng2024mixllm} tackles LM compression by assigning different bitwidths to the output features or channels in linear layers; those with high salience receive higher precision than the rest forming structured mixed-precision. It relies on a global mixed‐precision search procedure, where the precision of all output features in all linear layers is determined globally rather than locally. MixLLM relies on a global salience identification step that estimates how much each feature contributes to the final loss. First, Taylor Expansion is used to approximate $l(.)$, the loss function with respect to a single channel, relying on the second-order gradient (the Hessian matrix) with respect to the channel. The Hessian is further approximated by the Fisher information matrix (with respect to a channel) on a calibration dataset. MixLLM then applies a final approximation to get a single vector product, and final approximated saliency metric, $S_c$, of channel $c$ can be derived as follows:
\begin{equation}S_c=\frac{1}{|D|}\sum_{d\in D}|g_d^T(c_q-c_0)+\frac{1}{2}(g_d^T(c_q-c_0))^2|,\end{equation}
where $c_q$ is the quantized weight of the channel, $c_0$ represents the original weight, $D$ is the calibration dataset, and $g$ is the gradient of the loss with respect to the channel. First, the global precision search procedure calculates the salience of all output channels of all linear layers. Then, it sorts them in global descending order, and quantizes $N_{largebits}$ channels ($N_{largebits}$ is a global user-defined threshold) to 8-bits, while quantizing the rest of the channels into 4-bits. Group‐wise quantization is applied: for 4-bit weights, MixLLM uses group-wise asymmetric quantization, while for the 8-bit weights (including those “high salience” channels in the weight), it employs group-wise symmetric quantization. 

Additionally, activations are quantized to 8 bits with symmetric group‐wise quantization. It is worth noting that any quantization methodologies (like GPTQ \cite{frantar2022gptq}, clip search) can be applied independently to quantize into 4-bits and 8-bits. System‐wise, MixLLM uses a two-step dequantization where it partially dequantizes the weight, and then multiplies it by the quantized activation utilizing the 8-bit Tensor Core. After that, it multiplies this matrix-multiplication result by the two scales within each group, converting the integer to float instruction into two add/sub instructions, and fusing the integer subtraction into the Tensor Core Matrix Multiply-Accumulate. In addition, MixLLM designs an end-to-end software pipeline of the quantized kernel which overlaps memory transfers with computations, dequantization computation with Single Instruction Multiple Threads (SIMT) Core, and matrix-multiplication computation with Tensor Core for maximum efficiency. It can also minimize the overhead of group-wise dequantization. Moreover, MixLLM executes different sub-problems in parallel on the GPU with CUDA Graph, implementing this function with the fused epilogue of the matrix-multiplication kernel.

%Overall, MixLLM achieves near float-level accuracy with significantly reduced memory usage, surpassing many prior 4-bit and 8-bit quantization approaches in perplexity and throughput.

\subsubsection{ResQ: mixed-precision quantization of large language models with low-rank residuals}
\label{ResQ}
%\interlayertag \quad \intralayertag

ResQ \cite{saxena2024resq} is a PTQ mixed-precision weight, activation, and KV cache quantization framework for LMs which identifies, via Principal Component Analysis (PCA), a low-rank high-variance subspace; i.e., with high activation variances. Then it retains the coefficients of the projections along the basis in 8-bits. The rest of the coefficients are quantized to 4-bits (or 2-3 bits, depending on the target). For each subspace, ResQ applies an invariant random rotation to handle outliers. In particular, ResQ projects weights, activations, and the KV cache into an orthogonal basis, $U$, constructed as a combination of two rotation matrices having the following properties: 1) the more important components are captured by the low-rank space for high-precision quantization, and 2) the quantization error in the high-precision group and low-precision group is minimal. $U$ is defined as follows:
\begin{equation}U=PR=[P_lP_h]\begin{bmatrix}
R_l & 0 \\
0 & R_h 
\end{bmatrix},\end{equation}
where $P_l,R_l$ correspond to low-precision components and $P_h, R_h$ correspond to high-precision components. To reduce the outliers, $R_l,R_h$ are made random orthogonal matrices, and that way the rotated matrices are easier to quantize. ResQ shows that the low-rank subspace for high-precision quantization can be obtained by PCA,
while the subspace for low-precision quantization can be obtained using the following: $U_hU_h^T+U_lU_l^T=PhP_h^T+P_lP_l^T=I$, where $U_h$ represents bases of a low-rank space of high-precision components, and $U_l$ represents the complementary subspace of low-precision components.

Quantized activations ($X_q$) and weights ($W_q$) are obtained by projecting the input space of activations and weights by $U$ and $U^T$ respectively and quantizing the coefficients. Once the projection matrices are obtained, and during inference, the output of the layer, $X_qW_q$, is calculated by multiplying the weights and activations by $U$ as follows: $X_qW_q=Q_L(XU_l)Q_L(U_l^TW)+Q_H(XU_h)Q_H(U_h^TW)$, where Q(.) is the quantization function. $L$ represents the subscript for low-precision and $H$ is the subscript for high-precision. Weights can be projected and quantized offline, and the projection of an activation is merged to the weight of a previous linear layer. 

ResQ introduces four different kinds of projections based on decoder only LM architecture: projection $U_A$ at block boundaries (modifying the inputs across blocks and enabling better quantization), two projections $U_B$ and $U_C$ within the attention block enabling mixed precision quantization of KV cache, and a projection $U_D$ within the feedforward block projecting the activations and weights of $down\_proj$ layer. Only a single projection at block boundaries is shared across all layers while the rest of the projections are generated per layer. ResQ applies per-token asymmetric quantization to activations, per-channel symmetric quantization to weights, and per-head asymmetric quantization to the KV cache. It fuses the projection matrices $U_A$, $U_B$, and $U_D$ into adjacent weights and applies GPTQ \cite{frantar2022gptq} for weight quantization. Moreover, implement on-the-fly projections, $U_D$ is a Hadamard matrix and $U_C$ and its activations are quantized to a precision of 8. 

% Experiments show that ResQ surpasses leading baselines (e.g., SpinQuant, QuaRot, QUIK) by up to 4-33\% lower perplexity or up to 5\% higher accuracy on zero-shot tasks, even in fully quantized 4/4/4 configurations. The end result is near-float performance across tasks (language modeling, common-sense reasoning, MMLU, etc.) while drastically lowering memory and computation costs.

\subsection{Mixed weights, mixed activations quantization frameworks}

\subsubsection{Atom: low-bit quantization for efficient and accurate LLM serving}
\label{Atom}
%\interlayertag \quad \intralayertag

Atom \cite{zhao2024atom} is a low-bit weight-activation mixed-precision quantization framework for LM serving that combines: 1) mixed-precision quantization with channel reordering, 2) fine-grained group quantization, 3) dynamic activation quantization, and 4) KV-cache quantization. For mixed-precision quantization, Atom identifies activation outlier channels; i.e., those with high mean value, reorders them and their corresponding weight channels to the end of the matrix inspired by RPTQ \cite{yuan2023rptq}, and quantizes them to a higher precision (INT8) compared to the other channels, which are quantized to a lower bitwidth (4-bits). 

While weight reordering incurs a one-time cost (since outlier channels can be identified offline using a calibration dataset), activation reordering is performed online. To lower the cost of activation reordering, Atom fuses these reordering operators with prior operators. To tackle the limited representation capability of 4-bit precision, Atom uses group-wise quantization of the non-outlier channels. Specifically, the activation groups are first multiplied by their corresponding weight groups via Tensor Cores. Then, Cuda Cores are used to dequantize the temporary results (with different quantization parameters) to FP16 before performing addition. We note that Atom fuses the dequantization and addition into the GEMM kernel. In addition, Atom uses symmetric dynamic quantization for activations so that activation matrices would have their tailored quantization parameters during inference, where quantization operators are fused into prior operators. 

Atom also incorporates GPTQ \cite{frantar2022gptq} in an offline stage for weight matrix quantization. Meanwhile, the KV cache is quantized asymmetrically to shrink self-attention’s memory overhead, and is dequantized directly before performing FP16 calculation. Implementation wise, Atom relies on a low-precision unit for compute-bound layers, while fusing dequantization with FlashInfer \cite{ye2024accelerating}, an LM serving library. Moreover, it also uses PagedAttention \cite{kwon2023efficient} to enable efficient inference with large batch sizes.

% Tested on Llama, Llama-2, and others, Atom preserves near-baseline perplexities, minimal zero-shot accuracy drops (generally 1-2\%), and provides up to a $7.7\times$ speedup over FP16 and a $2.5\times$ improvement over 8-bit baselines.

\subsubsection{MASE: a dataflow compiler for efficient llm inference using custom microscaling formats}
\label{MASE}
%\interlayertag \quad \intralayertag

MASE \cite{cheng2023dataflow} is a dataflow compiler and co-design framework that leverages microscaling (MX) data formats, particularly MXInt, to balance accuracy and hardware efficiency for LM inference on hardware accelerators. A key contribution is the MASE Intermediate Representation (IR), a hardware-aware trainable software which allows exploring software–hardware optimizations like scheduling and parallelism for custom low-bit formats for both weights and activations at granularities ranging from the model level down to the bit level. MASE IR uses the static single-assignment form which enables an easy translation into a dataflow hardware representation. The hardware design attributes carried by a model in MASE IR include the shape of streaming tiles, the streaming order, the interface between hardware and data, and the estimated throughput. These attributes enable parallelism exploration by the model optimizer. 

MASE has 44 general and type-independent "analysis and optimization" passes which target the different granularities. For the "quantize" pass, a model is first quantized into a set of user-defined bitwidths. MASE supports both PTQ and QAT. After that the "parallelize" pass explores the best hardware resources for parallelism given a certain hardware resource budget. The "evaluate" pass estimates the accuracy and the area efficiency of the hardware because MASE IR has both software and hardware design parameters. Guided by a hardware-aware cost function, MASE’s "search" pass iterates over both quantization and hardware parallelism settings using existing algorithms like Tree-structured Parzen Estimator (TPE) \cite{ozaki2020multiobjective} to find mixed-precision assignments. 

Eventually, after a given number of iterations, MASE automatically maps the model into a SystemVerilog hardware design for the chosen configuration in the "emit" pass. To enable MASE to support exploration of custom data formats at scale, software emulators and hardware components must be added by users. While software emulators are used to indicate how to quantize/dequantize the value from/to the custom formats/floating-point numbers, MASE provides a Verilog template of a dataflow component with a set of parallelism parameters for each hardware component. MASE then automatically explores hardware designs by sweeping those parallelism parameters restricting the design space. We note that MASE IR provides a compact representation to explore large models up to billions of parameters.

% Evaluated on BERT, OPT, and LLaMA families, this framework preserves near-FP32 accuracy (often within 1-2\%) at 4-5 average bits, attaining area efficiency on par with or better than INT8 baselines, and yielding up to a 24\% improvement in the accuracy–bitwidth trade-off.

\subsubsection{BlockDialect: block-wise fine-grained mixed format for energy-efficient LLM inference}
\label{BlockDialect}
%\interlayertag \quad \intralayertag
 
BlockDialect \cite{jang2025blockdialect} is a block-wise, fine-grained, mixed-format quantization framework designed to boost energy efficiency and accuracy for LM inference at 4-bit precision, by focusing on "how to represent" data distribution over
"how to scale". Particularly, it introduces a “formatbook” of FP4 variants called DialectFP4, where each “dialect” adjusts large-magnitude values slightly differently. The dialects for the formatbook were determined based on profiling several models (like Llama3-8B, Llama2-7B, Mistal-7B, OPT-6.7B, and others), where the matrix is first divided into blocks (size 32), then each block is scaled by the shared exponent, and finally the magnitude distribution histograms for each block are accumulated. The profiling analysis reveals that FP4 E2M1 (referred to as FP4) aligns with the observed distribution, and is hence chosen as the dialects' base format. 

The authors then choose 16-Dialect DialectFP4 to meet three main criteria: 1) dialects cover all possible maximum magnitudes, 2) on the one hand, each pair of dialects shares the maximum magnitude, while on the other hand, the pairs differ in one large magnitude value, and 3) the unit of these dialects is set to $0.5$ to align with FP4. Accordingly, in addition to the 4-bit identifier assigned to each block, each data point in DialectFP4 has 1 bit for sign and 3 bits for index. Unlike weights whose optimal per-block dialect can be determined by calculating the MSE, determining the optimal per-block dialect for activations is done at runtime via a lightweight, two-stage selection process preceded by a pre-processing stage. In the pre-processing stage, a 5-bit shared exponent is computed, then each element’s exponent is adjusted by
subtracting the shared exponent. 

After that, BlockDialect shifts the mantissa by the adjusted exponent, and truncates the lower bits to form a 5-bit intermediate representation with 3 bits for the integer part and 2 bits for the fractional part. Following the pre-processing stage, BlockDialect determines the block’s maximum magnitude by rounding from the second fractional bit and limits the choice to the two dialects (FP4 variants) whose largest magnitudes are the same as the block's maximum. In the second stage and for each one of the two selected dialects in the first stage, BlockDialect looks at how many elements in the block lie within that dialect’s “beneficial range,” i.e., the specific segment of magnitudes where that dialect is more precise. The beneficial range is calculated as the midpoint between the differing value,
its adjacent value, and the differing value of the paired dialect. Whichever dialect can accommodate more elements in this beneficial range is assigned to the entire block. Implementation-wise, since DialectFP4 is compatible with 5-bit integer arithmetic operations, it enables two implementation options. BlockDialect can either use the general INT4 MAC with simple
logic to deal with residual bits for 5-bit multiplication or it can design optimized MACs with 4-bit unsigned integer multiplier and an additional XOR gate to deal with the sign bit.

% Tested on LLaMA, LLaMA-2, Mistral, and OPT families, BlockDialect delivers significantly higher accuracy than existing MX-format baselines, while remaining within ~5\% of full precision even in “full-path” quantization (covering activation–activation multiplications). This focus on “how to represent” each block, rather than just “how to scale” blocks allows for accurate outlier handling and hardware reuse of low-bit integer operations, substantially improving the trade-off between memory (cost) and model quality in LM inference.

\subsubsection{QUIK: towards end-to-end 4-bit inference on generative LLMs}
\label{QUIK}
%\interlayertag \quad \intralayertag

QUIK \cite{ashkboos2023quik} is a joint mixed-precision PTQ framework for both weights and activations in LMs, achieving speedups while maintaining near-baseline perplexity. QUIK quantizes most weights and activations into a precision of 4 bits, while keeping outliers in higher precision. Its selective-precision approach improves upon GPTQ \cite{frantar2022gptq} by identifying the outlier weight columns; i.e. those with the largest $l\infty$ norm, for each layer using a small calibration set. After that, the same outlier indices are used to identify the outlier columns. The outlier weight columns are rearranged to the end of the matrices and are quantized up to the index of the outliers offline. QUIK does not quantize the outliers to preserve them. 

At runtime, the activation columns corresponding to the outlier weights identified by the indices are extracted and kept in high precision while the rest of the columns are quantized. By aggregating the quantization errors to the columns kept in high precision (FP16 or BF16), and by excluding weight outliers from the 4-bit quantization scale, QUIK enhances GPTQ quantization. Moreover, QUIK identifies weight clipping thresholds by a linear search to improve end-to-end perplexity. We note that QUIK quantizes the down projection layers into 8 bits rather than 4 bits as these layers are more sensitive to quantization as suggested by the $l\infty$ norm analysis. 

While the weight columns kept in full-precision are multiplied by their corresponding full-precision activation columns regularly, the quantized columns go through the following quantized matrix multiplication pipeline: Quick first dynamically applies asymmetric quantization per token for activations, then, by the help of the GPU's INT8/INT4 tensor-cores, performs the matrix multiplication of the dynamically quantized activations and weights (which are quantized offline in a symmetric per output manner). Finally, it dequantizes the result back to full-precision. It also fuses the quantization and dequantization for efficient implementation. In addition, for some experiments, QUIK extends SparseGPT \cite{frantar2023sparsegpt} to jointly quantize and sparsify the models, while keeping the outliers in dense FP16.

\begin{figure}[!t]
    \centering
    % --- Subfigure (a) ---
    \subfloat[]{
        \includegraphics[width=0.45\textwidth]{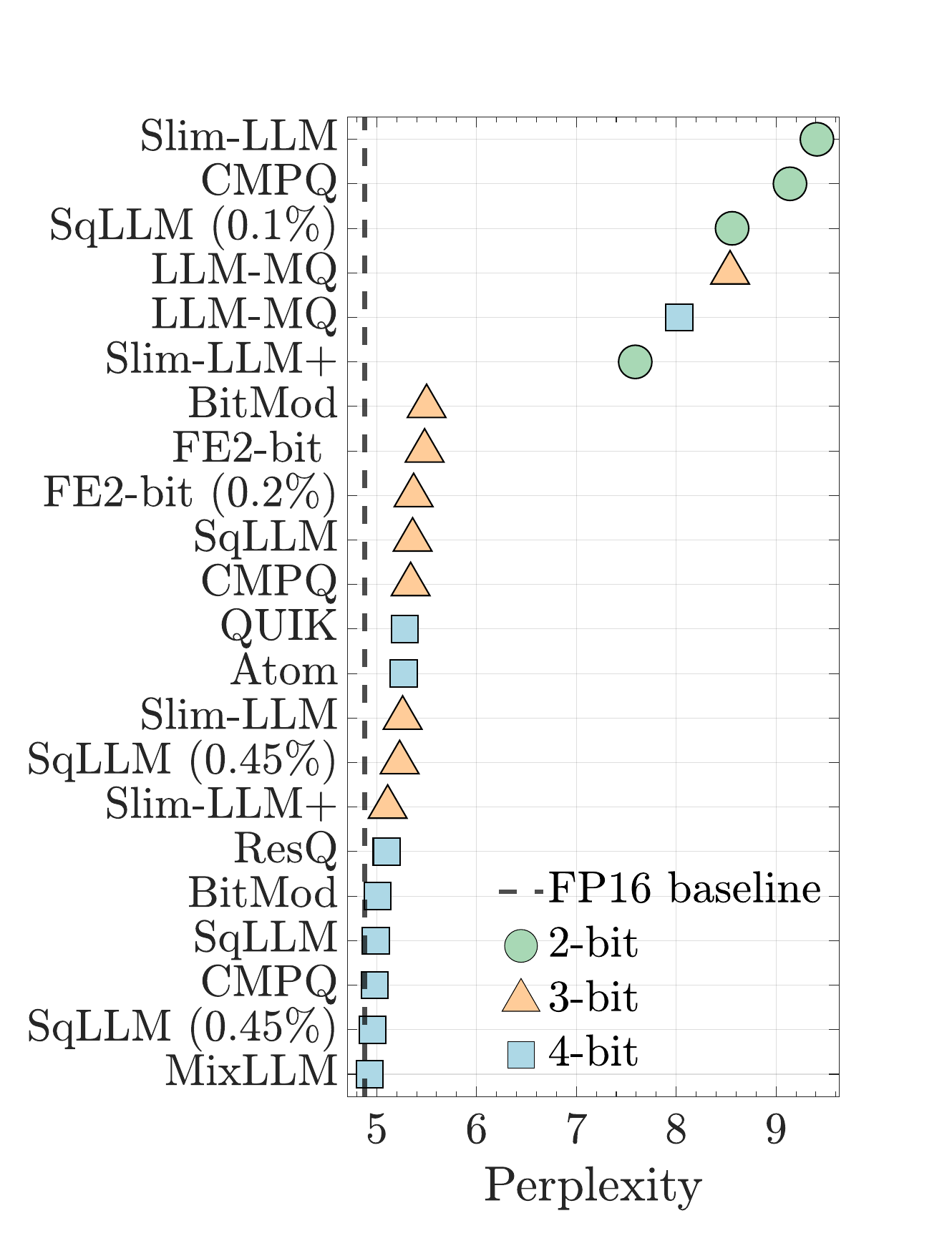}
        \label{fig:perplexity_llama2_13b}
    }
    \hfill
    % --- Subfigure (b) ---
    \subfloat[]{
        \includegraphics[width=0.45\textwidth]{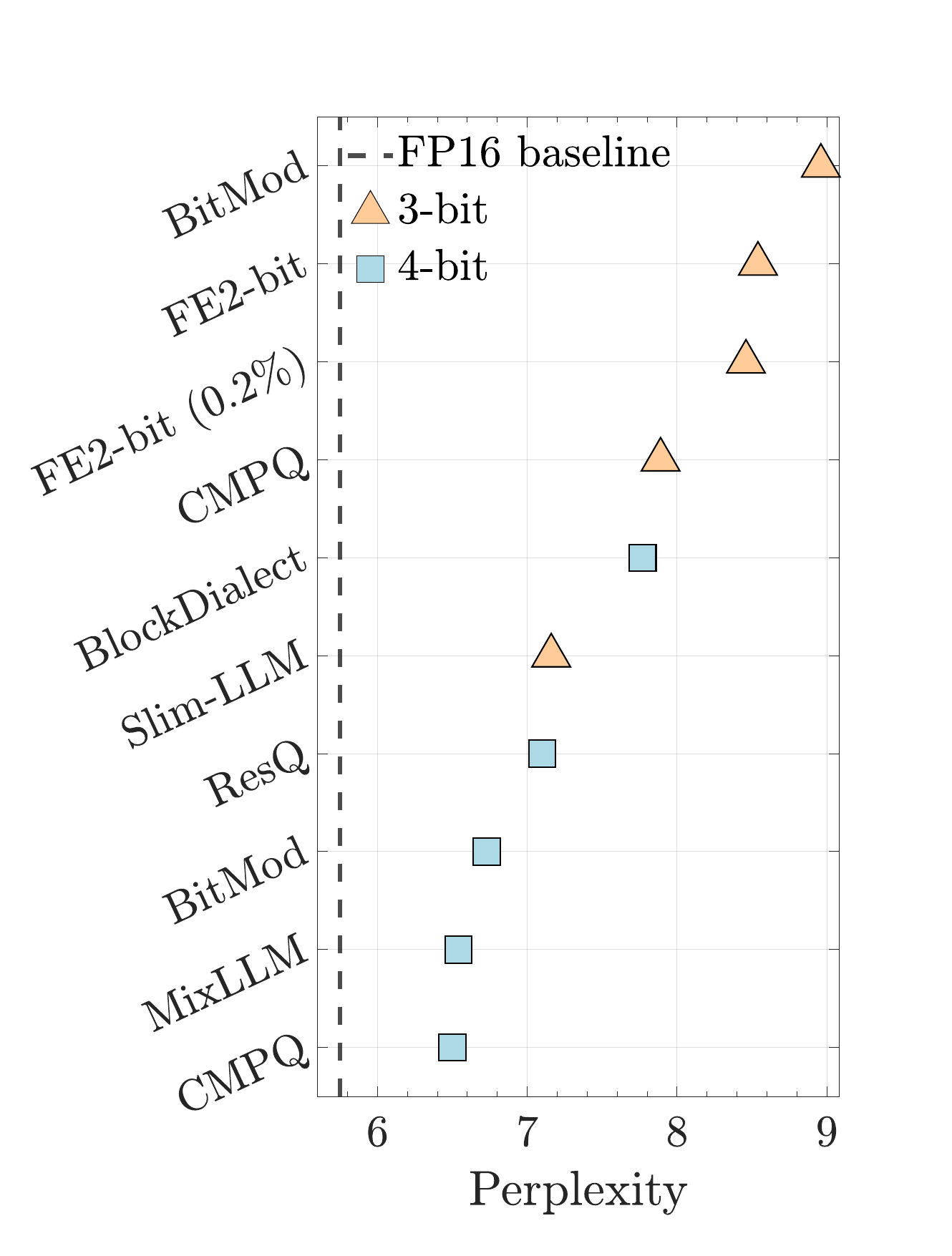}
        \label{fig:perplexity_llama3_8b}
    }

    \caption{Perplexity of MXPLM frameworks reported on (a) LLaMA2-13B and (b) LLaMA3-8B using WikiText2 for different average precisions. Note: ``(x\%)'' indicates sparsity of $x\%$.}
    \label{fig:perplexity_comparison}
\end{figure}
% Thanks to that design, QUIK often reduces memory by $3-4\times$ and obtains up to $3.4\times$ throughput gains on GPUs. 
% \begin{figure}[!ht]
%     \centering
%     \includegraphics[width=0.75\textwidth]{Figs/llama2_13b_perplexity.pdf}
%     \caption{Perplexity of MXPLM frameworks reported on LLaMA2-13b using Wikitext2 for different average precisions. Note: "(x\%)" means a sparsity of "x\%".}
%     \label{fig:perplexity_llama2_13b}
% \end{figure}

% \begin{figure}[!ht]
%     \centering
%     \includegraphics[width=0.75\textwidth]{Figs/llama3_8b_perplexity.pdf}
%     \caption{Perplexity of MXPLM frameworks reported on LLaMA3-8b using Wikitext2 for different average precisions. Note: "(x\%)" means a sparsity of "x\%".}
%     \label{fig:perplexity_llama3_8b}
% \end{figure}

% \begin{figure}[!ht]
%     \centering
%     \includegraphics[width=0.75\textwidth]{Figs/llama2_13b_zero_shot.pdf}
%     \caption{Zero-shot accuracy of MXPLM frameworks reported on LLaMA2-13b using zero-shot tasks for different average precisions. Note: "(x\%)" means a sparsity of "x\%" .}
%     \label{fig:zeroshotaccuracy_llama2_13b}
% \end{figure}

% \begin{figure}[!ht]
%     \centering
%     \includegraphics[width=0.75\textwidth]{Figs/llama3_8b_zero_shot.pdf}
%     \caption{Zero-shot accuracy of MXPLM frameworks reported on LLaMA3-8b using zero-shot tasks for different average precisions. Note: "(x\%)" means a sparsity of "x\%" .}
%     \label{fig:zeroshotaccuracy_llama3_8b}
% \end{figure}

\section{Insights \& discussions}
\label{sec:insights&discussions}

\subsection{Comparison between MXPLM frameworks}

Most existing MXPLM frameworks rely on PTQ due to its lower computational cost, with only a few exploring quantization-aware training QAT (like FE2-bit, PB-LLM, MASE). The selection of weights/activations to quantize is often guided by first- or second-order information, leveraging gradient-based methods such as the Hessian matrix. While weight quantization has been the primary focus, activation quantization remains more challenging, and a common approach is to use a single low-bitwidth setting alongside a higher-bitwidth setting for select outlier parameters. While the goal of MXPLM is to a hardware-friendly implementation, only few works (MASE and LLM-PQ) explicitly consider mixed-precision with hardware in the loop (these frameworks are called hardware-aware). Notably, some works in literature define mixed-precision merely as using different bitwidths for weights and activations while maintaining uniform precision within each category, but that was outside the scope of our definition of mixed-precision. As indicated in section \ref{hardwareInsights}, due to the lack of hardware support for mixed-precision computation, weights that are quantized to INT4 are usually dequantized to FP32 for GEMM operations, with accumulation typically performed in higher precision formats such as BF16 or FP32 \cite{shen2020q,zadeh2020gobo}. 

We henceforth analyze the perplexity/zero-shot accuracy behavior of MXPLM frameworks across LLaMA2-13B and LLaMA3-8B. We omit frameworks that do not evaluate on these models. As shown in Fig. \ref{fig:perplexity_llama2_13b}, for LLaMA2-13B, the FP16 baseline perplexity is $4.88$ and several 4-bit (on average) MXPLM frameworks, notably MixLLM, CMPQ, and SqLLM ($0.45\%$ sparsity) achieve near-baseline performance (perplexity ranges between $4.93-4.98$), and BitMod and ResQ are also competitive (within $2.66-4.51\%$ difference with respect to baseline). The best (perplexity-wise) 3-bit (on average) methods (Slim-LLM+, SqLLM ($0.45\%$), and Slim-LLM) remain below $7.17\%$ difference with respect to FP16. In contrast, the performance of 2-bit average precision frameworks such as Slim-LLM+, SqLLM ($0.1\%$), CMPQ, and Slim-LLM degrades severely as perplexity increases ($>55\%$ difference with respect to FP16).

For LLaMA3-8B (Fig. \ref{fig:perplexity_llama3_8b}, the FP16 baseline perplexity is $5.75$. The best performing MXPLM framework, CMPQ, has a $13.04\%$ higher perplexity than the baseline at an average precision of 4-bits, while BlockDialect performs worst among the frameworks at average precision of 4 with a perplexity of $7.77$. At an average precision of 3-bits, MXPLM frameworks exhibit larger performance loss, with BitMod approaching a perplexity of 9, indicating that no 2-bit or low 3-bit configuration is competitive with respect to the FP16 baseline.

\begin{table}[!t]
\centering
\begin{tabular}{llcccccc}
    \toprule
    \#Bits & Method & PIQA & ARC-e & ARC-c & BoolQ & HellaSwag & Winogrande \\
    \midrule
    {FP16 \cite{bondarenko2024low}}
     & - &      80.55 &80.52 &72.22& 77.44 & 48.98 &79.38 \\
     
    \midrule
    \multirow{1}{*}{2-bit}
     & LLM-MQ & \textbf{75.84} & \textbf{54.29} & – & – & \textbf{68.32} & \textbf{65.51} \\
    \midrule
    \multirow{5}{*}{3-bit}
     & SqLLM & 78.62 & 56.73 & – & – & 74.53 & 67.40 \\
     & LLM-MQ & 79.00 & 57.79 & – & – & 75.08 & 68.59 \\
     & BitMod & \textbf{79.22} & – & – & – & \textbf{76.79} & \textbf{72.37} \\
     & FE2-bit & 77.20 & 71.2 & – & – & 57.6 & 68.50 \\
     & FE2-bit (+0.2\%) & 77.80 & \textbf{71.3} & – & – & 58.8 & 68.60 \\
    \midrule
    \multirow{4}{*}{4-bit}
     & SqLLM & 78.94 & 57.70 & – & – & 76.05 & 69.14 \\
     & LLM-MQ & 79.49 & 58.50 & – & – & 76.31 & 69.30 \\
     & BitMod & \textbf{80.42} & – & – & – & \textbf{78.41} & \textbf{72.14} \\
     & ResQ & 79.10 & \textbf{76.10} & \textbf{49.1} & \textbf{79.70} & 77.90 & 69.90 \\
     & QUIK & 79.22 & 74.92 & 48.04 & – & 78.36 & 71.90 \\
    \bottomrule
\end{tabular}
\caption{Zero-shot accuracy (\%) on LLaMA2-13B across tasks (PIQA, ARC-e, ARC-c, BoolQ, HellaSwag, Winogrande). Frameworks with no reported results are omitted. Note: "(x\%)" means a sparsity of "x\%".}
\label{tbl:zero_shot_llama2-13b}
\end{table}

The best-performing frameworks in both models (MixLLM, SqLLM, CMPQ, BitMod, ResQ) belong primarily to the MPW and MPW,UPA categories, which exploit mixed-precision weights and uniform precision or FP precision for activations, with frameworks at an average precision of 4-bits performing best. MPW,MPA frameworks (Atom, BlockDialect, QUIK) can offer comparative performance with respect to FP16 at an average precision of 4 bits. {The perplexity results of all surveyed MXPLM frameworks can be found in Tables \ref{tbl:llama1_models_ppl_c4_full_table} and \ref{tbl:llama2-3_ppc_}, where the best performing framework metric is in bold.}

While perplexity provides a useful proxy for evaluating LM compression quality, it does not fully capture downstream task performance. Many real-world applications depend on zero-shot generalization across diverse benchmarks (like commonsense reasoning, natural language inference, or knowledge-intensive QA), where error propagation can be amplified even when perplexity remains close to the FP baseline. Evaluating on zero-shot tasks therefore complements perplexity analysis by exposing robustness and transferability under task-driven conditions. 
\begin{table}[!t]
    \centering
    \begin{tabular}{llcccccc}
        \toprule
        \#Bits & Method & PIQA & ARC-e & ARC-c & BoolQ & HellaSwag & Winogrande \\
            \midrule
    {FP16\cite{bondarenko2024low}}
     &– & 81.44 &80.79 &72.85& 77.74 &53.33 & 79.16 \\
        \midrule
        \multirow{3}{*}{3-bit}
         & BitMod & \textbf{77.91} & – & – & – & \textbf{73.56} & 70.32  \\
         & FE2-bit & 75.3 & 69.4 & – & – & 54.9 & 70.3 \\
         & FE2-bit (+0.2\%) & 76.8 & \textbf{70.3} & – & – & 55.4 & \textbf{70.5}  \\
        \midrule
        \multirow{4}{*}{4-bit}
         & BitMod & \textbf{79.98} & – & – & – & \textbf{78.49} & \textbf{73.09} \\
         & MixLLM & – & – & \textbf{53.67} & – & 78.2 & –  \\
         & ResQ & 78.3 & \textbf{75.0} & 49.2 & 72.5 & 76.5 & 71.0 \\
         & BlockDialect & 75.24 & 71.63 & 47.18 & \textbf{77.37} & 74.12 & 66.38  \\
        \bottomrule
    \end{tabular}%
    \caption{Zero-shot accuracy (\%) on LLaMA3-8B across tasks (PIQA, ARC-e, ARC-c, BoolQ, HellaSwag, Winogrande). Note: "(x\%)" means a sparsity of "x\%"}
    \label{tbl:llama3-8b_zero_shot}
\end{table}
In Tables \ref{tbl:zero_shot_llama2-13b} and \ref{tbl:llama3-8b_zero_shot}, we report the presented zero-shot accuracy results across 6 tasks: PIQA, ARC-e, ARC-c, BoolQ, HellaSwag, and Winogrande on LLaMA2-13B and LLaMA3-8B. The highest reported zero-shot accuracies on LLaMA2-13B are $80.42\%$ (BitMod), $76.10\%$ (ResQ), $49.1\%$ (ResQ), $79.70\%$ (ResQ), $78.41\%$ (BitMod), and $72.14\%$ (BitMod) across PIQA, ARC-e, ARC-c, BoolQ, HellaSwag, and Winogrande respectively. At an average precision of 4-bits, and on LLaMA3-8B, the highest reported zero-shot accuracies are $79.98\%$ (BitMod), $75.0\%$ (ResQ), $53.67\%$ (MixLLM), $77.37\%$ (BlockDialect), $78.49\%$ (BitMod), and $73.09\%$ (BitMod) across PIQA, ARC-e, ARC-c, BoolQ, HellaSwag, and Winogrande respectively. Again, the best performing frameworks are MPW and MPW, UPA. We note that ResQ which performs well across tasks quantizes weights, activations, and also the KV cache, offering even higher compression gains than frameworks that quantize weights/activations. {The zero-shot accuracy of all the surveyed frameworks on different models and across different tasks with the best performing metrics in bold can be found in Table~\ref{tbl:llama1_2_zero_shot}.}

\newpage
\begin{table}[H]
\centering
\begin{tabular}{ll*{8}{c}}
\toprule
& &\multicolumn{2}{c}{Llama1-7B} & \multicolumn{2}{c}{Llama1-13B} & \multicolumn{2}{c}{Llama1-30B} & \multicolumn{2}{c}{Llama1-65B}  \\
\#Bits & Method & WikiText2 &C4 &WikiText2 &C4 &WikiText2 &C4 &WikiText2 &C4 \\
\midrule
    FP16 &- &5.68 &7.08 &5.09 &6.61 &4.1 &5.98 &3.53 &5.62 \\
\midrule
\multirow{3}{*}{2-bit}
    &Slim-LLM &14.58 &32.91 &8.87 &13.85 &7.33 &11.27 &5.9 &10.95 \\
    &Slim-LLM+ &\textbf{9.68} &\textbf{14.99} &\textbf{7.17} &\textbf{10.22} &\textbf{6.41} &\textbf{9.33} &\textbf{5.74} &\textbf{7.52} \\
    &PB-LLM &24.61 &49.73 &17.73 &26.93 &12.65 &17.93 &7.85 &11.85 \\
\midrule
\multirow{9}{*}{3-bit}
    &Slim-LLM &6.4 &\textbf{6.14} &5.48 &\textbf{6.05} &4.61 &\textbf{6.33} &3.99 &\textbf{5.94} \\
    &Slim-LLM+ &\textbf{6.07} &7.75 &\textbf{5.37} &6.91 &\textbf{4.34} &6.36 &\textbf{3.72} &5.96 \\
    &SqLLM &6.32 &7.75 &5.6 &7.08 &4.66 &6.37 &4.05 &5.99 \\
    &SqLLM (+0.45\%) &6.13 &7.56 &5.45 &6.92 &– &– &– &– \\
    &OWQ &6.66 &– &5.66 &– &4.75 &– &4.25 &– \\
    &APTQ &6.76 &6.24 &– &– &– &– &– &– \\
    &FE2-bit &6.61 &– &5.92 &– &– &– &– &– \\
    &FE2-bit (+0.2\%) &6.56 &– &5.89 &– &– &– &– &– \\
    &Atom &11.77 &15.43 &8.4 &10.81 &6.94 &9.14 &5.89 &7.94 \\
\midrule
\multirow{6}{*}{4-bit}
    &SqLLM &5.79 &7.21 &5.18 &6.71 &\textbf{4.22} &\textbf{6.06} &3.76 &\textbf{5.69} \\
    &SqLLM (+0.45\%) &\textbf{5.77} &7.18 &\textbf{5.17} &\textbf{6.68} &– &– &– &– \\
    &OWQ &5.94 &– &5.25 &– &4.25 &– &3.74 &– \\
    &SpQR &5.87 &7.28 &5.22 &6.72 &4.25 &6.08 &\textbf{3.68} &5.7 \\
    &APTQ &6.45 &\textbf{5.23} &– &– &– &– &– &– \\
    &Atom &6.16 &7.7 &5.46 &7.03 &4.54 &6.35 &3.89 &5.92 \\
\bottomrule
\end{tabular}%
\caption{Perplexity across LLaMA1 model (WikiText2, C4). Note: "(x\%)" means a sparsity of "x\%"}
\label{tbl:llama1_models_ppl_c4_full_table}
\end{table}

\begin{table}[H]
\centering
{
\setlength{\tabcolsep}{3pt}
\begin{tabular}{ll*{10}{c}}
\toprule
 &  &\multicolumn{2}{c}{Llama2-7B} & \multicolumn{2}{c}{Llama2-13B} & \multicolumn{2}{c}{Llama2-70B} & \multicolumn{2}{c}{Llama3-8B} & \multicolumn{2}{c}{Llama3-70B} \\
\#Bits & Method & WikiText2 & C4 & WikiText2 & C4 & WikiText2 & C4 & WikiText2 & C4 & WikiText2 & C4 \\
\midrule
FP16 &- &5.47 &6.97 &4.88 &6.46 &3.31 &5.52 &5.75 &9.22 &2.9 &6.85 \\
\midrule
\multirow{6}{*}{2-bit}
&Slim-LLM &16.01 &16 &9.41 &\textbf{9.41} &6.28 &\textbf{7.01} &\textbf{39.66} &110 &\textbf{9.46} &\textbf{15.92} \\
&Slim-LLM+ &\textbf{10.87} &18.18 &\textbf{7.59} &10.24 &6.44 &8.4 &– &– &– &– \\
&SqLLM (+0.1\%) &13.64 &– &8.56 &– &\textbf{5.38} &– &– &– &– &– \\
&CMPQ &14.37 &\textbf{15.97} &9.14 &11.25 &– &– &120 &110 &– &– \\
&LLM-MQ &– &– &12.17 &– &– &– &– &– &– &– \\
&PB-LLM &25.37 &29.84 &49.81 &19.82 &– &8.95 &44.12 &\textbf{79.21} &11.68 &33.91 \\
\midrule
\multirow{7}{*}{3-bit}
&Slim-LLM &6.24 &7.74 &5.26 &\textbf{5.26} &3.67 &\textbf{5.09} &\textbf{7.16} &13.1 &\textbf{4.08} &\textbf{8.64} \\
&Slim-LLM+ &\textbf{5.94} &7.71 &\textbf{5.11} &6.9 &\textbf{3.35} &5.85 &– &– &– &– \\
&SqLLM &6.18 &7.72 &5.36 &6.97 &3.77 &5.83 &– &– &– &– \\
&SqLLM (+0.45\%) &5.96 &\textbf{7.51} &5.23 &6.82 &3.63 &5.73 &– &– &– &– \\
&CMPQ &6.14 &7.66 &5.34 &6.93 &– &– &7.89 &\textbf{11.3} &– &– \\
&FE2-bit &6.62 &– &5.48 &– &– &– &8.54 &– &– &– \\
&FE2-bit (+0.2\%) &6.59 &– &5.37 &– &– &– &8.46 &– &– &– \\
\midrule
\multirow{10}{*}{4-bit}
&SqLLM &5.62 &7.12 &4.99 &6.57 &3.41 &5.58 &– &– &– &– \\
&SqLLM (+0.45\%) &5.57 &\textbf{7.08} &4.96 &\textbf{6.54} &\textbf{3.39} &\textbf{5.57} &– &– &– &– \\
&CMPQ &5.61 &7.1 &4.98 &6.55 &– &– &\textbf{6.5} &\textbf{9.39} &– &– \\
&MixLLM &\textbf{5.55} &– &\textbf{4.93} &– &– &– &6.54 &9.62 &\textbf{3.3} &\textbf{7.24} \\
&ResQ &5.8 &– &5.1 &– &– &– &7.1 &– &4.1 &– \\
&Atom &6.03 &– &5.27 &– &3.68 &– &– &– &– &– \\
&LLM-MQ &– &– &8.03 &– &– &– &– &– &– &– \\
&BlockDialect &6.35 &– &– &– &– &– &7.77 &– &– &– \\
&QUIK &5.84 &– &5.28 &– &3.74 &– &– &– &– &– \\
&BitMod &5.72 &7.26 &5.01 &6.61 &– &– &6.73 &9.66 &– &– \\
\bottomrule
\end{tabular}%
}
\caption{Perplexity across LLaMA2 and Llama3 families (WikiText2, C4). Frameworks with no reported numbers are omitted. Note: "(x\%)" means a sparsity of "x\%"}
\label{tbl:llama2-3_ppc_}
\end{table}

\begin{table}[H]
\centering
\begin{tabular}{llcccccc}
\toprule
&  & PIQA & ARC-e & ARC-c & BoolQ & HellaSwag & Winogrande \\
\midrule
\#Bits & Method & \multicolumn{6}{c}{LLaMA1-7B} \\
\midrule
\multirow{2}{*}{2-bit}
&Slim-LLM &57.83 &33.46 &25.09 &56.05 &36.7 &52.64 \\
&Slim-LLM+ &\textbf{64.96} &\textbf{45.66} &\textbf{28.67} &\textbf{64.59} &\textbf{48.86} &\textbf{53.35} \\
\midrule
\multirow{6}{*}{3-bit}
&SpQR &\textbf{78.13} &65.87 &\textbf{38.05} &– &55.27 &67.48 \\
&APTQ &74.5 &57.9 &36.4 &– &68.3 &65.3 \\
&FE2-bit &76.6 &70.4 &– &– &72.8 &68.5 \\
&FE2-bit (+0.2\%) &76.8 &\textbf{71} &– &– &\textbf{73.8} & \textbf{68.7} \\
&Atom &65.56 &41.41 &30.72 &\textbf{61.77} &53.19 &55.56 \\
\midrule
\multirow{4}{*}{4-bit}
&SpQR &78.45 &67.13 &38.23 &– &56.01 &67.48 \\
&APTQ &\textbf{78.6} &\textbf{72.4} &\textbf{44.4} &– &\textbf{75.7} &69.3 \\
&PB-LLM &78 &69 &42.3 &– &74.3 &\textbf{69.7} \\
&Atom &76.28 &52.1 &38.99 &\textbf{69.79} &69.81 &63.69 \\
\midrule
& &  \multicolumn{6}{c}{LLaMA1-13B} \\
\midrule
\multirow{2}{*}{2-bit}
&Slim-LLM &73.19 &47.95 &36.27 &55.92 &63.04 &61.79 \\
&Slim-LLM+ &\textbf{74.15} &\textbf{50.26} &\textbf{37.04} &\textbf{64.31} &\textbf{63.57} &\textbf{63.11} \\
\midrule
\multirow{6}{*}{3-bit}
&SpQR &78.73 &\textbf{73.27} &\textbf{42.75} &– &58.22 &68.9 \\
&APTQ &74.4 &64.1 &41 &– &71.2 &68 \\
&FE2-bit &79.2 &72.7 &– &– &76.9 &71.6 \\
&FE2-bit (+0.2\%) &\textbf{79.3} &73 &– &– &\textbf{77.8} &\textbf{71.9} \\
&Atom &70.08 &47.94 &33.7 &\textbf{63.46} &62.93 &56.75 \\
\midrule
\multirow{4}{*}{4-bit}
&SpQR &78.94 &\textbf{74.37} &43.17 &– &59.02 &69.77 \\
&APTQ &\textbf{79.9} &73.9 &\textbf{47} &– &\textbf{78.8} &\textbf{72.1} \\
&PB-LLM &– &– &– &– &– &– \\
&Atom &77.69 &57.58 &42.92 &\textbf{67.46} &73.77 &68.51 \\
\midrule
& & \multicolumn{6}{c}{LLaMA1-30B} \\
\midrule
\multirow{2}{*}{2-bit}
&Slim-LLM &75.52 &51.3 &39.29 &62.01 &66.1 &64.07 \\
&Slim-LLM+ &\textbf{76.31} &\textbf{54.07} &\textbf{39.79} &\textbf{63.35} &\textbf{67.14} &\textbf{64.93} \\
\midrule
\multirow{2}{*}{3-bit}
&SpQR &\textbf{80.74} &\textbf{74.75} &\textbf{46.93} &– &61.96 &\textbf{73.32} \\
&Atom &72.47 &49.54 &37.8 &\textbf{66.75} &\textbf{66.99} &60.14 \\
\midrule
\multirow{2}{*}{4-bit}
&SpQR &\textbf{81.01} &\textbf{76.05} &\textbf{47.18} &– &62.5 &72.93 \\
&Atom &78.73 &58.92 &45.82 &\textbf{68.47} &\textbf{77.4} &\textbf{73.09} \\
\midrule
&  & \multicolumn{6}{c}{LLaMA1-65B} \\
\midrule
\multirow{2}{*}{2-bit} 
&Slim-LLM &77.09 &53.72 &40.25 &77.51 &72.05 &\textbf{70.91} \\
&Slim-LLM+ &\textbf{78.06} &\textbf{53.9} &\textbf{41.18} &\textbf{78.33} &\textbf{75.59} &69.99 \\
\midrule  
\multirow{2}{*}{3-bit}  
&SpQR &\textbf{81.18} &\textbf{74.37} &\textbf{45.05} &– &63.54 &\textbf{76.09} \\
&Atom &75.84 &51.43 &41.3 &\textbf{74.07} &\textbf{72.22} &64.33 \\
\midrule  
\multirow{2}{*}{4-bit}  
&SpQR &\textbf{81.56} &\textbf{75.25} &\textbf{46.93} &– &63.76 &\textbf{76.95} \\
&Atom &80.41 &58.12 &45.22 &\textbf{82.02} &\textbf{79.1} &72.53 \\
\midrule
&  & \multicolumn{6}{c}{LLaMA2-7B} \\
\midrule
\multirow{3}{*}{3-bit}  
&BitMod &\textbf{77.53} &– &– &– &\textbf{72.68} &66.22 \\
&FE2-bit &75.9 &66.9 &– &– &51.3 &66.4 \\
&FE2-bit (+0.2\%) &76.1 &\textbf{67.2} &– &– &52 &\textbf{66.8} \\
\midrule  
\multirow{3}{*}{4-bit} 
&BitMod &\textbf{78.45} &– &– &– &\textbf{75.43} &\textbf{68.19} \\
&ResQ &77.9 &\textbf{72.6} &44 &75.3 &74 &66.9 \\
&BlockDialect &75.84 &70.96 &\textbf{44.28} &\textbf{75.44} &73.28 &67.17 \\
\bottomrule
\end{tabular}
\caption{Zero-shot accuracy (\%) on Llama1-7B, Llama1-13, LLaMA1-30B, LLaMA1-65B, and LLaMA2-7B across tasks (PIQA, ARC-e, ARC-c, BoolQ, HellaSwag, Winogrande). Note: "(x\%)" means a sparsity of "x\%"}
\label{tbl:llama1_2_zero_shot}
\end{table}

\subsection{Comparison between optimization techniques of MXPLM frameworks and MXPDNN frameworks}

In the context of mixed-precision quantization for DNNs (MXPDNNs), Rakka et al.\ \cite{rakka2024review} identify categories of optimization techniques for bitwidth allocation, including heuristic, gradient-, Reinforcement Learning (RL)-based, and meta-heuristic (like evolutionary algorithms) methods.

Our survey of MXPLM frameworks reveals a clear convergence toward heuristic-based methods, where local weight or activation sensitivity information guides the precision assignment. The reliance on Hessian information in MXPLMs is driven by scalability and practicality: RL- and meta-heuristic approaches, though powerful in smaller-scale DNN settings, introduce prohibitive search overheads when applied to billion-parameter language models, where each evaluation cycle would require costly fine-tuning or extensive forward passes. Moreover, RL-based allocation strategies often depend on reward signals derived from task-level accuracy, which are challenging to compute in MXPLMs due to the vast training cost and the weak correlation between perplexity or zero-shot metrics and individual layer-level bitwidth choices. 

The divergence of MXPLM frameworks optimization techniques compared to those of DNNs underscores a critical distinction: while the mixed-precision quantization of DNNs has served as a testbed for a variety of search and optimization paradigms, current MXPLM frameworks demand methods that balance accuracy with feasibility at scale, effectively narrowing the optimization techniques toward sensitivity-driven approaches. Looking ahead, it is conceivable that RL- or meta-heuristic-based optimization could be revisited for MXPLMs if future research develops efficient surrogate objectives for perplexity or zero-shot accuracy, lightweight fine-tuning proxies to reduce evaluation cost, and hardware-in-the-loop frameworks capable of rapidly estimating energy-latency trade-offs; such advancements would enable more global search methods to become computationally tractable even at the scale of billion-parameter language models.

\section{Quantization-compatible hardware}
\label{sec:quantization_compatible_hw}

\begin{figure}[t!]
    \centering
    %\vspace{-0.5cm}
    \includegraphics[width=\textwidth]{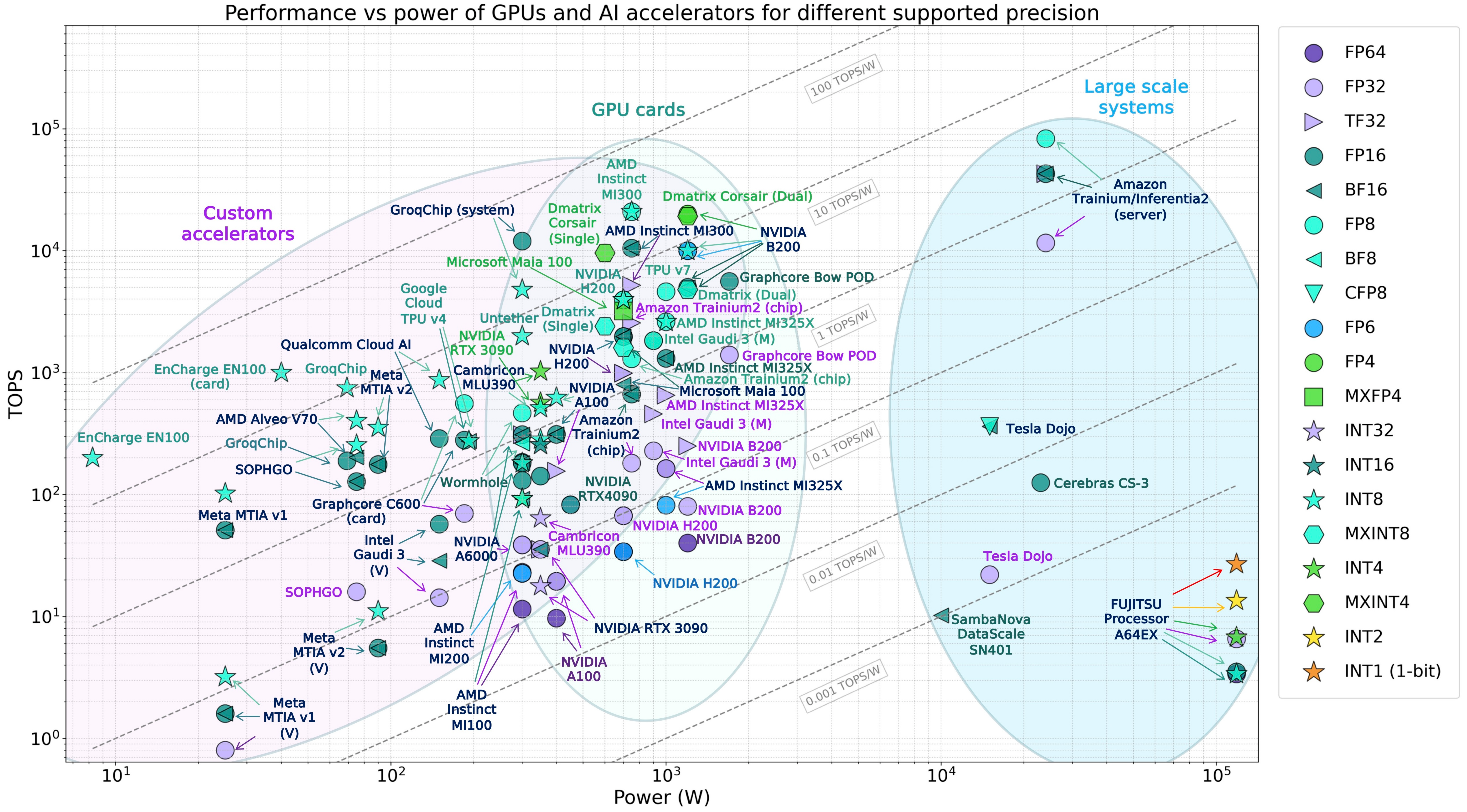}
    %\vspace{-0.6cm}
    \caption{Peak compute performance (in TOPS) versus power consumption for a range of AI hardware platforms, grouped into GPUs, custom AI accelerators, and large-scale systems. Multiple points per device indicate precision-dependent performance based on supported data formats.}
    \label{fig:tops_power}
    %\vspace{-0.5cm}
\end{figure}

Current-generation hardware for LMs includes advanced GPUs, specialized AI accelerators, and large-scale cloud systems. These platforms are designed for high-throughput execution of matrix-intensive workloads using low-precision arithmetic, balancing compute performance (in TFLOPS/TOPS) with memory bandwidth and interconnect efficiency. This section reviews the state-of-the-art in GPU architectures, such as NVIDIA’s Hopper \cite{nvidia_h200,nvidia_h200_datasheet} and Blackwell \cite{nvidia_blackwell_arch}, and AMD’s CDNA3 \cite{amd_mi200,techpowerup_mi200}, emerging custom accelerators, such as Tensor Processing Units (TPUs) \cite{google_tpu_v6e}, Intelligence Processing Units (IPUs) \cite{graphcore_c600}, and other novel AI chips \cite{untether_ai, etched_ai}. It also covers large-scale deployments in cloud infrastructure \cite{amazon_trainium2, futurum_cerebras_cs3}, with a focus on architectural design, supported precision formats, peak performance, and mixed-precision capabilities.
Figure~\ref{fig:tops_power} compares peak compute throughput (TOPS) versus power consumption across three hardware categories: GPUs, custom AI accelerators, and large-scale systems. Some devices include multiple data points corresponding to different supported numerical precisions (e.g., INT8, FP8, BF16), illustrating how performance scales with precision and highlighting the architectural trade-offs in energy efficiency and compute density.

\subsection{Precision casting}
\label{sec:precision_casting}
Before discussing the precision support of various GPUs and accelerators, the concept of precision casting should be introduced, particularly in the context of mixed-precision computation.
Precision casting refers to the conversion of data between different numerical formats (e.g., from FP32 to FP16 or FP8) \cite{micikevicius2022fp8}. During casting, data is reinterpreted and rescaled, which may involve rounding or applying scaling factors to minimize quantization error.
Down-casting reduces the bit width through techniques, such as exponent rebiasing, mantissa truncation, and rounding, often accompanied by additional scaling to preserve dynamic range. In contrast, up-casting restores precision by expanding the exponent and mantissa and reapplying the original scale.
State-of-the-art hardware, such as NVIDIA’s H100, supports automatic precision casting: often low-precision formats like FP8 are used for compute operations, while higher-precision formats (e.g., FP16 or FP32) are used for accumulation and weight updates \cite{nvidia_transformer_engine, luszczek2024batched}.

\subsection{Graphics processing units (GPUs)}
\label{sec:gpus}

Modern GPUs play a central role in LMs training and inference, offering high parallelism, memory bandwidth, and specialized matrix computation units. NVIDIA’s data center GPUs, such as the A100 (Ampere), H100 (Hopper) and B200 (Blackwell), are built for deep learning, utilizing Tensor Cores that execute matrix multiply-accumulate operations. The A100 \cite{nvidia_a100} introduced TF32 precision format (a 19-bit floating-point representation with FP32-range exponents and 10-bit mantissa, optimized for tensor operations) and enhanced support for FP16 and BF16, while the H100 \cite{nvidia_h100} added FP8 via its Transformer Engine \cite{nvidia_hopper_whitepaper, nvidia_hopper_blog}. Hopper’s Streaming Multiprocessors combine traditional compute units (FP64, FP32, INT32) with Tensor Cores that perform FP16/BF16/TF32 matrix operations (with FP32 accumulation) and FP8 operations (with higher-precision accumulation) \cite{nvidia_hopper_blog, nextplatform_hopper}. Two FP8 formats, E5M2 and E4M3, offer flexibility in dynamic range versus precision trade-offs \cite{nvidia_hopper_blog}. The Transformer Engine enables layer-wise mixed-precision and automatic precision casting, performing matrix multiplications in FP8 with accumulation in FP16 or FP32.
The newer H200 \cite{nvidia_h200, nvidia_h200_datasheet} and B200 \cite{nvidia_b200, nvidia_b200_datasheet} extend Hopper’s capabilities. The H200 significantly improves upon the H100, offering enhanced performance for memory-intensive and long-context LM inference. The B200, based on the Blackwell architecture, integrates two reticle-limited dies via chip-to-chip interconnect into a unified GPU \cite{nvidia_blackwell_arch}, and adds support for FP4 for ultra-low-precision inference \cite{wang2025optimizing}. With 192 GB of HBM3e memory and 8 TB/s bandwidth, it is well-suited for large model inference and memory-bound workloads.
Blackwell GPU is also an architecture used in state-of-the-art NVIDIA’s Edge AI solutions, e.g. Jetson Thor which also contains transformer engine and low-precision formats support and multi-instance GPU support for running multiple models simultaneously \cite{nvidia_jetson_thor}.

High-end desktop GPUs like the RTX 3090 \cite{nvidia_rtx3090,hothardware_rtx3090}, RTX 4090 \cite{nvidia_rtx4090,techpowerup_rtx4090}, and RTX A6000 \cite{nvidia_proviz} also support LMs. The RTX 3090 (Ampere) provides FP32 and FP16 support via third-gen Tensor Cores and supports automatic mixed-precision (AMP) through NVIDIA’s software stack (e.g., PyTorch AMP) \cite{nvidia_mixed_precision}. The RTX 4090 (Ada Lovelace) introduces fourth-gen Tensor Cores with support for FP8, BF16, FP16, and TF32. The A6000, also Ampere-based, offers 48 GB of GDDR6 and full Tensor Core support. While not designed for large-scale training of LMs, these GPUs are effective for fine-tuning, medium-scale training, and inference.

Overall, NVIDIA GPUs support a broad precision spectrum, from FP64 to INT4, with peak throughput at lower precisions via Tensor Cores (Fig. \ref{fig:tops_power}). In addition, the most recent solutions, like Blackwell architecture, support a novel custom-based 4-bit floating point format NVFP4 for inference. NVFP4 uses a 4-bit value (E2M1 format: 1 sign, 2 exponent, 1 mantissa) combined with a per-16-values FP8 scale factor (E4M3) and an additional per-tensor FP32 scale \cite{alvarez2025introducing}. Note that sparsity support is not shown in Fig. \ref{fig:tops_power}, but many GPUs. For instance, the A100 GPU can deliver 156 TFLOPS in TF32 precision, and this throughput is effectively doubled to 312 TFLOPS when structured sparsity is utilized.

AMD’s Instinct accelerators, based on the CDNA architecture, are another major platform for LMs \cite{amd_mi100}. The MI200 series (CDNA2) introduced multi-chip modules and Matrix Cores for efficient computation \cite{amd_mi200, techpowerup_mi200}. The MI250X integrates two GPU dies, each with 110 compute units (CUs) and HBM2e memory, connected via Infinity Fabric. It supports FP64, FP16, and BF16 matrix operations, with Matrix Cores operating similarly to NVIDIA’s Tensor Cores.
The latest MI300X (CDNA3) comprises 8 GPU chiplets (XCDs) and four memory/cache dies on a 2.5D interposer \cite{amd_mi300x, sth_mi300x_hotchips}, providing 192 GB of HBM3 suitable for models requiring hundreds of gigabytes of parameters. Its CUs include Matrix Cores supporting FP64, FP32, and low-precision Fused Multiply-Add (FMA) operations \cite{amd_mi300_cdna3_isa}. CDNA3 also introduces support for FP8 and INT8. Like NVIDIA, AMD supports mixed-precision, using low-precision arithmetic (FP8/FP16/BF16) \cite{perez2023training} combined with FP32 accumulation \cite{amd_cdna3_whitepaper}, and supports automated precision casting through libraries such as JAX \cite{amd_jax_mixed_precision}.

\subsection{Emerging custom AI accelerators}

Domain-specific AI accelerators for ML and AI workloads include Google’s Tensor Processing Units (TPUs)  \cite{google_tpu_v6e}, Graphcore’s Intelligence Processing Units (IPUs) \cite{graphcore_c600}, Groq processor \cite{sambanova_sn40l_blog}, Untether’s AI accelerator \cite{untether_ai,untether_ieee}, Etched “Sohu” Transformer ASIC \cite{etched_ai}, and others. The architectural design of domain-specific AI accelerators can be broadly categorized based on their approach to computation and memory into four classes: traditional tensor-core matrix engines, massively parallel many-core processors, near- or in-memory computing architectures, and wafer-scale or ultra-scalable platforms optimized for extreme workloads.

Matrix-compute accelerators such as Google’s TPU \cite{google_tpu_v4, google_tpu_v6e}, Intel’s Gaudi2 and Gaudi3 \cite{intel_gaudi, intel_gaudi3}, AMD’s Alveo V70 \cite{amd_alveo_v70}, Microsoft’s MAIA 100, Meta’s MTIA (v1 and v2) \cite{hostzealot_maia100, microsoft_maia100_azure}, Cambricon \cite{cambricon_ai},  SOPHGO SC7 \cite{sophon_sc7,sophon_sc7_doc}, and Qualcomm’s Cloud AI 100 \cite{qualcomm_cloud_ai100} rely on dense matrix engines, such as Tensor Cores or matrix-multiply units (MXUs) to execute high-throughput linear algebra operations. These accelerators can be further divided into those supporting training, e.g., Google’s TPU, Intel's Gaudi, Microsoft's MAIA 100, Meta's MTIA v2, Cambricon, and SOPHGO's SC7, and inference-optimized accelerators, e.g., AMD's Alveo V70, Meta's MTIA v1, and Qualcomm's Cloud AI 100.

Training-capable accelerators are optimized for forward and backward passes, and contain large memory, and high-precision accumulation, compared to inference-only accelerators.
Google’s TPUs are ASICs built for ML acceleration. TPU v4 incorporates multiple Tensor Cores (MXUs) designed for 128x128 systolic matrix operations supporting BF16 and INT8 precision, and also includes sparse execution cores \cite{google_tpu_v4}. The 6th-generation TPU improves performance with dual MXUs, a vector unit, and a scalar unit per core \cite{google_tpu_v6e}. The most recent 7th-generation TPU Ironwood, containing larger HBM, higher bandwidth and improved chip-to-chip communication, is mostly focusing on efficient inference achieving 4614 TFLOPS per chip for FP8 operation \cite{google_ironwood_inference}.
Intel Habana Gaudi (now Gaudi2) is deployed in AI datacenters (e.g., AWS DL1). It includes Matrix Math Engines (MMEs) for FP16/BF16/INT8 matrix operations \cite{intel_gaudi} and 24 integrated tensor-centric networking links (RoCE over Ethernet) for inter-node scaling; while Gaudi3 adds also FP8 support \cite{intel_gaudi3}. Gaudi accelerates AI computations in a GPU-like manner, offering high internal bandwidth and supporting BF16/FP16 multiplications with FP32 accumulation.
Microsoft Azure MAIA 100, a 5nm chip co-designed with OpenAI, contains high-density matrix engines and supports a wide range of precision types including the Microscaling (MX) formats \cite{microsoft_maia100}. It achieves peak performance of 0.8 PFLOPS in BF16, 1.5 PFLOPS in a 9-bit MX hybrid format, and 3.0 PFLOPS in 6-bit MX \cite{hostzealot_maia100, microsoft_maia100_azure}. MX Format operates as a block floating point (BFP), where a group of numbers shares a common exponent, allowing for more compact data representation and improved processing efficiency. MAIA performs most training in BF16, with 6-/9-bit modes for efficient forward passes or weight updates.
Meta’s MTIA (Meta Training and Inference Accelerator) supports recommender models and LM inference; while MTIA v2 (5nm) supports both training and inference \cite{meta_mtia_nextgen}, combining dense matrix engines with general-purpose cores \cite{techpowerup_meta_mtia_2024}.
Cambricon’s AI accelerators \cite{cambricon_ai} also support a wide range of formats, including FP32, FP16, BF16, INT16, INT8, INT4. The SOPHGO SC7 HP75-I card supports compute with FP32 down to INT8 \cite{sophon_sc7, sophon_sc7_doc} and includes a 24-core ARM cluster and a neural accelerator. 

Inference-only accelerators generally achieve higher throughput per watt compared to training-capable accelerators. AMD’s Alveo V70, built on the Versal AI Edge platform, is designed for power-efficient inference in edge and cloud environments \cite{amd_alveo_v70}. It supports FP32, FP16, BF16, INT8, and INT4, combining programmable logic and AI engines for mixed-precision inference, using lower precisions for compute and higher precision for accumulation. The other inference-focused accelerator is Meta’s first-generation MTIA v1 (7nm chip) \cite{meta_mtia}. Qualcomm’s Cloud AI 100 is also an inference-centric accelerator prioritizing energy efficiency and on-chip memory \cite{qualcomm_cloud_ai100}. It includes 16 AI cores and 144 MB of SRAM; therefore, large models need to be partitioned across chips for large-scale inference.

Many-core and distributed compute architectures, including Graphcore’s IPU \cite{graphcore_c600, graphcore_bowpod16}, Untether AI’s SpeedAI  \cite{untether_ai, untether_ieee}, and Tenstorrent’s Wormhole, utilize thousands of lightweight cores with distributed on-chip memory, offering fine-grained parallelism and flexible execution. 
Graphcore’s IPU is a massively parallel processor with distributed on-chip memory \cite{graphcore_c600, graphcore_bowpod16}. Each chip contains 1,472 IPU-Cores and 900 MB of SRAM, partitioned locally. It supports operation-wise mixed-precision, primarily in FP16 \cite{graphcore_mixed_precision, peng2024evaluating}. Multi-chip systems (IPU-PODs) scale IPUs via high-speed interconnects, with model sharding required due to the memory limits.
Untether AI’s SpeedAI employs a near-memory compute architecture, with more than 1,400 RISC-V cores adjacent to local SRAM banks \cite{untether_ai, untether_ieee}. Each memory bank contains 512 Processing Elements (PEs) supporting INT4/INT8/FP8/BF16, with structured sparsity and mixed-precision arithmetic (e.g., FP8 compute and BF16 accumulation).
In turn, Tenstorrent’s Wormhole processors combine RISC-V CPUs with specialized AI accelerators \cite{tenstorrent_wormhole}. Each chip contains 72 Tensix cores, with scalar units supporting various data types from FP32 to INT8. Mixed-precision is achieved by using low-precision for matrix operations and higher precision on scalar cores. The scalable “Galaxy” mesh interconnect enables efficient multi-chip deployment.

In-memory and near-memory compute accelerators, such as d-Matrix’s Corsair \cite{d_matrix_whitepaper, d_matrix_product}, Etched’s Sohu \cite{etched_ai}, EnCharge AI's EN100 \cite{encharge_en100}, Hailo-8 \cite{hailo8, hailo8_whitepaper}, and again Untether AI’s SpeedAI  \cite{untether_ai, untether_ieee}, reduce data movement by placing compute units directly within or adjacent to memory banks, which is especially advantageous for memory-bound inference tasks.
d-Matrix's Corsair targets low-latency LM inference using a digital in-memory compute (DIMC) system \cite{d_matrix_whitepaper, d_matrix_product}. Each Corsair's card contains 8 chiplets with 512 MB SRAM and utilizes a custom BFP12 (Block Floating Point 12-bit) format, where groups share an exponent. It performs INT8/BFP12 matrix operations with higher-precision accumulation.
Etched’s Sohu Transformer ASIC implements the transformer pipeline in hardware \cite{etched_ai}. Instead of generic MAC arrays, it uses dedicated “transformer cores” to sequentially process tokens through all layers. With 144 GB of HBM3E, performance is bandwidth-bound, and the entire model can reside in memory.
Hailo-8 is an efficient edge AI chip using a structure-defined dataflow architecture \cite{hailo8, hailo8_whitepaper}. It performs inference entirely within on-chip SRAM using INT8/INT4, avoiding costly DRAM access, and supports small-scale LM models.
The EN100 from EnCharge AI is an analog in-memory computing–based accelerator that supports 8-bit operations, delivering 200 TOPS per module at only 8.5 W and up to 1,000 TOPS per PCIe card (with four processing units) at 40 W, demonstrating exceptionally low power consumption. Each PCIe card integrates 128 GB of LPDDR memory and provides a memory bandwidth of 272 GB/s \cite{encharge_en100}.

Finally, wafer-scale and ultra-scalable designs, such as Groq’s Tensor Streaming Processor (TSP)  \cite{groq_chip} and Cerebras’s CS-2 and CS-3 wafer-scale engines \cite{futurum_cerebras_cs3, cerebras_cs2_datasheet} achieve extreme compute and memory scalability by spanning large die areas or statically pipelining entire model graphs across chip-scale dataflow fabrics.
Groq’s Tensor Streaming Processor (TSP) unrolls neural network graphs into a pipelined loop with static scheduling via compiler \cite{groq_chip}. Its single-core design achieves high throughput and low latency but has limited on-chip memory, requiring model distribution across chips \cite{sambanova_sn40l_blog}. Cerebras’s CS-2 and CS-3 wafer-scale engines are built as a single large “wafer-scale” chip. In terms of energy efficiency, it is closer to a large-scale system; therefore, it is discussed below \cite{futurum_cerebras_cs3, cerebras_cs2_datasheet}.

Across accelerators, mixed-precision arithmetic is implemented using similar principles, e.g., FP16/BF16 for training, INT8 or FP8 for inference, and hybrid schemes (such as block floating point) to balance accuracy and performance. Mixed-precision is often operation-wise, with lower precision for matrix multiplications and higher precision for sensitive operations (e.g., softmax, normalization).
Most accelerators rely on software–hardware co-design, with compiler/runtime support crucial for enabling precision casting and operation scheduling.
While GPUs and custom AI accelerators dominate LM training, some work explores LM inference on CPUs \cite{shen2023efficient, na2024understanding}. For example, Apple’s M3 Ultra is marketed as capable of running large language models on device, potentially leveraging its Neural Engine for inference support \cite{apple_m3_ultra}, though practical deployment and performance details remain unclear.

\subsection{Large-scale cloud-based systems}

Cloud-based AI infrastructure can be built using either GPU-based clusters or custom accelerator clusters. For example, Microsoft Azure uses clusters composed of tens of thousands of NVIDIA A100 GPUs interconnected via high-speed networking for large-scale training and inference workloads \cite{nvidia_azure_a100_v4}. NVIDIA also provides its own reference design for GPU superclusters under the DGX SuperPOD architecture \cite{nvidia_hopper_blog}. In addition, recently announced Microsoft Azure ND MI300X v5 Instances are based on AMD Instinct MI300X GPUs clusters \cite{azure_ndmi300xv5}.

In contrast, several cloud platforms deploy clusters based on custom accelerators. Google’s TPUv4 Pods comprise up to 4,096 TPU chips connected in a high-radix torus mesh network with optical circuit switching \cite{jouppi2023tpu}. Each Pod can deliver up to 1.1 exaFLOPS for BF16 or INT8 operations \cite{google_tpu_v4}. The most recent 7th-generation TPU Ironwood achieves 42.5 ExaFLOP per 9,216-chip pod for FP8 operation \cite{google_ironwood_inference}. Similarly, GroqRack is Groq’s cluster-scale architecture that integrates multiple GroqNodes for high-throughput inference and low-latency execution \cite{groq_rack, groq_scalability}.

Amazon Web Services (AWS) uses its own custom chips: Trainium for training and Inferentia for inference \cite{amazon_trainium2, semianalysis_trainium2, aws_inferentia}. Trainium supports a wide range of formats, including FP32, TF32, FP16, BF16, INT8, and a configurable FP8 (cFP8), where exponent and mantissa bits can be tuned for performance \cite{aws_neuron_dtypes}. These accelerators are organized into NeuronCores, with Trainium2 supporting 64-chip UltraClusters interconnected via NeuronLink, a mesh-style fabric for multi-node scaling. The NeuronCore-v2 ScalarEngine enables mixed-precision execution, performing arithmetic in BF16 and handling sensitive operations (e.g., LayerNorm, Softmax, accumulation) in FP32 \cite{aws_neuroncore_v2}.

SambaNova’s SN40L chip uses a Reconfigurable Dataflow Unit (RDU) to map neural networks onto a spatial array of functional units \cite{sambanova_datascale_sn30}. Each SN40L contains two AI cores, on-package HBM for fast access, and direct DDR connections for capacity \cite{sambanova_sn40l_blog}. A full system contains 16 SN40L chips. The RDU enables model-specific hardware reconfiguration, avoiding instruction overhead and optimizing compute patterns for GEMMs, convolutions, or attention. This chip supports FP32, BF16, and INT8, with mixed-precision modes and a two-tier memory hierarchy managed by runtime software \cite{prabhakar2024sambanova}.

Cerebras takes a radically different approach with its Wafer-Scale Engine (WSE). The second-generation WSE-2, used in the CS-2 system, is a full 7nm wafer with 850,000 AI cores and 40 GB of on-wafer SRAM \cite{cerebras_cs2_datasheet}. Its massive size (approximately 46,225 mm$^2$) eliminates off-chip communication, providing exceptional memory and inter-core bandwidth \cite{nextplatform_cerebras_heterogenous}. The WSE supports FP32, FP16, and BF16, and for larger models, implements weight streaming from external memory. The Cerebras CS-3 adds support for CB16, a custom 16-bit format with BF16-like 8 exponent bits for wide dynamic range and custom mantissa rounding or scaling optimized for AI kernels \cite{futurum_cerebras_cs3, cerebras_dynamic_loss_scaling}.

Tesla’s Dojo supercomputer is built around the custom Dojo D1 chip, designed for internal computer vision and AI workloads \cite{tesla_dojo, tesla_dojo_ieee}. Each D1 die has 354 training nodes arranged in a 2D mesh, all of which use local SRAM without DRAM. Chips are tiled together into larger “ExaPod” systems. Dojo performs training in BF16 or Tesla’s CFloat8, a custom 8-bit floating-point format allowing tunable exponent and mantissa allocation \cite{tesla_dojo_relayto}. Dojo achieves scalability through fast inter-chip communication and pipeline parallelism across chips rather than large per-chip memory.

Fujitsu’s A64FX CPU, used in the Fugaku supercomputer, demonstrates that general-purpose CPUs can still play a role in AI workloads \cite{fujitsu_a64fx, top500_a64fx}. A64FX features 48 ARM cores, 512-bit Scalable Vector Extension (SVE) vector units, and 32 GB HBM2, supporting FP16/INT8 arithmetic. Although slower than GPUs, CPUs like A64FX offer flexibility, large memory capacity, and the ability to run models with long sequence lengths or high parameter counts without compiler modifications.

% \newpage

\section{Prospects and future directions}
\label{sec:future_directions}
{
This section discusses prospects and future directions for fully quantized language models, spanning activation and KV-cache compression, attention/softmax quantization, training-time mixed precision, hardware and software enablement, and emerging low-bit numeric formats.
}

\subsection{Towards fully quantized LMs}
The majority of the mixed-precision methods discussed above focus on weight quantization, with a few also applying mixed-precision to activations \cite{zhao2024atom,cheng2023dataflow, jang2025blockdialect, saxena2024resq, ashkboos2023quik}. Quantizing weights is generally easier, as it can be performed offline with pre-computed scales stored in memory, whereas activation quantization often requires \textit{dynamic} (on-the-fly) scaling to avoid large accuracy drops in LLMs. This dynamic scaling introduces runtime overhead due to extra computations and data, reducing the overall efficiency benefits of quantization.

As both sequence length and model size increase, the memory footprint of the key–value (KV) cache in attention grows proportionally and can become a major bottleneck~\cite{wu2024layercondensedkvcacheefficient, Kwon2023PagedAttention}. In long-context scenarios, this growth can be quadratic in sequence length, making compression essential. Several works in our survey quantize the KV cache, either with uniform~\cite{zeng2024abq} or mixed-precision schemes~\cite{yang2024no,cheng2023dataflow,jang2025blockdialect}, reducing its size while aiming to minimize the resulting accuracy loss. Efficient KV-cache quantization is especially important for deployment in constrained memory settings, where the cache can otherwise dominate total inference memory.

Even when weights, activations, and the KV cache are quantized, many components in transformer inference pipelines remain in higher precision. These include normalization layers, non-linear activations, residual connections, the softmax function, and the query–key–value matrix multiplications in self-attention. Notably, even if the KV cache itself is stored in low precision, it is often dequantized back to higher precision before attention matmuls~\cite{lin2024qserve}. Both uniform~\cite{Ashkboos2025QuaRot} and mixed-precision~\cite{cheng2023dataflow,jang2025blockdialect} approaches have explored fully quantizing the self-attention matmuls. 

Recent works have started to address the challenge of low-bit softmax~\cite{shkolnik2024Exaq, hu2024illm, vasyltsov2021efficientsoftmaxapproximationdeep,  lin2023fqvit}, which remains one of the key obstacles to achieving truly fully quantized LMs. This is particularly relevant for edge deployments, where the non-linear nature and relatively high computational cost of softmax can become a latency bottleneck. A combination of robust softmax quantization and fully quantized attention could close one of the last major gaps toward end-to-end low-bit inference in LMs.

Reasoning Language Models (RLMs) decompose problems into intermediate steps; explicit long chain-of-thought (CoT) when elicited and implicit reasoning when not, enabling strong performance on math, logic, code, and multi-hop QA \cite{chen2025towards,li2025implicit}. Yet these behaviors are compute-hungry: long CoT traces and large parameter counts inflate memory and latency. Quantization addresses this by compressing parameters - and sometimes activations and the KV cache - into low-bit formats to cut memory and boost throughput with minimal architectural change. Recent reasoning-centric studies sharpen the picture: broad sweeps over weight/activation/KV settings find that W8A8 and W4A16 are typically near-lossless for CoT-style tasks, while pushing to $\le$ 3-bit induces disproportionate errors on harder multi-step problems \cite{li2025quantMeets,liu2025quantHurts}. Complementarily, evidence on smaller models suggests that careful training and post-training compression (including quantization) can deliver competitive reasoning, challenging “scale-only” assumptions \cite{srivastava2025towards}.

Looking ahead, the community would benefit from a standardized, CoT-sensitive evaluation protocol for quantized RLMs. Rather than ad-hoc leaderboards, future work could anchor comparisons to BF16/FP16 baselines while jointly tracking final-answer accuracy (e.g., GSM8K, MATH/AIME, BBH, MMLU) and process signals (tokens-to-answer, CoT length, self-consistency dispersion, refusal rate, and error taxonomies). Such a protocol would enable principled accuracy–efficiency frontiers across W8A8, W4A16, W4A8, and W4A4, and clarify where performance bends under compression, especially on competition-level math where recent studies suggest degradation accelerates below ~4-bit \cite{liu2025quantHurts,li2025quantMeets}. Future directions also include instance-level agreement analyses (how often quantized models follow the same intermediate steps as full precision), sensitivity to decoding hyperparameters, and robustness under dataset shift, yielding a more causal picture of how quantization perturbs multi-step reasoning.

Automated mixed-precision is a promising direction for better accuracy-efficiency tradeoff through learning per-layer/per-head bit-widths from sensitivity profiles. Fragile computations would retain higher precision while the rest go low-bit. This could generalize current recipes, weight-only W4 with saliency protection (AWQ), full-stack W8A8 stabilized by outlier handling (SmoothQuant) and 4-bit backbones adapted via higher precision LoRA (QLoRA), into data-driven allocation schemes \cite{lin2023awq,xiao2022smoothquant,dettmers2023qlora}. For long CoT, future systems may adopt adaptive KV-cache precision that varies across layers, spans, or tokens, building on KVQuant baselines and recent advances like PM-KVQ and OTT to compress more aggressively while preserving accuracy \cite{hooper2024kvquant,liu2025pmkvq,su2025ott}. Additional directions include dynamic precision during decoding (e.g., temporarily elevating precision for difficult steps), precision-aware training objectives that encourage quantization-robust internal states, and lightweight post-quantization finetunes targeted at the specific failure modes surfaced by the proposed evaluation suite.

\subsection{Microscaling format}
An important advancement in quantization and low-precision arithmetic is the emergence of the Microscaling (MX) formats~\cite{rouhani2023microscalingdataformatsdeep} that has been standardized by the Open Compute Project (OCP)~\cite{2023OcpMicroscaling}. Microscaling formats introduce a block floating-point or integer data representation that couples a shared scaling exponent per block with low-bit payloads, such as MXFP8, MXFP6, MXFP4, and MXINT8, to balance efficiency, accuracy, and deployment convenience. An MX block encodes a vector of \(k\) values using a single shared \textit{power-of-two} scale \(X\) and \(k\) quantized elements \(P_i\), with each reconstructed value given by:
\begin{equation}
v_i = X \cdot P_i,\quad i=1,\dots,k
\end{equation}
This fine-grained scaling enables robust quantization down to sub-8-bit regimes while preserving model quality, even when applied with minimal changes to training workflows~\cite{rouhani2023microscalingdataformatsdeep}.

These MX formats are a natural fit for mixed strategies in LM inference. For example, \cite{cheng2023dataflow} introduced a framework for MX-integer mixed precision, while more recent work from MicroMix~\cite{liu2025micromix} exploits MX’s flexibility by dynamically assigning tensor channels to different precisions, MXFP4, MXFP6, or MXFP8, based on quantization error thresholds. This enables high-speed mixed-precision computation while maintaining output in BF16.

Looking ahead, Microscaling is rapidly emerging as the de-facto standard for low-precision arithmetic in large-scale language models, driven by its strong support from both hardware vendors and deep learning frameworks. From the hardware side, Blackwell~\cite{nvidia_blackwell_mxfp_gemm} architecture natively accelerates MXFP4/8 GEMMs with tensor-core parity, and AMD’s CDNA\texttrademark~4 \cite{amd_cdna3_whitepaper} introduces dedicated instructions for MX arithmetic. From the software side, Microsoft has released experimental support via its \texttt{microxcaling} repository~\cite{ms_microxaling}, 
PyTorch has integrated MX into its quantization toolkit through \texttt{torchao}~\cite{torchao_mx}, 
and third-party projects such as \texttt{torchmx}~\cite{torchmx} provide additional research-grade implementations. Taken together, these developments position Microscaling as the likely gold standard for both training and inference of next-generation language models with mature hardware and software support.  

\subsection{Mixed-precision training for LMs}
\label{sec:mixed-precision-training}

{
Mixed-precision training (MPT) is now the default path to scale LLM training on NVIDIA GPUs, delivering large memory and throughput gains with minimal accuracy loss. Tooling such as NVIDIA Apex\footnote{A PyTorch Extension for mixed precision and distributed training} and native Automatic Mixed Precision (AMP) in PyTorch/TensorFlow automatically casts eligible operations to lower precision, applies dynamic loss scaling, and keeps numerically sensitive paths in higher precision—so models can train end-to-end in FP16 or BF16 with only small code changes~\cite{micikevicius2018mixedprecisiontraining}. Beyond op casting, modern stacks also vary gradient precision by statistics, while reserving FP32 for critical updates, accelerating both forward and backward passes. On recent hardware, NVIDIA’s Transformer Engine extends automatic mixed precision to FP8, selecting FP8 or FP16 per layer and handling re-casting to preserve convergence; emerging formats such as NVFP4 further push the efficiency frontier~\cite{deepseek2024,nvfp42025}. Existing tools such as NVIDIA Apex and pytorch/Tensorflow AMP do not have the full mixed precision support that is performed in afromentioned mixed precision frameworks in addition does not support low precision, including FP8/4 or INT8/4. NVIDIA Apex and AMP in PyTorch/TensorFlow do not provide full mixed-precision coverage like the frameworks discussed above, and they lack native support for ultra-low precisions (e.g., FP8/FP4 or INT8/INT4). In this section, we explore the challenges and possible research directions.

\subsubsection{Challenges in mixed-precision training}
Here, we outline the challenges of mixed precision training of language models. 

\textbf{Numerical stability and dynamic range management:} The most significant challenge in MPT is maintaining numerical stability when using lower-precision formats~\cite{nvfp42025,deepseek2024}. The limited dynamic range of reduced-precision formats causes gradient values to frequently fall outside the representable range, leading to underflow where small gradients are rounded to zero~\cite{nvfp42025}. This gradient underflow prevents proper weight updates and can cause training divergence, particularly in deeper networks where gradients diminish through multiple layers~\cite{nvidia_mixed_precision,deepseek2024}. The challenge intensifies as precision decreases from FP16 to FP8 and further to FP4, requiring progressively more sophisticated techniques to maintain stability~\cite{nvfp42025,deepseek2024,lee2024fp8}. In parallel, activations, weights, and gradients in transformer-based language models exhibit significant outliers that disproportionately affect low-precision training~\cite{deepseek2024,nvfp42025}. Addressing these outliers necessitates fine-grained quantization strategies to extend the effective dynamic range of narrow formats~\cite{deepseek2024}. DeepSeek-V3 leverages tile-wise grouping ($1 \times N_c$ elements) or block-wise grouping ($N_c \times N_c$ elements) for FP8 training, while NVFP4 uses 16-element blocks with two-level scaling (per-block E4M3 scales combined with per-tensor FP32 scales) to better capture local dynamic range~\cite{deepseek2024,nvfp42025}.

\textbf{Bit-allocation search :} 
Bit-allocation search is computationally expensive because the configuration space grows exponentially with both network depth and the number of candidate precisions per layer. Even “relaxed” differentiable search methods still require large-scale proxy training and bilevel optimization to evaluate allocations at scale~\cite{rethinking_diff_search_2020}.
Heuristic strategies, such as greedy layer ranking or gradient-free metaheuristics, can narrow the search, but they still demand many calibrated evaluations or short fine-tuning runs to estimate accuracy and latency. This quickly becomes prohibitive for LLMs or long calibration sets~\cite{greedy_mixed_precision_ptq,gradient_free_bit_alloc}.

Hardware constraints make the problem even more complex. Legal tile sizes, accumulation rules, and scale formats (e.g., FP8 tile/block groupings or NVFP4’s 16-element two-level scaling) invalidate many nominal precision assignments, requiring hardware-aware feasibility checks that further inflate the search cost~\cite{nvfp42025,deepseek2024}.
Reliable validation also depends on hardware-accurate kernels and end-to-end timing. For training-time mixed precision, consistency must be maintained across forward, backward, and optimizer paths, often necessitating partial retraining or repeated calibration passes~\cite{mixed_precision_survey_2024,rethinking_diff_search_2020}.

Sensitivity-guided methods can reduce exploration but introduce their own costs: computing Hessian traces, saliency metrics, or activation/KV sensitivities at LLM scale is expensive, especially when repeated across layers, blocks, and tensor granularities~\cite{hawq_v2}.
Finally, bit allocations are not static during training. Configurations that work early can degrade during learning-rate decay or curriculum transitions, prompting reallocation or precision upshifts—effectively embedding the search process within training itself~\cite{nvfp42025}.

\textbf{Layer-wise and component sensitivity:} Different architectural components exhibit varying sensitivity to precision reduction~\cite{nvfp42025,deepseek2024}. Embedding layers, output projection heads, attention mechanisms (including softmax and attention score computations), normalization layers, and gating modules in MoE architectures consistently require higher precision~\cite{deepseek2024,nvfp42025}. For NVFP4, keeping the final 15\% of linear layers in BF16 proves necessary for convergence, with sensitivity concentrated in the final blocks of the network~\cite{nvfp42025}. DeepSeek-V3 similarly maintains the first and last few blocks in BF16 while executing the majority of GEMMs in FP8~\cite{deepseek2024}.

\textbf{Training stability across scale and duration:} Maintaining stability becomes increasingly difficult as both model size and training duration grow~\cite{nvfp42025}. NVFP4 training of a 12B-parameter model on 10 trillion tokens, the longest publicly documented FP4 run—showed that relative loss error stays below 1\% during stable phases but widens to around 1.5\% during learning-rate decay~\cite{nvfp42025}.
Techniques that prove essential at scale, such as Random Hadamard transforms, 2D weight scaling, or stochastic rounding, may have little measurable effect on smaller models or shorter training runs, highlighting the scale-dependent nature of precision sensitivity~\cite{nvfp42025}.

\subsubsection{Future research directions}
Below, we outline key directions to address the challenges of mixed-precision training.

\noindent\textbf{Knowledge transfer from pretrained models through leveraging inference quantization insights:}
Mixed-precision training can be accelerated by transferring precision sensitivity patterns learned from pretrained models to guide precision allocation in new runs~\cite{quantization_transfer_2022,sensitivity_aware_ptq,best_source_selection}.
Post-training quantization (PTQ) methods such as GPTQ, APTQ, and SliM-LLM estimate layer sensitivity using Hessian-based metrics. They consistently identify attention projections (especially key matrices), the first and last transformer blocks, embeddings, and normalization layers as requiring higher precision across architectures~\cite{sensitivity_aware_ptq,hessian_mixed_precision,deepseek2024,nvfp42025}. This prior knowledge can initialize mixed-precision policies by assigning higher precision formats, such as FP8 or BF16, to sensitive components and lower precision formats, such as FP4, to more robust ones. These allocations can then be refined during training through periodic Hessian-based re-evaluations on checkpoints~\cite{quantization_transfer_2022,sensitivity_aware_ptq,hessian_mixed_precision,deepseek2024,nvfp42025}.

% Mixed-precision training can be accelerated by transferring layer- and component-level precision sensitivity learned from pretrained models to guide precision allocation in new runs~\cite{quantization_transfer_2022,sensitivity_aware_ptq,best_source_selection}. PTQ methods, such as GPTQ, APTQ, and SliM-LLM, typically use Hessian-based sensitivity metrics to rank layers and consistently flag attention projections (particularly key matrices), early and final transformer blocks, embeddings, and normalization layers as requiring higher precision across architectures~\cite{sensitivity_aware_ptq,hessian_mixed_precision,deepseek2024,nvfp42025}. This prior knowledge can initialize mixed-precision policies (e.g., assign FP8/BF16 to sensitive components and FP4 elsewhere), which can then be adaptively refined during training via periodic Hessian-based re-evaluations on checkpoints~\cite{quantization_transfer_2022,sensitivity_aware_ptq,hessian_mixed_precision,deepseek2024,nvfp42025}.

First-order gradients from pretrained checkpoints provide efficient sensitivity maps that can guide precision allocation using gradient magnitude, $|\nabla_{\mathbf{w}}\mathcal{L}|$, even when gradients are non-zero at convergence~\cite{gradient_aware_quant,sensitivity_aware_ptq}.
Community sensitivity priors capture common architectural patterns: MoE gating, softmax, and routing operations need higher precision; most feed-forward GEMMs tolerate FP8; and the final 15\% of layers are often more fragile~\cite{deepseek2024,nvfp42025,hessian_mixed_precision,best_source_selection}. These priors can serve as templates that are refined online through loss or Hessian-based feedback. A practical procedure would be to initialize from small-data sensitivity maps, assign the top 20–30\% most sensitive parameters to FP8 or BF16, reassess periodically, and gradually lower precision for robust regions as training stabilizes. The resulting precision patterns can be transferred across model sizes within the same architecture family~\cite{nvfp42025,hessian_mixed_precision,sensitivity_aware_ptq,best_source_selection,quantization_transfer_2022}.

% Complementarily, first-order gradients from pretrained checkpoints yield efficient sensitivity maps, despite non-zero gradients at convergence, allowing parameter ranking by $|\nabla_{\mathbf{w}}\mathcal{L}|$ to guide allocation~\cite{gradient_aware_quant,sensitivity_aware_ptq}. Community “sensitivity priors” can capture architecture-level regularities (e.g., MoE gating/softmax/routing need higher precision; most FFN GEMMs tolerate FP8; the final ~15\% of layers are often fragile) and serve as templates refined online via loss/Hessian signals~\cite{deepseek2024,nvfp42025,hessian_mixed_precision,best_source_selection}. Practically, initialize from small-data sensitivity maps (assign top 20–30\% most sensitive to FP8/BF16), reassess at intervals, progressively lower precision for robust parts as training stabilizes, and transfer patterns across scales within an architecture family~\cite{nvfp42025,hessian_mixed_precision,sensitivity_aware_ptq,best_source_selection,quantization_transfer_2022}.

\textbf{Selective precision transitions during training:} NVFP4 experiments demonstrate that transitioning from FP4 to higher precision during learning rate decay phases can close the loss gap with higher-precision baselines~\cite{nvfp42025}. This suggests a broader research direction: learning optimal precision schedules that vary not just between training phases but potentially between model components, batch positions, or training dynamics~\cite{nvfp42025}. Meta-learning approaches could discover when to upshift precision based on gradient statistics, loss surface sharpness, or convergence indicators, maximizing efficiency while preserving final model quality~\cite{deepseek2024}.

\textbf{Automated mixed-precision search and hyperparameter tuning:} Neural architecture search (NAS) techniques could be adapted to discover optimal mixed-precision configurations across the training pipeline, such as zero- and one-shot NAS \cite{li2024zero}. Recent work, such as AMQ (Automated Mixed-precision Quantization)\cite{lee2025amq} demonstrates the potential of using quality predictors to efficiently navigate the mixed-precision search space for weight-only quantization, while FLIQS \cite{dotzel2023fliqs} enables one-shot mixed-precision search across both floating-point and integer formats. This includes searching over precision assignments for forward/backward passes, identifying which operations benefit most from higher precision, and determining layer-specific quantization granularity. The search space could extend to quantization-specific hyperparameters like block sizes, scale factor formats, and rounding strategies, potentially discovering configurations superior to manually designed approaches~\cite{nvfp42025,deepseek2024}.

\textbf{Theoretical understanding of low-precision training dynamics:} Despite empirical success, theoretical understanding of why low-precision training converges remains limited~\cite{nvfp42025}. Research should investigate how quantization noise affects gradient descent trajectories, loss landscape geometry, and generalization properties~\cite{reduced_precision_stability}. Understanding the conditions under which stochastic rounding provides sufficient gradient estimation, or when chain rule violations from inconsistent quantization remain acceptable, would inform better training methodologies~\cite{nvfp42025}. By clarifying how low-precision updates behave, we can design mixed-precision schemes that allocate bits where they matter most and reduce sensitivity to outliers and dynamic-range limits.

\subsection{Hardware and software support}
\label{hardwareInsights}

One of the key areas where hardware support is lagging behind the rapid development of LM models is in low-precision arithmetic. Only a few systems currently support INT4 or FP4 formats, and FP4 remains non-standardized. In contrast, NVIDIA has standardized the FP8 format with E5M2 and E4M3 variants \cite{nvidia_hopper_blog}, enabling the development of robust software stacks, improved portability, and greater adoption. More recently, NVIDIA’s Blackwell architecture added support for the 4-bit floating-point inference format NVFP4 \cite{alvarez2025introducing}. However, the support for ultra-low-precision formats, such as INT2 and INT1, is missing across state-of-the-art accelerators, except the Fujitsu A64FX processor \cite{fujitsu_a64fx,top500_a64fx}. Efficient use of these extremely low-precision formats requires not only hardware support but also co-designed software and training algorithms that ensure stable convergence under reduced precision.

Another emerging insight from current accelerator designs is the lack of fine-grained precision control. While many platforms support mixed-precision using lower precision (e.g., FP8 or INT8) for matrix multiplications and higher precision (e.g., FP16 or FP32) for accumulation, irregular or intra-matrix precision variation is not supported. Allowing operations within a kernel or matrix to use different precisions could unlock further efficiency gains in LM training and inference, but would require significant hardware changes and sophisticated scheduling logic. Some recent academic works \cite{tahmasebi2024flexibit,sharma2018bit,rakka2024bf} have proposed hardware accelerators for fine-grained mixed-precision, but these designs are often evaluated in isolation, without integration into broader CPU/GPU-based systems. As a result, significant challenges remain in translating such research-proposed hardware from the academic domain into deployable commercial-grade industrial hardware.

Sparsity support is another critical challenge. Structured pruning is primarily supported by NVIDIA and AMD GPUs and Google’s TPU SparseCores. For example, NVIDIA’s support for structured sparsity in the A100 is limited to specific patterns \cite{nvidia_a100}. Custom accelerators like Untether AI and d-Matrix support fine-grained sparse computation via pattern-based zero skipping \cite{untether_ai, untether_ieee}. However, unstructured or dynamic sparsity, which is more representative of real model pruning, presents irregular memory access patterns that most hardware struggles to handle efficiently. Combining sparsity with mixed-precision, e.g. using low-precision formats for dense regions and skipping zero-valued blocks, offers further efficiency potential but requires fine-grained control and coordinated software–hardware co-design to avoid underutilization of compute units.

Another open problem is integrating quantization into the hardware pipeline. Currently, quantization is handled as a preprocessing step during model export or compilation. A more advanced design would enable hardware to dynamically determine quantization scales during execution, especially embedding quantization-aware calibration within the compute pipeline. 
Incorporating more quantization-aware intelligence in hardware could reduce the gap between INT4/INT8 and full precision inference quality, but commercial accelerators have not yet offered this capability.

From a broader hardware perspective, a consistent limitation is the imbalance between memory bandwidth and compute throughput. While GPUs and AI accelerators now reach multi-petaflop compute performance, memory capacity and bandwidth (typically in the 80–200 GB range) have not kept pace. This limits performance for large-scale LMs with hundreds of billions of parameters \cite{cerebras_cs3_vs_b200}, as high compute throughput is only effective when the memory subsystem can feed it fast enough. Solutions such as larger caches, activation reuse, and gradient compression are becoming essential to improve memory efficiency.
Finally, power and thermal limits are growing concerns. The compute demands of large LMs often require merging multiple AI accelerators into massive clusters, resulting in extremely high power consumption. Achieving higher efficiency for training and inference increasingly depends on improving energy efficiency per operation, which will require innovations in both circuit design and system-level thermal management.

\subsubsection{Hardware-aware MXPLMs}
\label{subsubsec:hardware_aware}

For DNNs, hardware-aware methods have been investigated that integrate device constraints such as memory bandwidth, latency, and energy directly into the bit-allocation optimization process \cite{rakka2024review}. These approaches often rely on hardware-in-the-loop evaluation or hardware proxies to guide bitwidth allocation, enabling quantization decisions that are not only accuracy-preserving but also resource-efficient on the target accelerator.

While effective for DNNs, our survey of MXPLM frameworks reveals that very few works have extended this line of thinking to LMs. Among the surveyed methods, only MASE and LLM-PQ explicitly incorporate hardware-awareness in their formulation. The majority of MXPLM frameworks remain accuracy-centric, optimizing perplexity while assuming hardware costs can be addressed post hoc. This gap stems from the unique challenges of LMs: the scale of billions of parameters renders hardware-in-the-loop evaluation prohibitively expensive, simulation environments for LMs are less standardized than for DNNs, and the latency/energy profile of transformer blocks depends strongly on sequence length, cache management, and memory bandwidth all of which complicate proxy modeling.

Looking forward, enabling hardware-aware MXPLMs will require the development of lightweight and accurate hardware proxies tailored to LM workloads, such as analytic cost models for transformer inference (based on energy, latency, area, etc.), or learned surrogates trained on representative workloads. Such methods would allow the quantization search to jointly optimize perplexity and hardware cost, bridging the current disconnect between accuracy-driven MXPLM design and deployability on real hardware platforms.

\subsubsection{Software support of MXPLMs}

Modern AI compiler stacks expose first-class quantization flows, though scope and maturity differ by project: in PyTorch, the quantization axis is PyTorch AO (“TorchAO”), which provides end-to-end APIs for post-training and training-time quantization and integrates with PyTorch’s compilation/export toolchain \cite{torchao_mx,torchao-api}. NVIDIA TensorRT offers both calibration-driven INT8 and explicit quantization via Q/DQ graphs and, at the kernel/runtime level, enumerates low-precision types including FP8, INT4 (weight-only), and FP4 on supported hardware; at very low precision, production support is strongest in NVIDIA’s stack, where current releases document FP8 and 4-bit variants that trade precision for bandwidth/throughput, with quantized values represented in INT8, FP8 (E4M3), INT4, or FP4 (E2M1) and explicit rules for de/quantization and range handling \cite{tensorrt-quantized,tensorrt-capabilities}. Apache TVM compiles both pre-quantized graphs exported from upstream frameworks and TVM-side quantized graphs via its QNN dialect, whose Relay operators include \texttt{quantize}, \texttt{dequantize}, and \texttt{requantize}; its design keeps integer domains first-class in QNN and frontends, enabling import of pre-quantized models while providing TVM-side passes and kernels to execute these graphs portably across targets \cite{tvm-qnn,tvm-prequant}. Within the Modular ecosystem, MAX (backed by Mojo kernels) supports loading and executing pre-quantized models and provides a quantization module that specifies the admissible encodings and tunable configuration parameters for deployment \cite{max-quant-guide,max-quant-api,mojo-quant}. As a summary, TorchAO serves as the framework-level quantizer, TVM as a portable compiler for quantized IR, TensorRT as a hardware-tuned deployer with aggressive low-precision coverage, and MAX/Mojo as a runtime and kernel toolchain that consumes quantized graphs.

Mixed-precision execution is handled explicitly in deployment compilers. TensorRT supports mixed-precision inference, combining FP32, TF32, BF16, and FP16 within the same network. Users can maintain numerically sensitive operations in higher precision while using reduced precision for less sensitive operations. These mechanisms coexist with explicit quantization graphs and extend to FP8/INT4/FP4 where kernels exist and constraints are met \cite{tensorrt-accuracy,tensorrt-bestpractices,tensorrt-capabilities}. TVM provides automated mixed-precision transformations that lower FP32 graphs to FP16 (optionally retaining FP32 accumulation), and can be composed with quantization workflows or applied when importing mixed-precision graphs from external frameworks \cite{tvm-mixed-adreno}.
PyTorch formalizes mixed floating point execution via Automatic Mixed Precision (AMP), which selects dtypes of op level (e.g., BF16 / FP16 for tensor core friendly ops and FP32 for critical reductions) to improve throughput and memory efficiency while maintaining model quality; this is the standard mechanism for heterogeneous precision during training and inference in the PyTorch stack and is routinely paired with export/compile backends for deployment \cite{pytorch-amp}.

\section{Conclusions}
\label{sec:conclusion}
Mixed-precision quantization has emerged as an effective strategy for balancing efficiency and accuracy in language models. Our survey shows that most mixed-precision frameworks with a 4-bit average precision achieve near floating-point accuracy/perplexity, while some 2/3-bit frameworks are competitive. Compared to mixed-precision quantization techniques for DNNs, MXPLM frameworks are distinguished by their scale, memory-bound inference characteristics, and reliance on sensitivity-driven heuristics rather than search-based optimization. 

Beyond efficiency gains, MXPLM frameworks hold transformative potential for managing the computational intensity of large-scale LMs. By adaptively distributing precision budgets across layers or tokens, they can significantly mitigate the quadratic scaling of attention and reduce the massive memory footprint of KV-cache operations, the dominant cost drivers in inference. As LMs continue to expand into trillion-parameter regimes, such adaptive precision strategies could become essential not just for inference compression but for enabling training and fine-tuning within feasible energy and latency bounds.

Looking forward, several advancements are required to fully realize the potential of mixed-precision quantization for LMs. First, hardware and software co-design must evolve to natively support fine-grained mixed-precisions, enabling efficient execution without costly dequantization overhead. Second, activation and KV-cache quantization remain open bottlenecks, demanding more robust methods that preserve accuracy. Third, scalable optimization techniques that move beyond local Hessian heuristics, such as lightweight reinforcement learning or gradient methods, are needed to enable global precision allocation at billion-parameter scale. Finally, integration with multi-level memory hierarchies and dynamic precision scheduling offers a promising path toward sustainable deployment across cloud and edge platforms.

%Bibliography
%\bibliographystyle{unsrt}  
%\bibliography{references}  

\end{document}